\documentclass[fleqn,10pt]{wlscirep}
\usepackage[utf8]{inputenc}
\usepackage[T1]{fontenc}
\usepackage{hyperref}       
\usepackage{url}            
\usepackage{booktabs}       
\usepackage{amsfonts,amssymb}       
\usepackage{nicefrac}       
\usepackage{microtype}      
\usepackage{amsmath}
\usepackage{mathtools}
\usepackage{algorithm}
\usepackage[noend]{algpseudocode}
\usepackage{siunitx}
\usepackage{graphicx}
\usepackage{subcaption}
\usepackage{adjustbox}
\usepackage{color}
\usepackage{xcolor}
\usepackage{multirow}
\usepackage{longtable}
\usepackage{comment}

\newcommand{\cX}{\mathbb{X}}
\newcommand{\cZ}{\mathbb{Z}}
\newcommand{\cS}{\mathsf{S}}
\newcommand{\bbR}{\mathbb{R}}
\newcommand{\enc}{\textrm{Enc}}
\newcommand{\dec}{\textrm{Dec}}
\newcommand{\xhat}{\widehat{x}}
\newcommand{\loss}{\texttt{Loss}}
\newcommand{\nablahat}{\widehat{\nabla}}

\title{Optimizing Molecules using Efficient Queries from Property Evaluations}

\author[1]{Samuel C.~Hoffman}
\author[1]{Vijil Chenthamarakshan}
\author[1]{Kahini Wadhawan}
\author[1,*]{Pin-Yu Chen}
\author[1,*]{Payel Das}
\affil[1]{IBM Research, Yorktown Heights, NY 10598, USA}

\affil[*]{pin-yu.chen@ibm.com and daspa@us.ibm.com}

\begin{abstract}
Machine learning based methods have shown potential for optimizing existing molecules with more desirable properties, a critical step towards accelerating new chemical discovery. Here we propose \textbf{QMO}, a generic query-based molecule optimization framework that exploits latent embeddings from a molecule autoencoder. QMO improves the desired properties of an input molecule based on efficient queries, guided by a set of molecular property predictions and evaluation metrics. 
We show that QMO outperforms existing methods in the benchmark tasks of optimizing small organic molecules for drug-likeness and solubility under similarity constraints. We also demonstrate significant property improvement using QMO on two new and challenging tasks that are also important in real-world discovery problems: (i) optimizing existing potential SARS-CoV-2 Main Protease inhibitors toward higher binding affinity; and (ii) improving known antimicrobial peptides towards lower toxicity. Results from QMO show high consistency with external validations, suggesting  effective means to facilitate material optimization problems with design constraints.

\end{abstract}
\begin{document}

\flushbottom
\maketitle
%
%
\thispagestyle{empty}

\section*{Introduction}

\label{sec_intro}
Molecule optimization (MO) for improving the structural and/or functional profile of a molecule is an essential step for many scientific and engineering applications including chemistry, drug discovery, bioengineering and material science. 
Without further modeling or use of prior knowledge, the challenge of MO lies in searching over the prohibitively large space comprised of all possible molecules and generating new, valid, and optimal ones. 
In recent years, machine learning has shown to be a promising tool for MO by combining domain knowledge and data-driven learning for efficient discovery \cite{bartok2017machine,tkatchenko2020machine,button2019automated,kotsias2020direct}. Compared to traditional high-throughput  wet lab experiments or  computer simulations that are time-consuming and expensive \cite{polishchuk2013estimation, zhavoronkov2018artificial}, 
machine learning can significantly accelerate MO
by enabling iterative improvements based on instant feedback from real-time model prediction and analysis \cite{sun2019machine,ekins2019exploiting}, and thereby reducing the gap between initial discovery and subsequent optimization and production of materials for various applications.
For example, machine learning driven MO can enable prompt design of optimized  candidates starting from existing lead molecules, leading to potentially better inhibition of SARS-CoV-2 proteins. It is now well-accepted that the majority of existing drugs fail to show desired binding (and  inhibition) to SARS-CoV-2 targets, mostly due to the novel nature of the SARS-CoV-2 virus \cite{wu2020analysis,yang2020molecular}. Therefore,  optimization of  existing lead  molecules toward better SARS-CoV-2 target binding affinity while keeping the molecular similarity high appears a promising first step for optimal drug design for COVID-19.  Similarly, an efficient MO method can guide  design of antimicrobials with better-optimized toxicity to fight against resistant pathogens, one of the biggest threats to global health \cite{coates2011novel}. Without loss of generality, we refer a lead molecule as the starting molecule to be optimized in order to meet a set of desired properties and constraints.  

Many recent research studies that focus on machine learning enabled MO represent a molecule as a string consisting of chemical units. For small organic molecules, the SMILES representation \cite{weininger1988smiles} is widely used, whereas for peptide sequences, a text string comprised of amino acid characters is a popular representation. 
Often, for efficiency reasons, the optimization is performed on a learned  representation space of the system of interest, which describes molecules as embedding vectors in a low-dimensional continuous space. 
A sequence-to-sequence encoder-decoder model, such as a (variational) autoencoder,
can be used to learn continuous representations of the molecules in a latent space. Moreover, different optimization or sampling techniques based on the latent representation can be used to improve a molecule with external guidance from a set of molecular property predictors and simulators. 
The external guidance can be either explicitly obtained from physics-based simulations, (chem/bio-)informatics, wet-lab experiments, or implicitly learned from a chemical database. 

Based on the methodology, the related works on machine learning for MO can be divided into two categories: \textit{guided search} and \textit{translation}.
Guided search uses guidance from the predictive models and/or evaluations from statistical models, where the search can be either in the discrete molecule sequence space or through a continuous latent space (or distribution) learned by an encoder-decoder. Genetic algorithms \cite{reutlinger2014multi,yuan2011ligbuilder,nigam2019augmenting} and Bayesian optimization (BO) \cite{korovina2019chembo} 
have been proposed for searching in the discrete sequence space, but their efficiency can be low in the case of high search dimension. Recent works have exploited latent representation learning and different optimization/sampling techniques for efficient search. Examples include the combined use of variational autoencoder (VAE) and BO \cite{gomez2018automatic,skalic2019shape,griffiths2020constrained,jimenez2019deltadelta},
VAE and Gaussian sampling \cite{Boitreaud_optimol},
VAE and sampling guided by a predictor \cite{jin2018junction,fu2019core}, VAE and evolutionary algorithms \cite{winter2019efficient}, deep reinforcement learning and/or a generative network \cite{olivecrona2017molecular,guimaraes2017objective,sanchez2017optimizing,you2018graph,zhou2019optimization}, and attribute-guided rejection sampling on an autoencoder \cite{das2020accelerating}.
On the other hand, translation-based approach treats molecule generation as a sequence-to-sequence translation problem \cite{griffen2011matched,dossetter2013matched,dalke2018mmpdb,bahdanau2014neural}. Examples are translation with
junction-tree \cite{jin2018learning,yang2020improving}, shape features \cite{skalic2019shape}, hierarchical graph \cite{jin2019hierarchical}, and transfer learning  \cite{maragakis2020deep}.
Comparing to guided search, translation-based approaches require the additional knowledge of paired sequences for learning to translate a lead molecule into an improved molecule. This knowledge may not be available for new MO tasks with limited information. For example, in the task of optimizing a set of known inhibitor molecules to better bind to SARS-CoV-2 target protein sequence while preserving the desired drug properties,  a sufficient number of such paired molecule sequences is unavailable.
We also note that these two categories are not exclusive. Guided search can be jointly used with translation.


\begin{figure}[t]
\centering
\includegraphics[width=\textwidth]{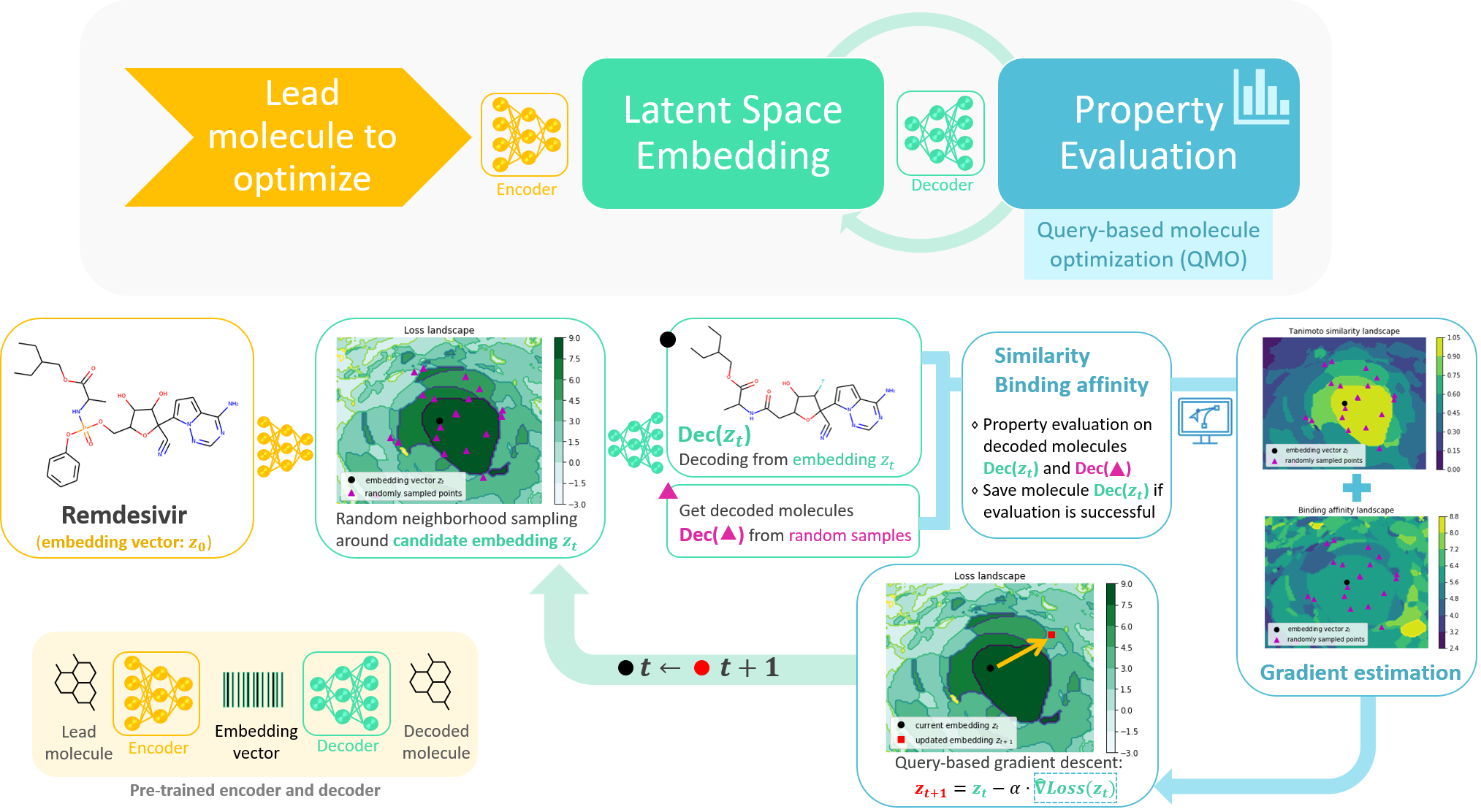}
  \caption{System illustration of the proposed query-based molecule optimization (QMO) framework. The QMO system progressively optimizes an input lead molecule (e.g., Remdesivir) according to a set of user-specified properties (e.g., binding affinity and Tanimoto similarity) by leveraging the learned molecule embeddings from a pair of pre-trained encoder and decoder (i.e. an autoencoder), and by evaluating the properties of the generated molecules. Given a candidate embedding $z_t$ at the optimization step $t$, QMO randomly samples the neighboring vectors of  $z_t$ in the embedding space, evaluates the properties of the corresponding decoded molecules, and uses the evaluations for gradient estimation (see equation \eqref{eqn_gradient_est}) and queried-based gradient descent (see equation \eqref{eqn_ZO_descent}) 
   for finding the next candidate embedding vector $z_{t+1}$.    }
  \label{fig:QMO_system}
\end{figure}

In this paper, we propose a novel \textbf{Q}uery-based \textbf{M}olecule \textbf{O}ptimization (QMO) framework, as illustrated in Figure \ref{fig:QMO_system}.
In this context, a query to a designed loss function for QMO gives the corresponding numerical value obtained through the associated property evaluations. Efficiency refers to the performance of the optimization results given a query budget.
QMO uses an encoder-decoder and external guidance, but it differs from existing works in the following aspects: (i) QMO is a generic end-to-end optimization framework that reduces the problem complexity by decoupling representation learning and guided search. It applies to any \textit{plug-in} (pre-trained) encoder-decoder with continuous latent representations. It is also a unified and principled approach that incorporates multiple predictions and evaluations made directly at the molecule sequence level into guided search without further model fitting. (ii) To achieve efficient end-to-end optimization with discrete molecule sequences and their continuous latent representations, QMO adopts a novel query-based guided search method based on zeroth order optimization \cite{ghadimi2013stochastic,liu2020primer}, a technique that performs efficient mathematical optimization using only function evaluations (see Section \ref{sec_ZO_opt} in the Supplementary Material for more details). 
Its query-based guided search enables direct optimization over the property evaluations provided by chemical informatics/simulation software packages or prediction APIs, and it supports guided search with exact property evaluations that only operate at the molecular sequence level instead of latent representations or surrogate models. To the best of our knowledge, this work is the first study that facilitates molecule optimization by disentangling molecule representation learning and guided search, and by exploiting zeroth order optimization for efficient search in the molecular property landscape. The success of QMO can be attributed to its data efficiency, by exploiting the latent representations learned from abundant unlabeled data and the guidance for property prediction trained on relatively limited labeled data.


We first demonstrate the effectiveness of QMO through two sets of standard benchmarks. On
 two existing and simpler MO benchmark tasks of optimizing 
 drug-likeness (QED) \cite{bickerton2012quantifying} and penalized logP (reflecting octanol-water partition coefficient) \cite{jin2018junction} with similarity constraints, QMO attains superior performance over existing baselines, showing at least 15\% higher success on QED optimization and an absolute improvement of 1.7 on penalized logP.   Performance of QMO on these molecular physical property benchmarks shows its potential for optimizing the material design prior to synthesis, which is critical in many applications, such as in food industry, agrochemicals, pesticides, drugs, catalysts, and waste chemicals.

\begin{figure}[t]
\centering
  \begin{subfigure}[t]{0.38\textwidth}
    \centering
    \includegraphics[width=\textwidth]{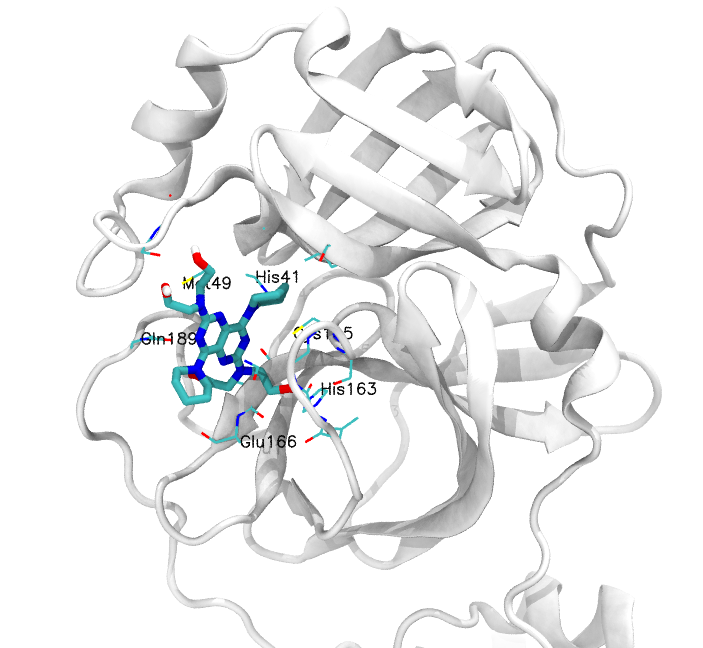}
    \caption{Original (Dipyridamole)}
    \label{fig:dipyridamole_orig}
  \end{subfigure}
  \hfill
  \begin{subfigure}[t]{0.38\textwidth}  
    \centering 
    \includegraphics[width=\textwidth]{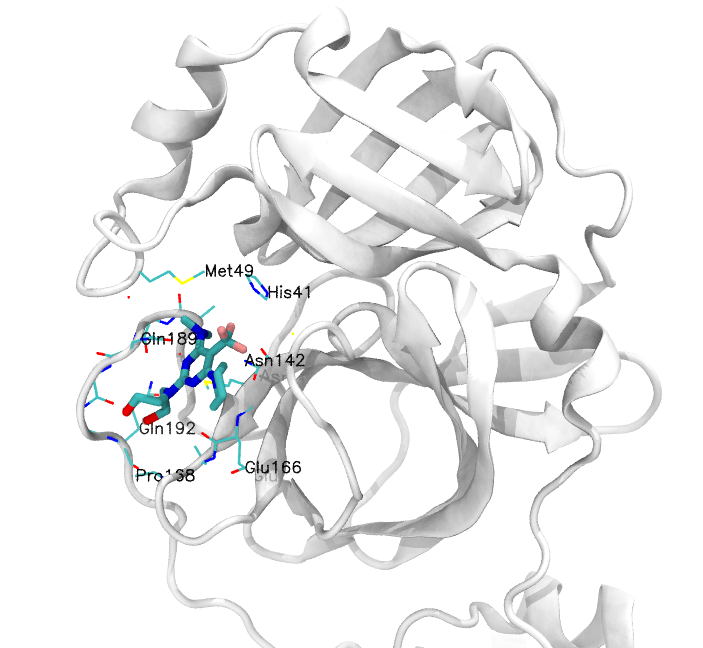}
    \caption{QMO-optimized}
    \label{fig:dipyridamole_imp}
  \end{subfigure}
  \hfill
  \begin{minipage}[b][2.6in][t]{0.21\textwidth}
    \centering
    \begin{subfigure}[t]{\textwidth}
      \includegraphics[width=\textwidth,trim={0 55px 0 45px},clip]{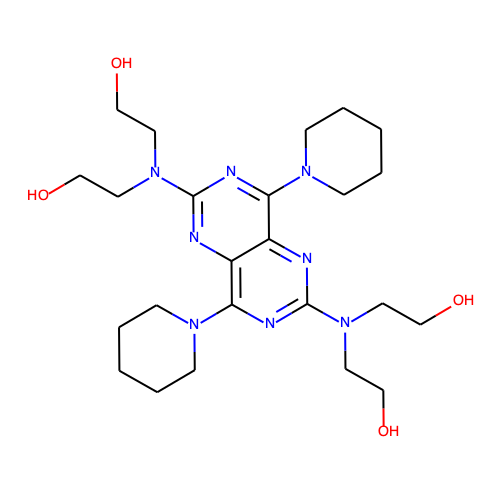}
      \caption{Original (Dipyridamole)}
    \end{subfigure}
    \vfill
    \begin{subfigure}[t]{\textwidth}
      \includegraphics[width=\textwidth,trim={0 45px 0 45px},clip]{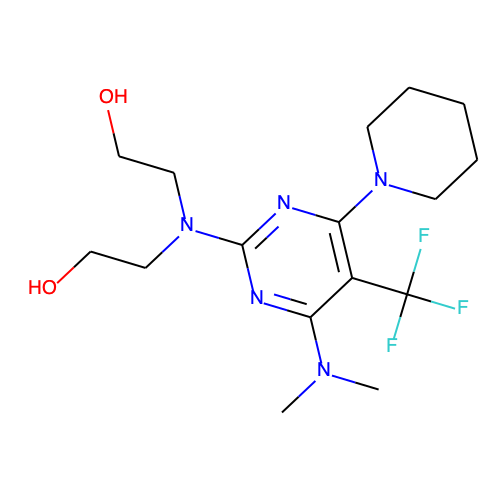}
      \caption{QMO-optimized}
    \end{subfigure}
  \end{minipage}
  \caption{Top docking poses of Dipyridamole and its QMO-optimized variant with SARS-CoV-2 M\textsuperscript{pro}, as obtained using AutoDock Vina. Their 2D structures are also shown. QMO optimizes the predicted affinity for the Dipyridamole variant from 3.94 to 7.59, while maintaining a Tanimoto similarity score of 0.58 and without changing the binding pocket significantly. MM/PBSA calculations for these poses show binding free energy improvement from $-11.49$ to $-25.65$ kcal/mol. Important residues from M\textsuperscript{pro} substrate-binding pocket are also shown. See Table \ref{table_covid_targets} in the Supplementary Material for details.}
  \label{fig:dipyridamole}
\end{figure}

Next, as a motivating discovery use-case 
that also, at least to some extent,  reflect the complexity of real discovery problems \cite{coley2019autonomous}, we demonstrate how QMO can be used to improve binding affinity of a number of  existing inhibitor molecules to the SARS-CoV-2 Main Protease (M\textsuperscript{pro}) , one of the most extensively studied drug targets for SARS-CoV-2.  As an illustration,  Figure \ref{fig:dipyridamole} shows the top docking poses of Dipyridamole and its QMO-optimized variant with SARS-CoV-2 M\textsuperscript{pro}.
We formulate this task as an optimization over predicted binding affinity (obtained using a pre-trained machine learning model) starting from an existing molecule of interest  (i.e. a lead molecule).
Since experimental $\text{IC}_{50}$ values are widely available, we use them  ($\text{pIC}_{50} = -\log_{10} (\text{IC}_{50})$) as a measure for protein-ligand binding affinity.  
 $\text{pIC}_{50}$ of the optimized molecule is constrained to be above 7.5, a sign of good affinity, while the Tanimoto similarity between the optimized and the original molecule is maximized. Retaining high similarity while optimizing the initial lead molecule means important chemical characteristics can be maximally preserved. Moreover,
a high similarity to existing leads is important for rapid response to a novel pathogen such as SARS-CoV-2, as then it is more likely to leverage existing knowledge and manufacturing pipeline for synthesis and wet lab evaluation of the optimized variants. Moreover, the chance of optimized variants inducing adverse effects is potentially low. Our results
show that QMO can find  molecules with high similarity and improved affinity, while preserving other properties of interest such as drug-likeness. 

  We also consider the task of optimizing existing antimicrobial peptides toward lower selective toxicity, which is critical for accelerating safe antimicrobial discovery. In this task QMO shows high success rate ($\sim$ 72\%) in improving the toxicity of antimicrobial peptides, and the properties of optimized molecules are consistent with external toxicity and antimicrobial activity classifiers. Finally, we perform property landscape visualization and trajectory analysis of QMO to illustrate its efficiency and diversity in finding improved molecules with desired properties.
  
  We emphasize that QMO is a generic-purpose optimization algorithm that enables optimization over discrete spaces (e.g. sequences, graphs), which involves  searching over a latent space of the system by using guidance from (expensive) black-box function evaluations. Beyond the organic and biological molecule optimization applications considered in this study, QMO can be applied to optimization of other classes of materials, e.g.  inorganic solid-state materials like metal alloys or metal oxides.

\section*{Results}


\paragraph{Representations of Molecules}
In our QMO framework, we model a molecule as a discrete string of chemical or amino acid characters (i.e. a sequence). Depending on the downstream MO tasks, the sequence representation can either be a string of natural amino acids \cite{qin2020artificial, das2020accelerating}, or a string designed for encoding small organic chemicals. In particular, the simplified molecular input line entry specification (SMILES) representation \cite{weininger1988smiles} describes the structure of small organic molecules  using short ASCII strings.
Without loss of generality, we define $\cX^m:=\cX \times \cX \cdots \times \cX$ as the product space containing every possible molecule sequence of length $m$, where $\cX$ denotes the set of all chemical characters. 
To elucidate the problem complexity, considering the 20 protein-building amino acids as characters in a peptide sequence,
the number of possible candidates in the space of  sequences with length $m=60$ is already reaching the number of atoms in the known universe ($\sim$ $10^{80}$). Similarly,
the space of small molecules with therapeutic potential is estimated to be on the order of $10^{60}$ \cite{bohacek1996art,reymond2012enumeration}.
Therefore, the problem of MO in the ambient space $\cX^m$ can be computationally inefficient as the search space grows combinatorially with the sequence length $m$. 

\paragraph{Encoder-Decoder for Learning Latent Molecule Representations}

To address the issue of large search space for molecule sequences, QMO adopts an encoder-decoder framework.
The encoder $\enc: \cX^m \mapsto \bbR^d$ encodes a sequence $x \in \cX^m$ to a low-dimensional continuous real-valued representation of dimension $d$, denoted by an embedding vector $z=\enc(x)$.
The decoder  $\dec: \bbR^d \mapsto \cX^{m^\prime}$ decodes the latent representation $z$ of $x$ back to the sequence representation, denoted by $\xhat=\dec(z)$. We note that depending on the encoder-decoder implementation, the input sequence
$x$ and the decoded sequence $\xhat$ may be of different length. On the other hand, the latent dimension $d$ is universal (fixed) to all sequences. In particular, Winter et al. \cite{winter2019learning} proposed a novel molecular descriptor and used it for an autoencoder to learn latent representations featuring high similarity between the original and the reconstructed sequences.
QMO 
applies to any \textit{plug-in} (pre-trained) encoder-decoder with continuous latent representations and thus decouples representation learning and guided search, in order to reduce the problem complexity of MO.


\paragraph{Molecule Optimization Formulation via Guided Search}
\label{QMO_formulation}

In addition to leveraging learned latent representations from a molecule encoder-decoder, our QMO framework incorporates molecular property prediction models and similarity metrics at the sequence level as external guidance. Specifically, for any given sequence $x \in \cX^m$, we use 
a set of $I$ separate prediction models $\{f_i(x)\}_{i=1}^I$ to evaluate the properties of interest for MO. 
In principle, for a candidate sequence $x$, a set of thresholds $\{\tau_i\}_{i=1}^I$ on its property predictions $\{f_i(x)\}_{i=1}^I$ is used for validating the condition $f_i(x) \geq \tau_i$ for all $i  \in [I]$, where $[I]$ denotes the integer set $\{1,2,\ldots,I\}$.
Moreover, we can simultaneously impose a set of $J$ separate constraints $\{g_j(x|\cS) \geq \eta_j \}_{j=1}^J$ in the optimization process, such as molecular similarity,  relative to a set of reference molecule sequences denoted by $\cS$. 

Our QMO framework covers two practical cases in MO: (i) \textit{optimizing molecular similarity while satisfying desired chemical properties} and (ii) \textit{optimizing chemical properties with similarity constraints}. It can be easily extended to other MO settings that can be formulated via $\{f_i(x)\}_{i=1}^I $ and  $\{g_j(x|\cS)\}_{j=1}^J $.
In what follows, we formally define our designed loss function of QMO for Case (i). Given a starting molecule sequence $x_0$ (i.e. a lead molecule) and a pre-trained encoder-decoder, let $x=\dec(z)$ denote a candidate sequence decoded from a latent representation $z\in \bbR^d$.
Our QMO framework aims to find an optimized sequence by solving the following continuous optimization problem:
\begin{align}
    \label{eq_unconstrained_search}
    \text{Minimize}_{z \in \bbR^d}~\underbrace{\sum_{i=1}^I \max\{\tau_i - f_i(\dec(z)),0\}}_{\text{property validation loss (to be minimized)} }
    - \underbrace{\sum_{j=1}^J \lambda_j \cdot  g_j(\dec(z)|\cS)}_{\text{molecular score (to be maximized)}}
\end{align}
The first term $\sum_{i=1}^I \max\{\tau_i - f_i(\dec(z)),0\}$ quantifies the loss of property constraints and 
is presented as the sum of hinge loss over all property predictions, which approximates the binary property validation relative to the required thresholds $\{\tau_i\}_{i=1}^I$. It achieves the optimal value (i.e. 0) only when the candidate sequence $x=\dec(z)$ satisfies all the desired properties, which is equivalent to the condition that $f_i(\dec(z)) \geq \tau_i$ for all $i \in [I]$. 
The second term $\sum_{j=1}^J \lambda_j \cdot  g_j(\dec(z)|\cS)$ corresponds to a set of molecular similarity scores to be maximized (therefore a minus sign in the minimization formulation). The reference sequence set $\cS$ can be the starting sequence such that $\cS=\{x_0\}$, or a set of molecules.
The positive coefficients $\{\lambda_j\}_{j=1}^J$ are associated with the set of molecular similarity scores $\{g_j(\dec(z)|\cS)\}_{j=1}^J$, respectively.
It is worth mentioning that the use of the latent representation $z$ as the optimization variable in a low-dimensional continuous space 
greatly facilitates the original MO problem in a high-dimensional discrete space.  The optimization variable $z$ can be initialized as the latent representation of $x_0$, denoted by $z_0 = \enc(x_0)$. 

Similarly, for Case (ii), the  optimization problem is formulated as
\begin{align}
    \label{eq_unconstrained_search_2}
    \text{Minimize}_{z \in \bbR^d}~
    \underbrace{\sum_{j=1}^J  \max\{  \eta_j - g_j(\dec(z)|\cS), 0\} }_{\text{molecular constraint loss (to be minimized)}} -
    \underbrace{\sum_{i=1}^I \gamma_i \cdot f_i(\dec(z))}_{\text{property score (to be maximized)}}
\end{align}
where  $\{\eta_j\}_{j=1}^J$ are the similarity score constraints and $\{\gamma_i\}_{i=1}^I$ are positive coefficients of the property scores $\{f_i(\dec(z))\}_{i=1}^I$.

\paragraph{Query-based Molecule Optimization (QMO) Procedure}
\label{subsec_QMO_algo}

Although we formulate MO as an unconstrained continuous minimization problem,
we note that solving it for a feasible candidate sequence $x=\dec(z)$ is not straightforward because: (i) The output of the decoder $x=\dec(z)$ is a discrete sequence, which imposes challenges on any gradient-based (and high-order) optimization method since acquiring the gradient of $z$ becomes non-trivial. Even resorting to the Gumbel-softmax sampling trick for discrete outputs \cite{jang2017categorical}, the large output space of the decoder may render it ineffective; (ii) In practice, many molecular property prediction models and molecular metrics are computed in an access-limited environment, such as prediction APIs and chemical softwares, which only allow inference on a queried sequence but prohibit other functionalities such as gradient computation. To address these two issues, we use zeroth order optimization in our QMO framework (see Methods section for detailed procedure) to provide a generic and model-agnostic approach for solving the problem formulation in \eqref{eq_unconstrained_search} and \eqref{eq_unconstrained_search_2} using \textit{only} inference results of $\{f_i\}_{i=1}^I$ and $\{g_j\}_{j=1}^J$ on queried sequences.

Let $\loss(z)$ denote the objective function to be minimized, as defined in either \eqref{eq_unconstrained_search} or \eqref{eq_unconstrained_search_2}. 
Our QMO framework uses zeroth order gradient descent to find a solution, which mimics the descent steps on the loss landscape in gradient-based solvers but only uses the function values \loss($\cdot$) of queried sequences. Specifically,
at the $t$-th iteration of the zeroth order optimization process, the iterate (candidate embedding vector) $z^{(t+1)}$ is updated by 
\begin{align}
    \label{eqn_ZO_descent}
    z^{(t+1)} = z^{(t)} - \alpha_t \cdot \nablahat \loss(z^{(t)}),
\end{align}
where $\alpha_t \geq 0$ is the step size at the $t$-th iteration, and the true gradient  $\nabla \loss(z^{(t)})$ (which is challenging or infeasible to compute) is approximated by the pseudo gradient $\nablahat \loss(z^{(t)})$. The pseudo gradient $\nablahat \loss(z^{(t)})$ is estimated by $Q$ independent random directional queries defined as
\begin{align}
    \label{eqn_gradient_est}
    \nablahat \loss(z^{(t)}) = \frac{d}{\beta \cdot Q} \sum_{q=1}^Q \left[ \loss(z^{(t)}+\beta u^{(q)}) - \loss(z^{(t)}) \right]   \cdot u^{(q)},
\end{align}
where $d$ is the dimension of the latent space of the encoder-decoder used in QMO, and $\beta > 0$ is a smoothing parameter used to perturb the embedding vector $z^{(t)}$ for neighborhood sampling with $Q$ random directions $\{u^{(q)}\}_{q=1}^Q$ that are independently and identically sampled on a $d$-dimensional unit sphere. See Figure \ref{fig:QMO_system} for the illustration of random neighborhood sampling.
In our implementation, we sample $\{u^{(q)}\}_{q=1}^Q$ using a zero-mean $d$-dimensional isotropic Gaussian random vector divided by its Euclidean norm, such that the resulting samples 
are drawn uniformly from the unit sphere. Intuitively, the gradient estimator in \eqref{eqn_gradient_est} can be viewed as an average of $Q$ random directional derivatives along the sampled directions $\{u^{(q)}\}_{q=1}^Q$. The constant $\frac{d}{\beta \cdot Q}$ in \eqref{eqn_gradient_est} ensures the norm of the estimated gradient is at the same order as that of the true gradient\cite{ghadimi2013stochastic,liu2020primer}. 

A schematic example of the QMO procedure is illustrated in Figure \ref{fig:QMO_system} using binding affinity and Tanimoto similarity as property evaluation criterion.
Note that based on the iterative optimization step in \eqref{eqn_ZO_descent}, QMO only uses function values queried at the original and perturbed sequences for optimization.
The query counts made on the \loss  
~function for computing $\nablahat \loss(z^{(t)})$ is $Q+1$ per iteration. 
Larger $Q$ further reduces the gradient estimation error at the price of increased query complexity.
When solving \eqref{eq_unconstrained_search}, an iterate $z^{(t)}$ is considered as a valid solution if its decoded sequence $\dec(z^{(t)})$ satisfies the property conditions $f_i(\dec(z^{(t)})) \geq \tau_i$ for all $i \in [I]$. Similarly, when solving \eqref{eq_unconstrained_search_2}, a valid solution $z^{(t)}$ means $g_j(\dec(z^{(t)}|\cS)) \geq \eta_j$ for all $j \in [J]$. Finally, QMO returns a set of found solutions (returning null if in vain). Detailed descriptions for the QMO procedure are given in Methods section.

\paragraph{Three Sets of Molecule Optimization Tasks with Multiple Property Evaluation Criterion}
In what follows, we demonstrate the performance of our proposed QMO framework on three sets of tasks that aim to optimize molecular properties with constraints, including standard MO benchmarks and challenging tasks relating to real-world discovery problems.
The pre-trained encoder-decoder and the hyperparameters of QMO for each task are specified in Methods section and in Section \ref{appendix_model_data} of the Supplementary Material.






\paragraph{Benchmarks on QED and Penalized logP Optimization}
\label{subsec_QED_log}

We start with testing QMO on two single property targets: penalized logP and Quantitative Estimate of Drug-likeness (QED) \cite{bickerton2012quantifying}. LogP
is the logarithm of the partition ratio of the solute between octanol and water. Penalized logP is defined as the logP minus the synthetic accessibility (SA) score  \cite{jin2018junction}.
Given a similarity constraint, finding an optimized molecule that improves drug-likeness of compounds using the QED score (from range [0.7, 0.8] to [0.9, 1.0]) \cite{bickerton2012quantifying} or improves the penalized logP score \cite{jin2018junction}, are two widely used benchmarks. For a pair of original and optimized sequences $(x_0,x)$,
we use the QMO formulation in \eqref{eq_unconstrained_search_2} with the Tanimoto similarity (ranging from 0 to 1) over
Morgan fingerprints \cite{rogers2010extended} as $g_{\text{Tanimoto}}(x|x_0)$ and the interested property score (QED or penalized logP) as $f_{\text{score}}(x)$. Following the same setting as existing works,
the threshold $\delta$ for $g_{\text{Tanimoto}}(x|x_0)$ is set as either 0.4 or 0.6. We use RDKit\footnote{RDKit: Open-source cheminformatics; \url{http://www.rdkit.org}} to compute QED and logP, and use MOSES \cite{polykovskiy2018molecular} to compute SA (synthetic accessibility), where $f_{\text{penalized logP}}(x)=\text{logP}(x) - \text{SA}(x)$.

In our experiments, we use the same set of 800 molecules with low penalized logP scores and 800 molecules with QED $\in[0.7,0.8]$ chosen from the ZINC test set \cite{sterling2015zinc} as in Jin et al. \cite{jin2018junction} as our starting sequences. We 
compare QMO with various guided-search and translation-based methods in Tables \ref{table_QED} and \ref{table_penalized_logP}. Baseline results are obtained from the literature \cite{jin2018learning,maragakis2020deep} that use machine learning for solving the same task. 

For the QED optimization task, the success rate defined as the percentage of improved molecules having similarity greater than $\delta=0.4$ is shown in Table \ref{table_QED}.
QMO outperforms all baselines by at least 15\%. For penalized logP task,
the molecules optimized by QMO outperform the baseline results by a significant margin, as shown in Table \ref{table_penalized_logP}. The increased standard deviation in QMO is an artifact of having some molecules with much improved penalized logP scores (see Section \ref{appendix_addtl_vis} in the Supplementary Material).

\begin{table}
\begin{minipage}[t]{.35\textwidth}
\centering
\adjustbox{max width=0.76\textwidth}{
\begin{tabular}[t]{l|l}
\toprule
Method
 & Success (\%) \\ \midrule
 MMPA \cite{dalke2018mmpdb} & 32.9 \\ 
 JT-VAE \cite{jin2018junction} & 8.8 \\ 
 GCPN \cite{you2018graph} & 9.4 \\ 
 VSeq2Seq \cite{bahdanau2014neural} &  58.5 \\ 
 VJTNN+GAN \cite{jin2018learning}& 60.6  \\ 
 AtomG2G \cite{jin2019hierarchical} & 73.6 \\ 
 HierG2G \cite{jin2019hierarchical} & 76.9 \\ 
 DESMILES \cite{maragakis2020deep} & 77.8 \\ \hline
  \textbf{QMO} & \textbf{92.8} \\ 
 \bottomrule
\end{tabular}
}
\caption{Performance of drug likeness (QED) task with Tanimoto similarity constraint $\delta=0.4$.}
\label{table_QED}
\end{minipage}%
\hfill
\hspace{-7mm}
\begin{minipage}[t]{.36\textwidth}
\centering
\adjustbox{max width=0.95\textwidth}{
\begin{tabular}[t]{l|cc}
\toprule
Method & \multicolumn{2}{c}{Improvement}\\
& $\delta=0.6$ & $\delta=0.4$ \\ \midrule
JT-VAE \cite{jin2018junction} & 0.28 $\pm$ 0.79 & 1.03 $\pm$ 1.39\\
GCPN \cite{you2018graph} & 0.79 $\pm$ 0.63 & 2.49 $\pm$ 1.30\\
MolDQN \cite{zhou2019optimization} & 1.86 $\pm$ 1.21 & 3.37 $\pm$ 1.62\\
VSeq2Seq \cite{bahdanau2014neural} & 2.33 $\pm$ 1.17 & 3.37 $\pm$ 1.75\\
VJTNN \cite{jin2018learning} & 2.33 $\pm$ 1.24 & 3.55 $\pm$ 1.67 \\ 
GA \cite{nigam2019augmenting} & 3.44 $\pm$ 1.09 & 5.93 $\pm$ 1.41 \\ \hline
\textbf{QMO} & $\mathbf{3.73 \pm 2.85}$ & $\mathbf{7.71 \pm 5.65}$\\
\bottomrule
\end{tabular}
}
\caption{Performance of penalized logP task at various Tanimoto similarity constraint value $\delta$.}
\label{table_penalized_logP}
\end{minipage}
\hfill
  \begin{minipage}[t]{0.28\textwidth}\vspace{-10pt}%
    \centering
    \includegraphics[width=1.05\textwidth]{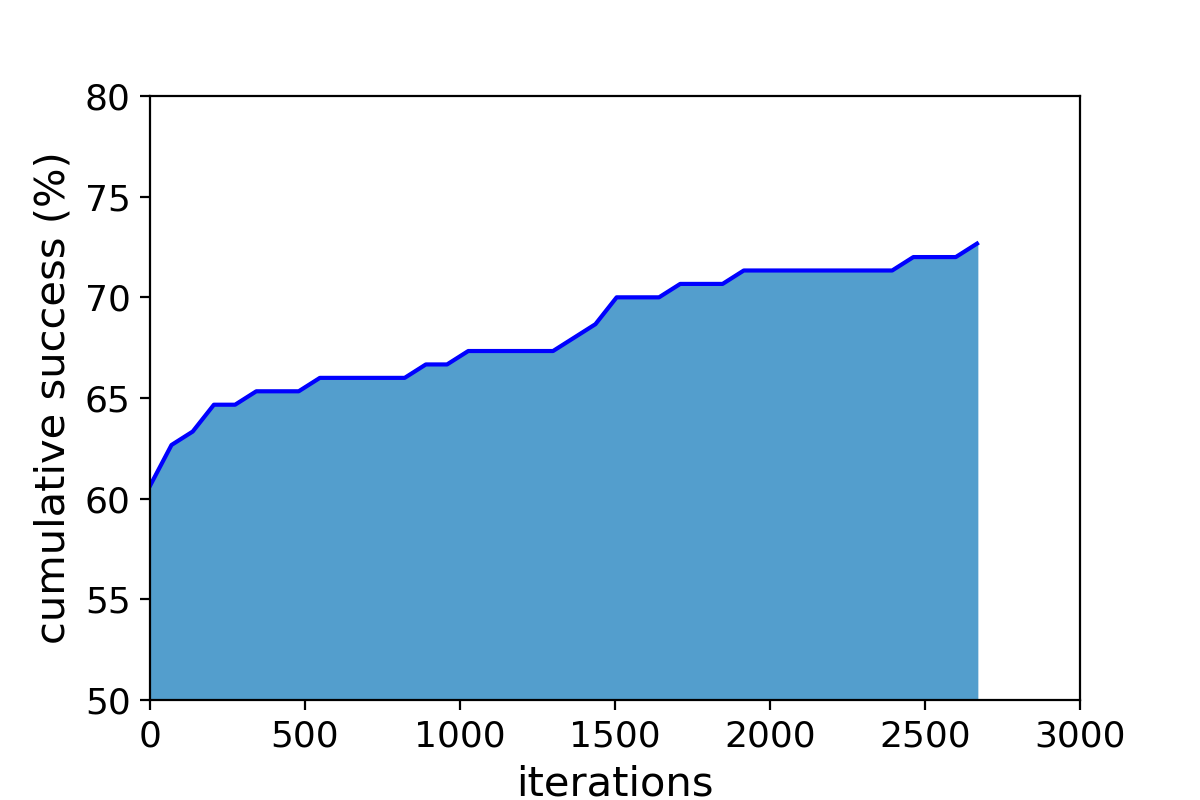}
    \captionof{figure}{Cumulative success rate of antimicrobial peptide (AMP) sequence optimization v.s. iterations using QMO.}
    \label{fig:cdf_AMP}
  \end{minipage}
\end{table}

Although the above-mentioned molecular property optimization tasks provide well-defined benchmarks for testing our QMO algorithm, it is well-recognized that such tasks are easy to solve and  do not capture the complexity associated with real-world   discovery \cite{brown2019guacamol}. For example, it is trivial to achieve state-of-the-art results for logP optimization by generating long saturated hydrocarbon chains \cite{RENZ2020}. Coley et al. \cite{coley2019autonomous} has proposed that molecular optimization goals that better reflect the complexity of real discovery tasks might include binding or selectivity attribute. Therefore, in the remaining of this paper, we consider two such tasks: (1) optimizing binding affinity of existing SARS-CoV-2 M\textsuperscript{pro} inhibitor molecules  and (2) lowering toxicity of known antimicrobial peptides. 

\paragraph{Optimizing Existing SARS-CoV-2 Main Protease Inhibitor Molecules toward better IC$_{50}$.}
\label{subsec_COVID}

To provide a timely solution and accelerate the drug discovery against a new virus such as SARS-CoV-2, it is a sensible practice to optimize known leads to facilitate design and production as well as minimize the emergence of adverse effects. 
Here we focus on the task of optimizing the parent molecule structure of a set of existing SARS-CoV-2 M\textsuperscript{pro} inhibitors.
Specifically, we use the QMO formulation in \eqref{eq_unconstrained_search}, a pre-trained binding affinity predictor\cite{chenthamarakshan2020targetspecific} $f_{\text{affinity}}$ (output is $\text{pIC}_{50}$ value), and the Tanimoto similarity $g_{\text{Tanimoto}}$  between the original and optimized molecules.
Given a known inhibitor molecule $x_0$, we aim to find an optimized molecule $x$ such that $f_{\text{affinity}}(x)\geq \tau_{\text{affinity}}$ while $g_{\text{Tanimoto}}(x|x_0)$ is maximized.

For this task, we start by assembling 23 existing molecules shown to have weak to moderate affinity with SARS-CoV-2 M\textsuperscript{pro} \cite{jin2020structure,huynh2020insilico}. These are generally in the \si{\micro \meter} range of $\text{IC}_{50}$, a measure of inhibitory potency (see Section \ref{appendix_model_data} in the Supplementary Material for experimental $\text{IC}_{50}$ values). We choose the target affinity threshold $\tau_{\text{affinity}}$ 
as $\text{pIC}_{50}\ge7.5$, which implies strong affinity. 
Table \ref{tab:selected_covid_targets} shows the final optimized molecules compared to their initial state (i.e., the original lead molecule). We highlight common substructures and show a similarity map to emphasize the changes.
The results of all 23 inhibitors
are summarized in Table \ref{table_covid_targets} of the Supplementary Material.

\begin{table}[t]
\centering
\adjustbox{max width=0.97\textwidth}{
\begin{tabular}{lp{0.24\textwidth}p{0.24\textwidth}p{0.24\textwidth}p{0.24\textwidth}}
  \toprule
  & \multicolumn{1}{c}{Dipyridamole} & \multicolumn{1}{c}{Favipiravir} & \multicolumn{1}{c}{Umifenovir} & \multicolumn{1}{c}{Kaempferol} \\
  \midrule
  \multirow{2}{*}{Original} & \parbox[c]{0.24\textwidth}{\includegraphics[width=0.24\textwidth]{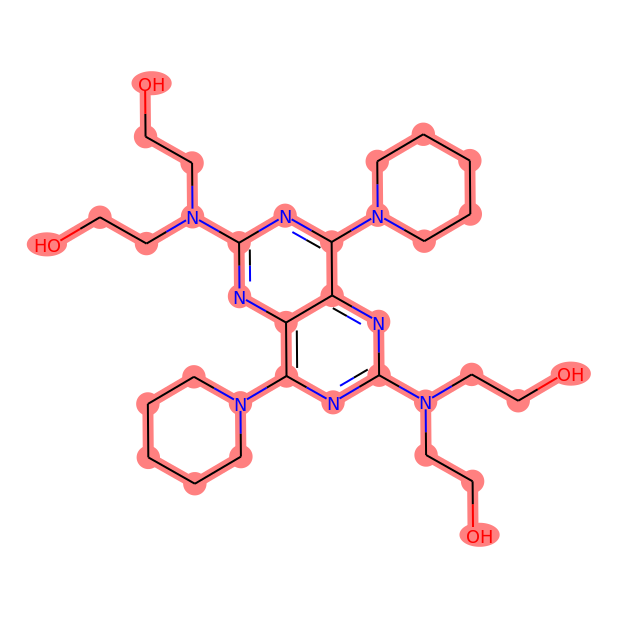}} &
  \parbox[c]{0.24\textwidth}{\includegraphics[width=0.24\textwidth]{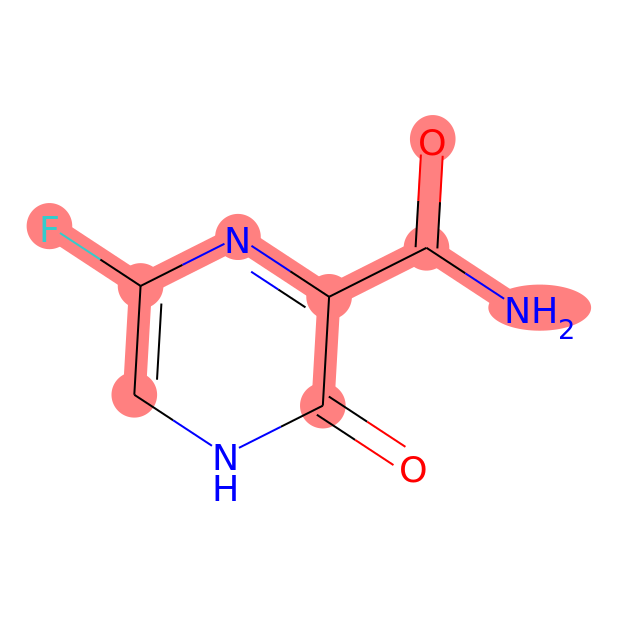}} &
  \parbox[c]{0.24\textwidth}{\includegraphics[width=0.24\textwidth]{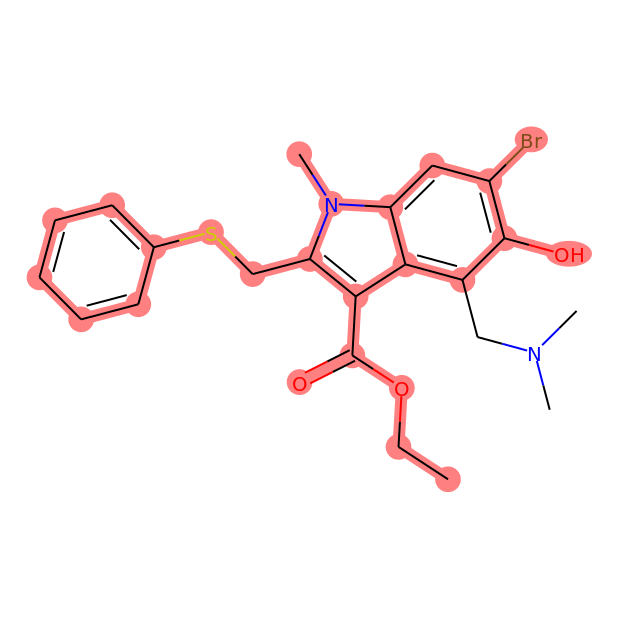}} &
  \parbox[c]{0.24\textwidth}{\includegraphics[width=0.24\textwidth]{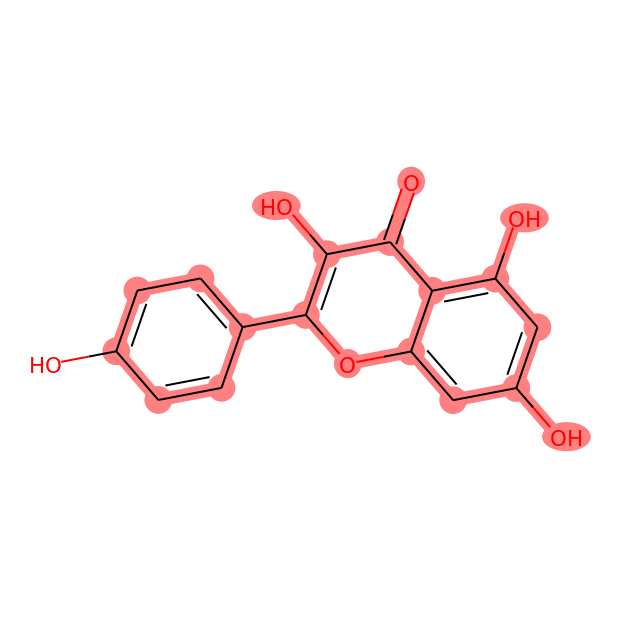}} \\
  & \multicolumn{1}{c}{$-11.49$} & \multicolumn{1}{c}{$-0.77$} & \multicolumn{1}{c}{$-16.08$} & \multicolumn{1}{c}{$-11.86$} \\
  \midrule
  \multirow{2}{*}{Similarity} & \parbox[c]{0.24\textwidth}{\includegraphics[width=0.24\textwidth]{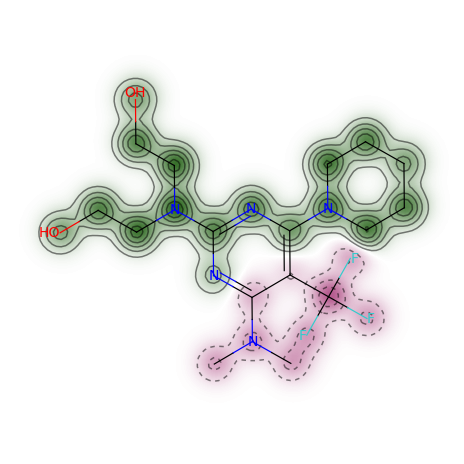}} &
  \parbox[c]{0.24\textwidth}{\includegraphics[width=0.24\textwidth]{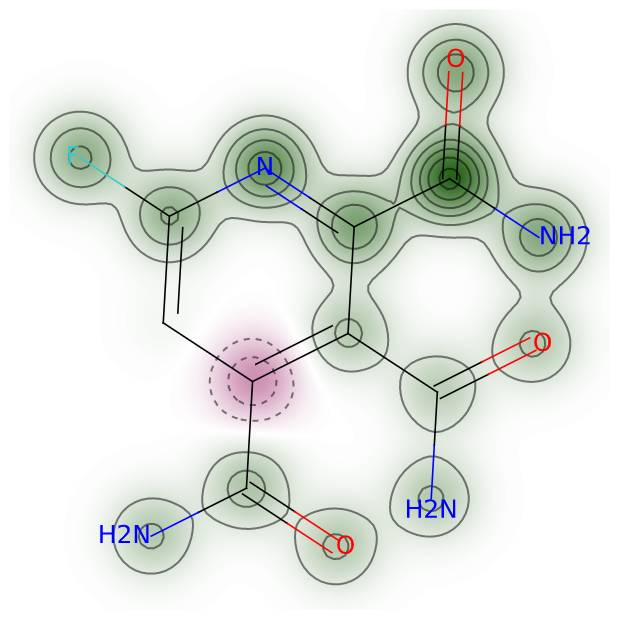}} &
  \parbox[c]{0.24\textwidth}{\includegraphics[width=0.24\textwidth]{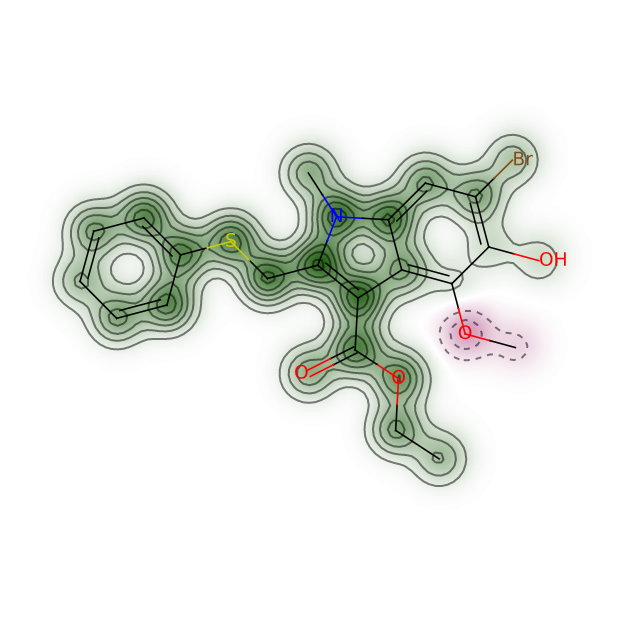}} &
  \parbox[c]{0.24\textwidth}{\includegraphics[width=0.24\textwidth]{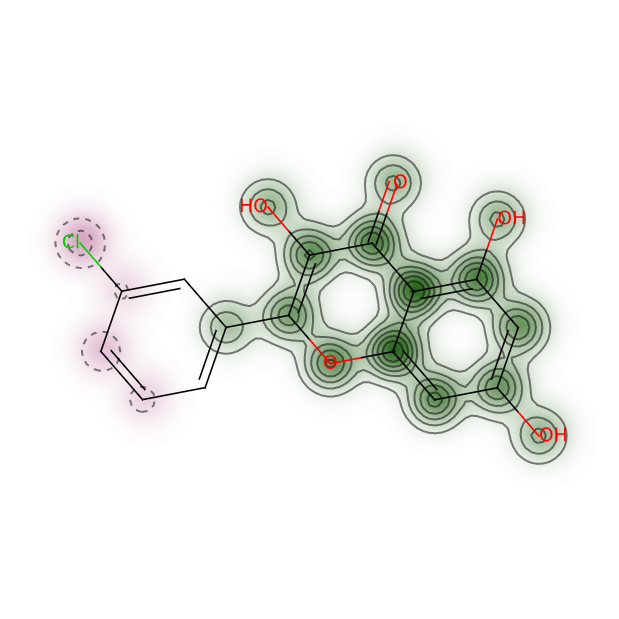}} \\
  & \multicolumn{1}{c}{0.58} & \multicolumn{1}{c}{0.46} & \multicolumn{1}{c}{0.73} & \multicolumn{1}{c}{0.67} \\
  \midrule
  \multirow{2}{*}{Optimized} & \parbox[c]{0.24\textwidth}{\includegraphics[width=0.24\textwidth]{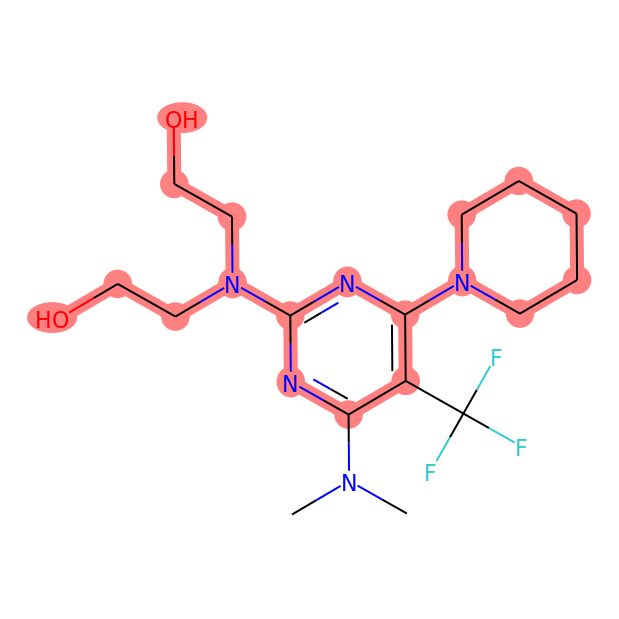}} &
  \parbox[c]{0.24\textwidth}{\includegraphics[width=0.24\textwidth]{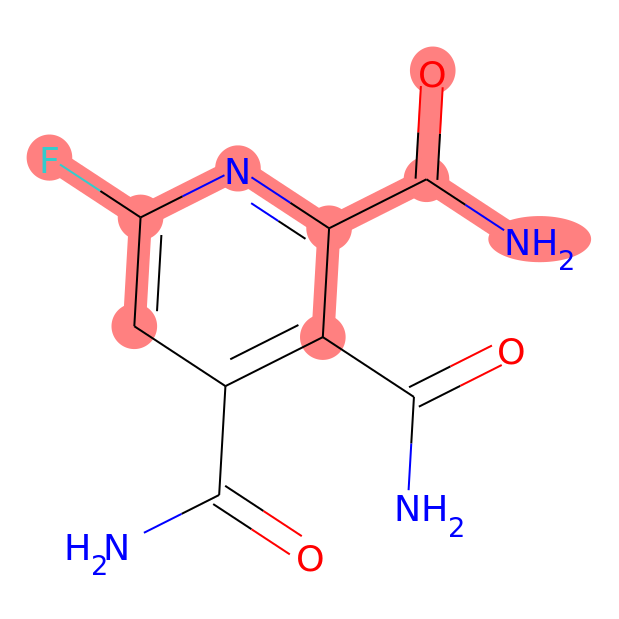}} &
  \parbox[c]{0.24\textwidth}{\includegraphics[width=0.24\textwidth]{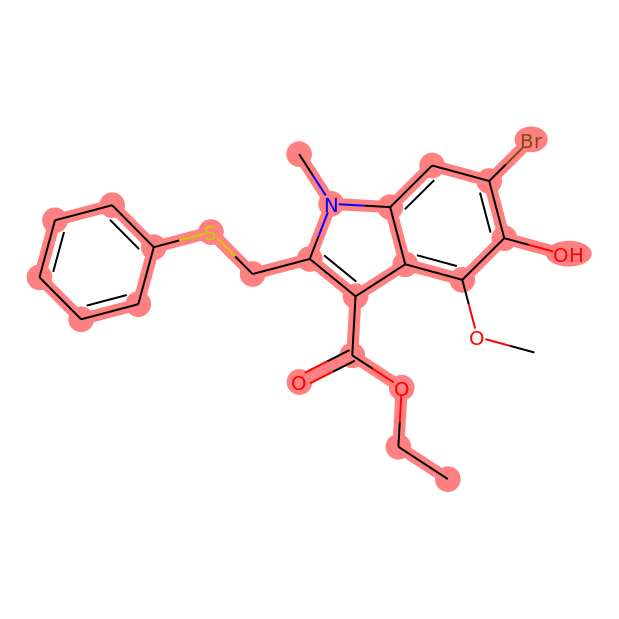}} &
  \parbox[c]{0.24\textwidth}{\includegraphics[width=0.24\textwidth]{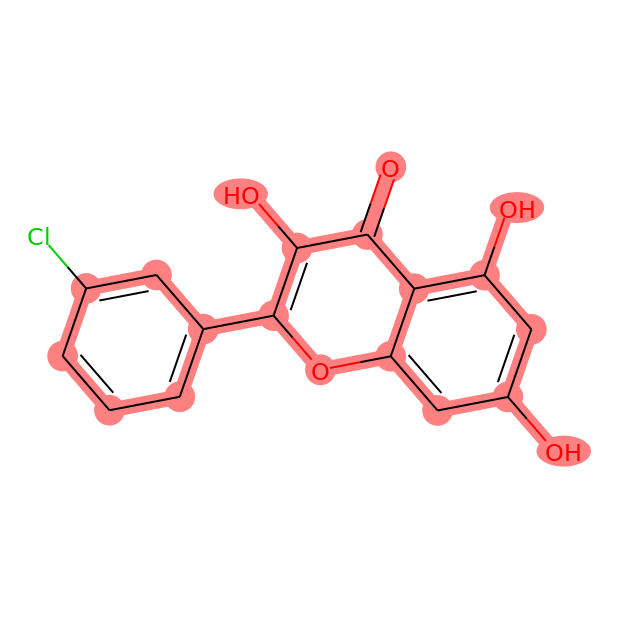}} \\
  & \multicolumn{1}{c}{$-25.65$} & \multicolumn{1}{c}{$-10.93$} & \multicolumn{1}{c}{$-20.87$} & \multicolumn{1}{c}{$-13.48$} \\
  \bottomrule
\end{tabular}
}
\caption{Rows from top to bottom (images): original molecule with common substructure highlighted, Tanimoto similarity map of improved molecule with respect to original molecule (green indicates similar regions, purple indicates dissimilar regions), QMO-optimized molecule with common substructure highlighted. Rows from top to bottom (numbers): original molecule MM/PBSA binding free energy (BFE) estimate, overall Tanimoto similarity, QMO-optimized molecule BFE estimate. Both BFE calculations are done with the molecules docked to the substrate-binding pocket of M\textsuperscript{pro}.}
\label{tab:selected_covid_targets}
\end{table}

Since all of these 23 inhibitors are reported to bind to the substrate-binding pocket of M\textsuperscript{pro}, we investigate possible binding mode alterations of the QMO-optimized molecules. It should be noted that, direct comparison of $\text{IC}_{50}$ with binding free energy is not always
possible, as the  relationship of binding affinity and $\text{IC}_{50}$ for a given compound
varies depending on the assay conditions and the compound’s
mechanism of inhibition \cite{cournia2017relative}. Further, high fidelity binding free energy estimation requires accounting for factors such as conformational entropy and explicit presence of the solvent.  Nevertheless, we report binding free energy and mode for the QMO-optimized variants. For simplicity, we limit the analysis to achiral molecules. First, we run blind docking simulations using AutoDock Vina \cite{trott2010autodock} over the entire structure of M\textsuperscript{pro} with the $\text{exhaustiveness}$ parameter set to 8. 
We further rescore top 3 docking poses for each of the original and QMO-optimized molecules using the  Molecular Mechanics/Poisson Boltzmann Surface Area (MM/PBSA)  method and AMBER forcefield \cite{wang2019farppi}, that is known to be more rigorous and accurate than the scoring  function used in docking.
Next we inspect if any of the top-3 docking poses of  the original as well as of QMO-optimized variants involves the substrate-binding pocket of M\textsuperscript{pro}, as favorable interaction with that pocket is crucial for M\textsuperscript{pro} function inhibition. 
As an illustration, Figure \ref{fig:dipyridamole} shows top docking pose of Dipyridamole and its QMO-optimized variant to the M\textsuperscript{pro} substrate-binding pocket. Consistent with more favorable MM/PBSA binding free energy, the QMO-optimized variant forms 14\% more contacts (with a 5 \AA~ distance cutoff between heavy atoms) with M\textsuperscript{pro} substrate-binding pocket compared to  Dipyridamole. Some of the M\textsuperscript{pro} residues that explicitly form contacts with the Dipyridamole variant are LEU167, ASP187, ARG188, and GLN192. Similar observations, e.g. higher number of contacts with M\textsuperscript{pro} substrate-binding pocket, which involves TYR54, were found for  other exemplars of  QMO variants, such as for Favipiravir and Umifenovir.
See Section \ref{appendix_addtl_vis} in the Supplementary Material for extended blind docking analysis.

\definecolor{mycolor1}{rgb}{0.558, 0.108, 0.478}

\begin{figure}[t]
\centering
   \begin{minipage}[t]{0.65\textwidth} \begin{subfigure}[h]{0.5\textwidth}
    \centering
    \caption{}
    \includegraphics[width=\textwidth]{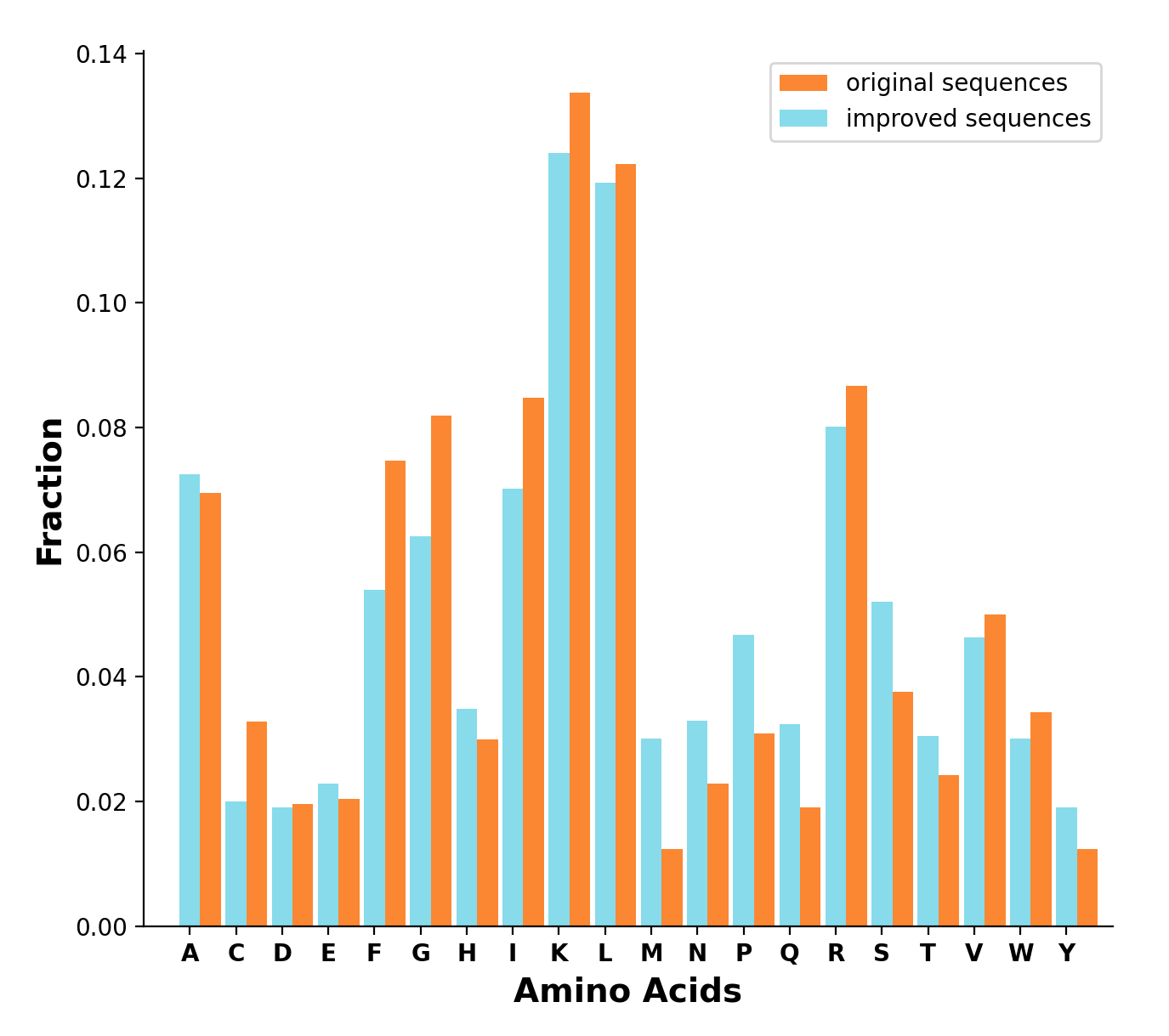}
  \end{subfigure}
  \hfill
  \hspace{-30mm}
  \begin{subfigure}[h]{0.5\textwidth}
    \centering
    \caption{}
    \includegraphics[width=\textwidth]{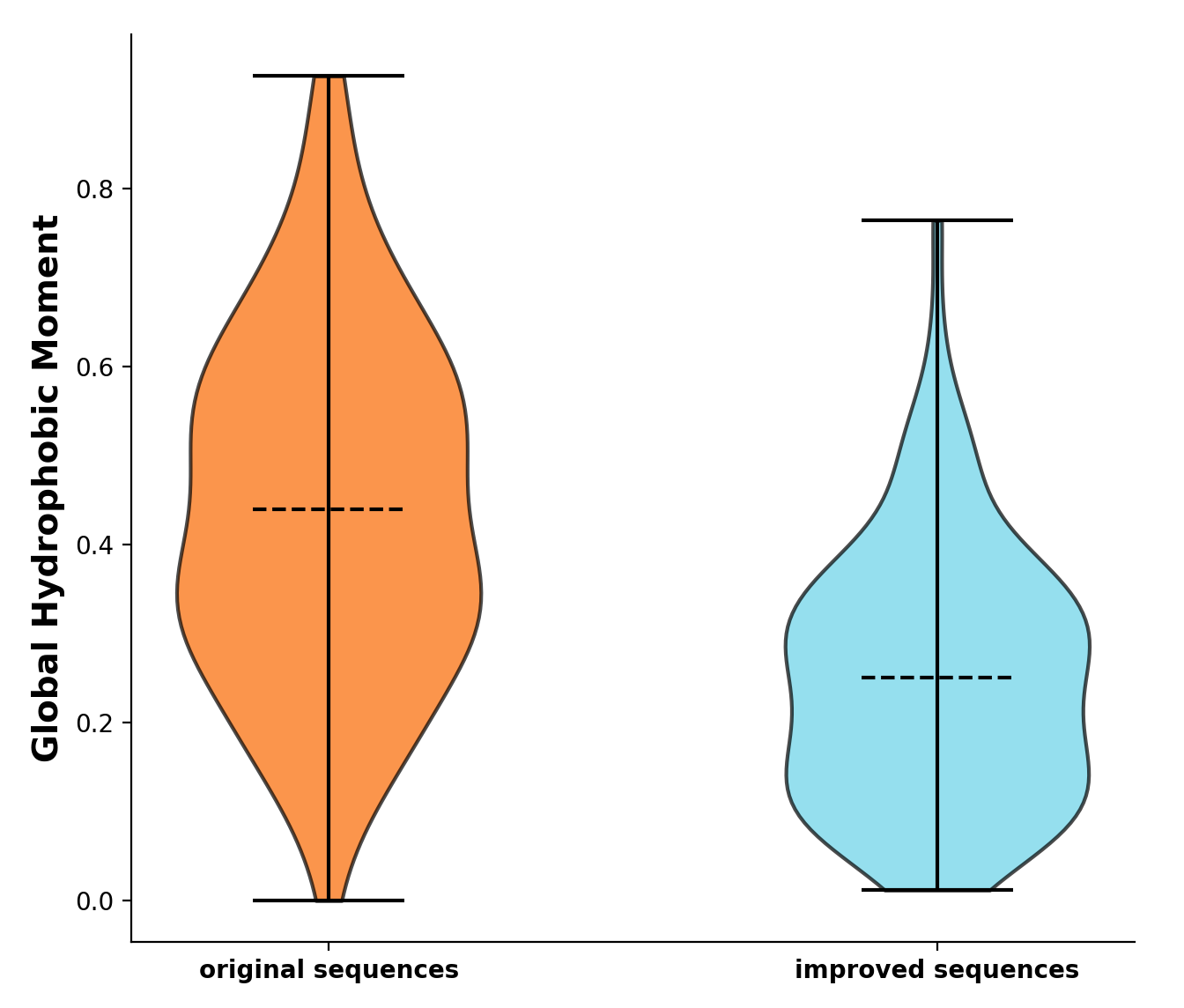}
  \end{subfigure}
  \end{minipage}
\hfill
\hspace{-5mm}
   \begin{minipage}[t]{0.35\textwidth} 
\adjustbox{max width=0.97\textwidth}{
 \begin{tabular}[h]{@{}l|cc|rr@{}}
\multicolumn{2}{l}{{\Large \textbf{(c)}} Peptide sequences and their similarity ratio} \\ 
\toprule
 Original (top) -- Optimized (bottom) &
 \multicolumn{2}{l}{Similarity Ratio}  
 
\\ \midrule

    \texttt{\textcolor{blue}{FFHH}\textcolor{red}{IFR}\textcolor{blue}{G}\textcolor{red}{IV}\textcolor{blue}{H}\textcolor{red}{V}\textcolor{blue}{A}\textcolor{red}{K}\textcolor{blue}{TIHR}\textcolor{red}{L}\textcolor{blue}{VT}-{}-\textcolor{red}{G}}  &  \multirow{1}{*}{0.5378} \\
    \texttt{\textcolor{blue}{FFHH}\textcolor{red}{VHV}\textcolor{blue}{G}\textcolor{red}{VA}\textcolor{blue}{H}\textcolor{red}{A}\textcolor{blue}{A}\textcolor{red}{H}\textcolor{blue}{TIHR}\textcolor{red}{T}\textcolor{blue}{VT}VV\textcolor{red}{T}} &   \\
    \hline
  
    \texttt{\textcolor{blue}{AKKVFKRLG}\textcolor{red}{IGAVL}\textcolor{blue}{WV}\textcolor{red}{L}\textcolor{blue}{TTG}} &  \multirow{1}{*}{0.7128} \\
    \texttt{\textcolor{blue}{AKKVFKRLG}\textcolor{red}{DAILV}\textcolor{blue}{WV}\textcolor{red}{T}\textcolor{blue}{TTG}} &    \\ 
    \hline
    
    \texttt{\textcolor{blue}{WFHHI}\textcolor{red}{FR}\textcolor{blue}{G}\textcolor{red}{IV}\textcolor{blue}{H}\textcolor{red}{V}\textcolor{blue}{G}\textcolor{red}{K}\textcolor{blue}{TIHR}\textcolor{red}{L}\textcolor{blue}{VTG}} & 
    \multirow{1}{*}{0.7460}  \\
    \texttt{\textcolor{blue}{WFHHI}\textcolor{red}{HS}\textcolor{blue}{G}\textcolor{red}{VI}\textcolor{blue}{H}\textcolor{red}{E}\textcolor{blue}{G}\textcolor{red}{S}\textcolor{blue}{TIHR}\textcolor{red}{Q}\textcolor{blue}{VTG}} &    \\
    \hline
      
    \texttt{\textcolor{blue}{F}\textcolor{red}{W}\textcolor{blue}{G}\textcolor{red}{A}\textcolor{blue}{LA}\textcolor{red}{KG}\textcolor{blue}{ALKL}I\textcolor{blue}{GS}\textcolor{red}{L}\textcolor{blue}{FS}\textcolor{red}{S}\textcolor{blue}{FSKKD}} &  \multirow{1}{*}{0.5784}  \\
    \texttt{\textcolor{blue}{F}\textcolor{red}{Y}\textcolor{blue}{G}\textcolor{red}{M}\textcolor{blue}{LA}\textcolor{red}{ML}\textcolor{blue}{ALKL}-\textcolor{blue}{GS}\textcolor{red}{V}\textcolor{blue}{FS}\textcolor{red}{K}\textcolor{blue}{FSKKD}} &    \\
    \hline
 
    \texttt{\textcolor{blue}{IGGI}\textcolor{red}{I}\textcolor{blue}{SFFK}-\textcolor{blue}{RLF}} & 
    \multirow{1}{*}{0.7332}  \\
    \texttt{\textcolor{blue}{IGGI}\textcolor{red}{S}\textcolor{blue}{SFFK}K\textcolor{blue}{RLF}} &  \\
    \hline
 
    \texttt{\textcolor{blue}{FLP}\textcolor{red}{I}\textcolor{blue}{LAGLAA}\textcolor{red}{K}\textcolor{blue}{I}\textcolor{red}{V}\textcolor{blue}{P}\textcolor{red}{KL}\textcolor{blue}{FC}\textcolor{red}{L}\textcolor{blue}{A}\textcolor{red}{T}\textcolor{blue}{KKC}} & 
    \multirow{1}{*}{0.7152}   \\
    \texttt{\textcolor{blue}{FLP}\textcolor{red}{M}\textcolor{blue}{LAGLAA}\textcolor{red}{V}\textcolor{blue}{I}\textcolor{red}{A}\textcolor{blue}{P}\textcolor{red}{AA}\textcolor{blue}{FC}\textcolor{red}{A}\textcolor{blue}{A}\textcolor{red}{A}\textcolor{blue}{KKC}}  &   \\

\bottomrule
\end{tabular}
}
   \end{minipage}
\caption{(a) Amino acid distribution comparison of original and QMO optimized 109 sequences. (b) Comparison of global hydrophobic moment. (c) Pairwise alignment and similarity ratio between few original and improved sequences. Sequences were validated by external toxicity classifier HAPPENN \cite{timmons2020happenn} and AMP activity predictor with iAMP-2L\cite{xiao2013iamp}. 
Color coding follows  \textcolor{red}{red=mismatch}, \textcolor{blue}{blue=match}, and \textcolor{black}{black=gap}. Similarity ratio is the ratio of similarity between the original and optimized sequences with respect to the  self similarity of the original sequence. Similarity is estimated by using the global alignment function in Biopython. }
\label{fig:bio_properties}
\end{figure}

\paragraph{Optimization of Existing Antimicrobial Peptides (AMPs) toward Improved Toxicity}
\label{subsec_AMP}


As an additional motivating use-case, 
discovering new antibiotics at a rapid speed is critical to tackling the looming crisis of a global increase in antimicrobial resistance \cite{coates2011novel}. AMPs are considered as promising candidates for next generation antibiotics. Optimal AMP design requires balancing between multiple, tightly interacting attribute objectives\cite{tallorin2018discovering,porto2018silico}, such as high potency and low toxicity. As an attempt toward addressing this challenge, we show how QMO can be used to find improved variants of known AMPs with reported/predicted toxicity, such that the variants have lower predicted toxicity and high sequence similarity, when compared to original AMPs. 

For the AMP optimization task, a peptide molecule is represented as a sequence of 20 natural amino acid characters. 
Using the QMO formulation in \eqref{eq_unconstrained_search}, subject to the constraints of toxicity prediction value ($f_{\text{tox}}$) and AMP prediction value ($f_{\text{AMP}}$), we aim to find most similar molecules for a set of toxic AMPs.
The sequence similarity score ($g_{\text{sim}}$) to be maximized is computed using Biopython\footnote{\url{http://www.biopython.org}}, 
which uses global alignment between two sequences (normalized by the length of the starting sequence) to evaluate the best concordance of their characters. See Methods section for detailed descriptions.
The objective of QMO is to search for improved AMP sequences by maximizing similarity while satisfying AMP activity and toxicity predictions (i.e. classified as being AMP and non-toxic based on  predictions from pre-trained deep learning models\cite{das2020accelerating}).

In our experiments,
we use QMO to optimize 150 experimentally-verified toxic AMPs collected from public databases \cite{singh2015satpdb, pirtskhalava2016dbaasp} by Das et al. \cite{das2020accelerating} as starting sequences. Note, the toxic annotation here does not depend on a specific type of toxicity, such as hemolytic toxicity.
Figure \ref{fig:cdf_AMP} shows their cumulative success rate (turning toxic AMPs to non-toxic AMPs) using QMO up to the $t$-th iteration. Within the first few iterations,
more than 60\% molecules were successfully optimized. Eventually, about $72.66\%$ (109/150) molecules can be successfully optimized. 
Analysis over all 109 original-improved pairs reveals notable physicochemical changes, e.g.   lowering of  hydrophobicity and hydrophobic moment in the QMO-optimized AMP sequences (see Figure \ref{fig:bio_properties} (a) and (b) and also Table \ref{tab:bio_properties} in the Supplementary Material). This trend  is consistent with reported positive correlation of hydrophobicity and hydrophobic moment with cytotoxicity and hemolytic activity \cite{hawrani2008origin, sun2019characterization}.
Figure \ref{fig:bio_properties} (c) shows  examples of known AMPs and their QMO optimized variant sequences.  Sequence alignment and similarity ratio relative to the original sequence are also shown, indicating that sequences resulting from QMO differ widely from the initial ones.
Figure \ref{fig:dist_impr} in the Supplementary Material depicts the optimization process of some AMP sequences. QMO can further improve similarity while maintaining low predicted toxicity and high AMP values for the specified thresholds after first success.

\begin{figure}[t]
\centering
  \begin{subfigure}[t]{0.48\textwidth}
    \centering
    \includegraphics[width=\textwidth]{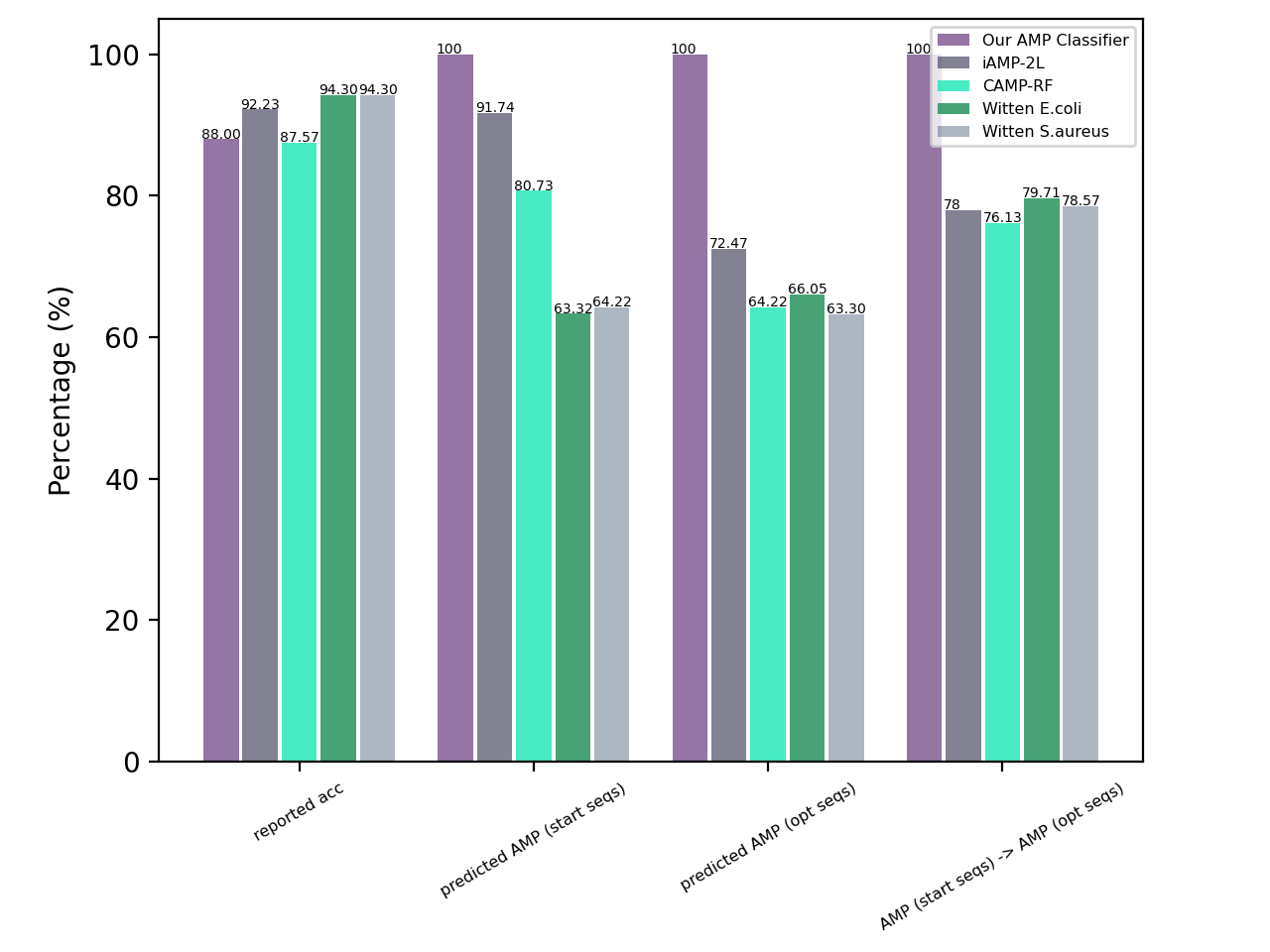}
    \caption{AMP}
  \end{subfigure}
      \hspace{-40cm}
  \hfill
  \begin{subfigure}[t]{0.48\textwidth}  
    \centering 
    \includegraphics[width=\textwidth]{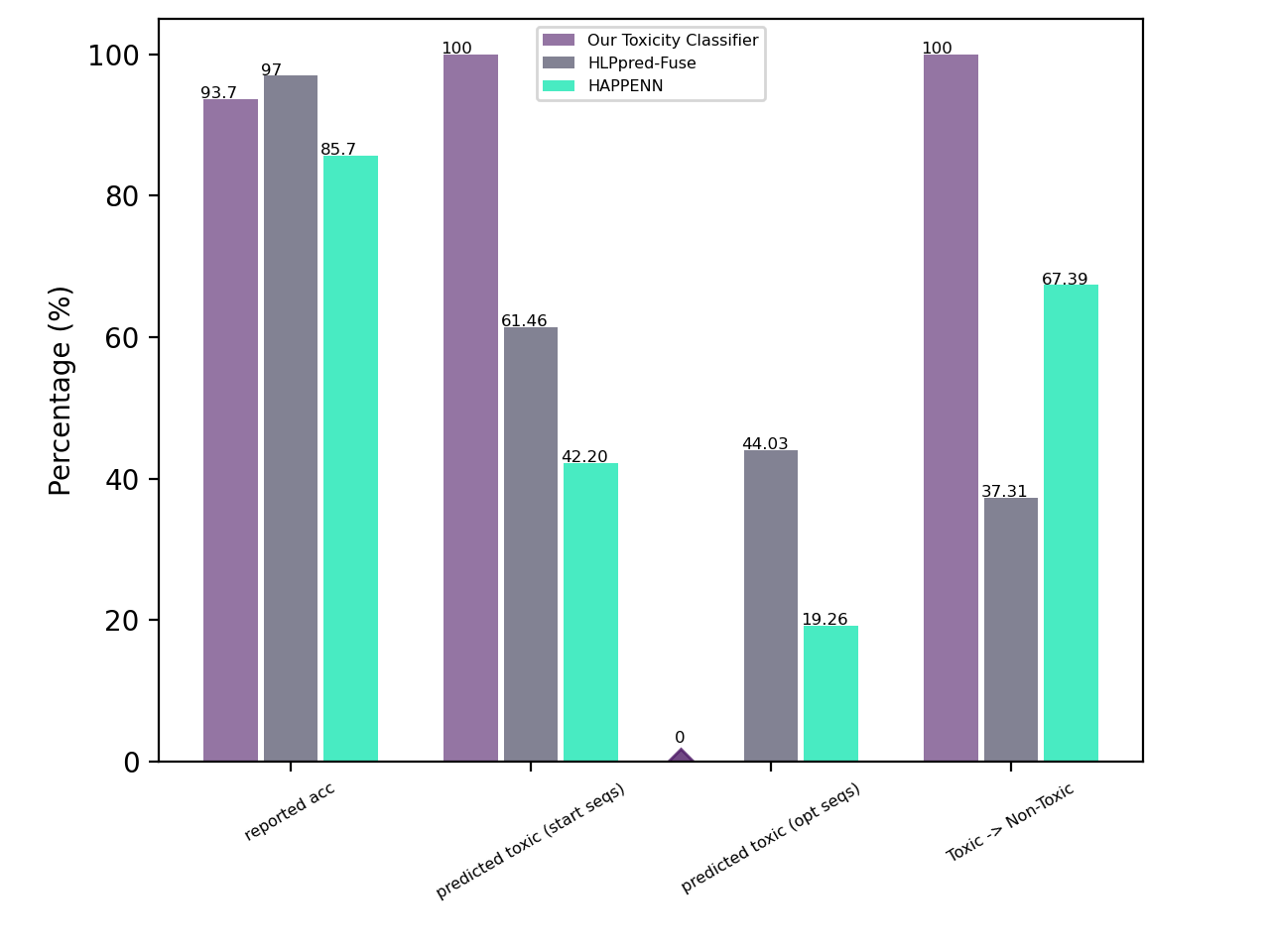}
    \caption{Toxicity}
  \end{subfigure}
  \caption{Reported accuracy, prediction rate, and property improvement for 109 pairs of starting and QMO-optimized sequences 
    based on different AMP and Toxicity classifiers. The 109 starting sequences are experimentally verified toxic AMPs and are correctly predicted by the AMP and toxicity classifiers used in QMO.  
    The external classifiers have varying prediction accuracy as they may yield incorrect predictions  on some of starting sequences. The prediction rate on QMO-optimized sequences is defined as the fraction of AMP and/or toxin predictions.
    About 56.88\% of QMO-optimized sequences are predicted as non-toxic AMPs by iAMP-2L + HAPPENN, showing high agreement with the classifiers used in QMO. The complete results are reported in Table \ref{tab:ext-classifiers} of the Supplementary Material.} 
  \label{fig:AMP}
\end{figure}

We perform additional validation of our optimization results by comparing QMO-optimized sequences using a number of state-of-the-art AMP and toxicity predictors that are external classifiers not used in the QMO framework. Figure \ref{fig:AMP} shows the external classifiers' prediction results on 109  original and improved sequence pairs that are successfully optimized by QMO. 
We note that these external classifiers vary on training data size and type, as well as on model architecture, and report a range of accuracy. Data and models for the toxicity prediction task are more rare, compared to those for the AMP classification problem. Further, external toxicity classifiers such as HAPPENN \cite{timmons2020happenn} and HLPpred-fuse \cite{hasan2020hlppred} target explicitly predicting hemolytic nature.
For these reasons, the predictions of the external classifiers on the original lead sequences may vary, when compared to ground-truth labels (see the third column in Table \ref{tab:ext-classifiers} of Supplementary Material).
Nonetheless,  predictions on the QMO-optimized sequences using external classifiers show high consistency in terms of toxicity improvement, when compared with the predictors used in QMO.
Specifically, the predictions from iAMP-2L \cite{xiao2013iamp} and  HAPPENN \cite{timmons2020happenn} (hemolytic toxicity prediction) show that 56.88\% (62/109) QMO-optimized molecules are predicted as non-toxic AMPs. In Section \ref{sup_encoder} of the Supplementary Material, we also show that the use of a better encoder-decoder helps the optimization performance of QMO.

\begin{figure}[tbh]
\centering
\includegraphics[width=\textwidth]{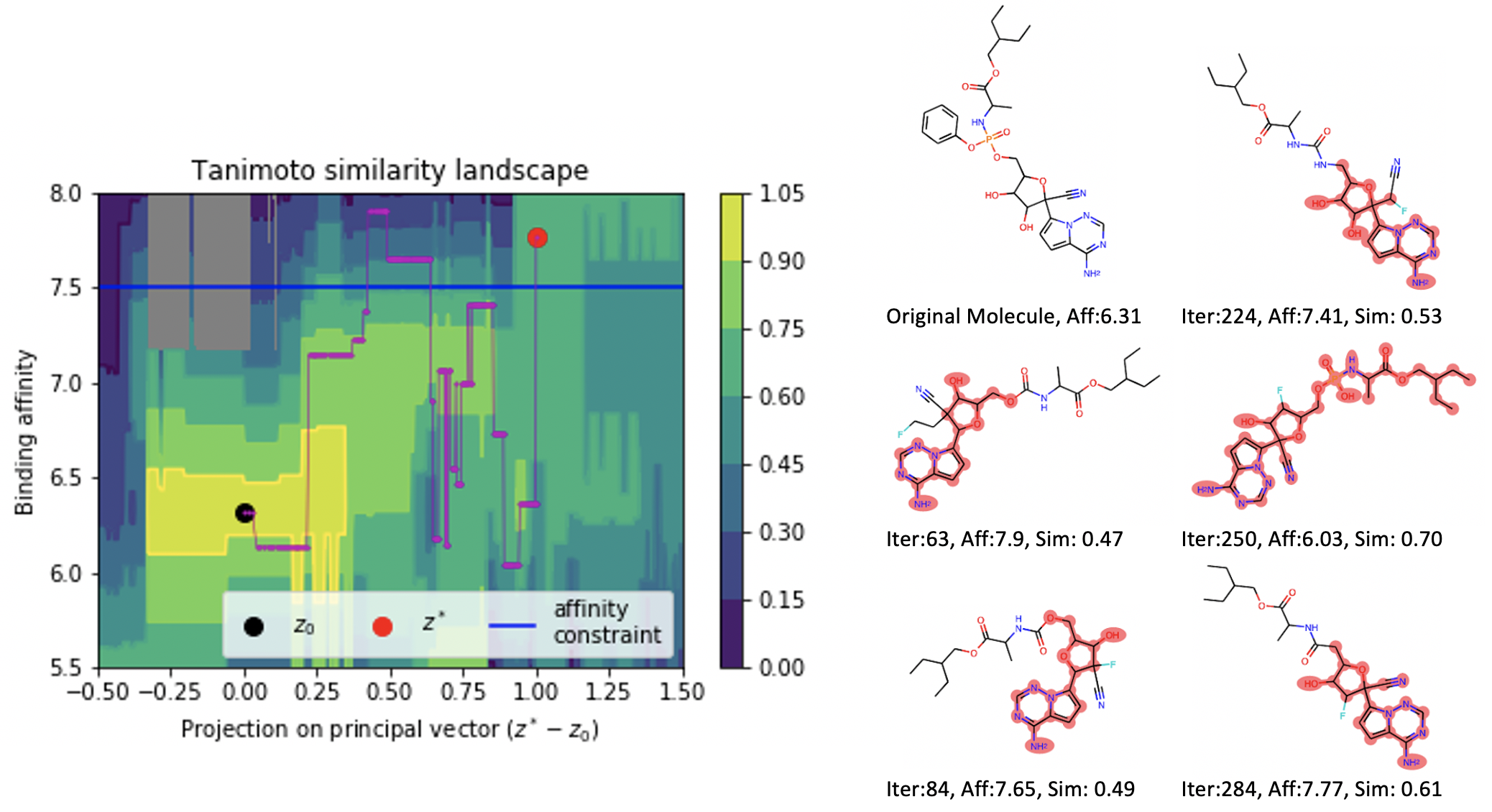}
  \caption{\textbf{Left:} QMO trajectory visualization on the landscape of Tanimoto similarity v.s. binding affinity prediction when using Remdesivir as the lead molecule. The optimization objective is to maximize Tanimoto similarity while ensuring the predicted binding affinity is above a defined threshold of 7.5. The gray area indicates infeasible region according to local grid search.
  \textbf{Right:} Common substructures over the QMO optimization process with respect to the Remdesivir structure and their property predictions. Iter denotes iteration index in QMO, Aff denotes affinity, and Sim denotes Tanimoto similarity. The highlighted part in red color indicates subgraph similarity.}
  \label{fig:QMO_loss_landscape}
\end{figure}

\begin{figure}[t]
\centering
\includegraphics[width=0.8\textwidth]{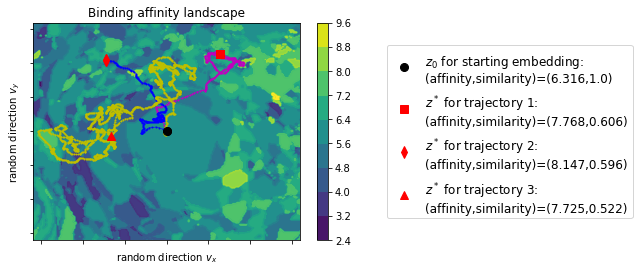}
  \caption{Three sets of QMO trajectory visualization on the landscape of predicted binding affinity when using Remdesivir as the lead molecule. The trajectory differs by the rand seed used in QMO for query-based guided search with random samples.
  The optimization objective is to maximize Tanimoto similarity while ensuring the predicted binding affinity is above a defined threshold of 7.5. The visualization suggests that QMO can find a diverse set of improved molecules. }
  \label{fig:QMO_diversity}
\end{figure}

\paragraph{Property Landscape Visualization and Trajectory Analysis}

To gain better understanding of how QMO optimizes a lead molecule with respect to the property constraints and objectives, we provide visual illustration of the property landscapes and search trajectories via QMO using a two-dimensional local interpolation on the molecule embedding space. Specifically, given the original embedding $z_0$ and the embedding of the best candidate $z^*$ returned by QMO, we perform local grid sampling following two selected directions $v_x$ and $v_y$, and then evaluate the properties of the decoded sequences from the sampled embeddings for property landscape analysis. For the purpose of visualizing the property landscape in low dimensions, we project the high-dimensional search trajectories $\{z_t\}_{t=1}^T$ to the two directions $v_x$ and $v_y$. Figure \ref{fig:QMO_loss_landscape} shows the landscape of Tanimoto similarity v.s. binding affinity prediction when using Remdesivir as the lead molecule, with the optimization objective of maximizing Tanimoto similarity while ensuring the predicted binding affinity is above a defined threshold 7.5. The two directions are the principal vector $z^*-z_0$ and a random vector orthogonal to the principal vector (see Methods section for more details). The trajectory shows how QMO leverages the evaluations of similarity and binding affinity for optimizing the lead molecule.
Figure \ref{fig:QMO_loss_landscape} also displays the common substructure of candidate molecules in comparison to the Remdesivir molecule in terms of subgraph similarity and their predicted properties over sampled iterations in QMO. 


In addition to demonstrating the efficiency in optimizing lead molecules, we also study the diversity of the optimized molecules by varying the random seed used in QMO for query-based guided search. Figure \ref{fig:QMO_diversity} shows three different sets of trajectory on the landscape of predicted binding affinity when using Remdesivir as the lead molecule  (see Methods section for more details). 
  The optimization objective is the same as that of Figure \ref{fig:QMO_loss_landscape}. The visualization suggests that the trajectories are distinct and the best candidate molecule in each trajectory is distant from each other in the embedding space, suggesting that QMO can find a diverse set of improved molecules with desired properties. We also provide a quantitative study on the diversity and novelty of the QMO-optimized sequences when varying the similarity threshold  in Section \ref{sup_diversity} of the Supplementary Material. Setting a lower similarity threshold in QMO results in more novel and diverse sequences.

\section*{Discussion and Conclusion}

In this paper we propose QMO, a generic-purpose molecule optimization framework that readily applies to any pre-trained molecule encoder-decoder with continuous latent molecule embeddings and any set of property predictions and evaluation metrics. It features efficient guided search with molecular property evaluations and constraints obtained using predictive models and cheminformatics softwares.  More broadly, QMO is a machine learning gear that can be incorporated into different scientific discovery pipelines with deep generative models, such as generative adversarial networks, for efficient guided optimization with constraints. As a demonstration, Section \ref{sup_objective} and \ref{sup_rand_z} of the Supplementary Material show the QMO results on the SARS-CoV-2 Main Protease inhibitor optimization task with alternative objectives and randomly generated lead sequences, respectively. QMO is able to
perform successful optimization with respect to different objectives, constraints, and starting sequences.
The proposed QMO framework can be applied in principle to other classes of materials, for example metal oxides, alloys, and genes.

On the simpler benchmark tasks for optimizing drug-likeness and penalized logP scores with similarity constraints, QMO demonstrates superior performance over baseline results. We also apply QMO to improve the binding affinity of existing inhibitors of the SARS-CoV-2 Main Protease, and to improve the toxicity of antimicrobial peptides. The QMO-optimized variants of existing drug molecules show favorable binding free energy with SARS-CoV-2 Main Protease  upon blind docking and MM/PBSA rescoring, whereas the QMO-optimized peptides are consistently predicted to be antimicrobial and non-toxic  by  external peptide property predictors. The property landscape analysis and low-dimensional visualization of the optimization trajectories provide insights on how QMO efficiently navigates in the property space to find a diverse set of improved molecules with the desired properties.
Our results show strong evidence for QMO to serve as a novel and practical tool for molecule optimization and other process/product design problems as well to aid accelerating chemical discovery with constraints. In Section \ref{appen_ablation} of the Supplementary Material, we provide an ablation study of QMO for additional performance analysis, including the effect of encoder-decoder, the difference between sequence-level and latent-space classifiers, and the comparison between different gradient-free optimizers. The results show that QMO is a query-efficient end-to-end molecule optimization framework, and a better encoder-decoder can further improve its performance.
Future work will include integrating multi-fidelity expert feedback into the QMO framework for human-AI collaborative material optimization, and using QMO for accelerating discovery of novel, high-performance and low-cost materials.



\section*{Methods}
\label{sec_methods}


\subsection*{Procedure Descriptions of the QMO Framework}
\label{method_QMO}


\begin{itemize}
    \item \textbf{Procedure Inputs:} Pre-trained encoder-decoder; Molecular property predictors $\{f_i\}_{i=1}^I$ and thresholds $\{\tau_i\}_{i=1}^I$; Molecular similarity metrics $\{g_j\}_{j=1}^J$ and thresholds $\{\eta_j\}_{j=1}^J$; Total search iteration $T$; Step size $\{\alpha_t\}_{t=0}^{T-1}$; Starting lead molecule sequence $x_0$; Reference sequence set $\cS$; \loss~function from \eqref{eq_unconstrained_search} or \eqref{eq_unconstrained_search_2}
    \item \textbf{Procedure Initialization:} $z^{(0)}=\enc(x_0)$; $\cZ_{\text{solution}}\leftarrow \{\varnothing \}$
    \item Repeat the following steps for $T$ times, starting from $T=0$:
    \item \textbf{Gradient Estimation:}  Generate $Q$ random unit-norm perturbations $\{ u^{(q)}\}_{q=1}^Q$ and compute \\
    $\nablahat \loss(z^{(t)}) = \frac{d}{\beta \cdot Q} \sum_{q=1}^Q \left[ \loss(z^{(t)}+\beta u^{(q)}) - \loss(z^{(t)}) \right]    \cdot u^{(q)}$
    \item \textbf{Pseudo Gradient Descent:}  $z^{(t+1)} = z^{(t)} - \alpha_t \cdot \nablahat \loss(z^{(t)})$
    \item \textbf{Molecular property and constraint verification:} If solving for formulation \eqref{eq_unconstrained_search}, check $f_i(\dec(z^{(t)})) \geq \tau_i$ for all $i \in [I]$. If solving for formulation \eqref{eq_unconstrained_search_2}, check $g_j(\dec(z^{(t)})|\cS) \geq \eta_j$ for all $j \in [J]$. 
    \item \textbf{Update valid molecule sequence:} $\cZ_{\text{solution}} \leftarrow \cZ_{\text{solution}} \cup \{z^{(t)}\} $
\end{itemize}

\subsection*{Procedure Convergence Guarantee and Implementation Details for QMO}
Inherited from zero order optimization, QMO has algorithmic convergence guarantees. Under mild conditions on the true gradient (Lipschitz continuous and bounded gradient), the zeroth order gradient descent following \eqref{eqn_ZO_descent} ensures QMO takes at most $O(\frac{d}{T})$ iterations to be sufficiently close to a local optimum in the loss landscape for a non-convex objective function \cite{ghadimi2013stochastic,liu2020primer}, where $T$ is the number of iterations. In addition to the standard zeroth order gradient descent method, our QMO algorithm can naturally adopt different zeroth order solvers, such as zeroth order stochastic and accelerated gradient descent. Our implementation of gradient estimation gives $Q+1$ loss function queries per iteration. If the decoder outputs a SMILES string, we pass the string to RDKit for validity verification and disregard invalid strings.

In our QMO implementation, we use the zeroth-order version of the popular Adam optimizer \cite{kingma2014adam} that automatically adjusts the step sizes $\{\alpha_t\}_{t=1}^{T-1}$ with an initial learning rate $\alpha_0$ (see Section \ref{apprndix_algo} in the Supplementary Material for more details).  Empirically, we find that Adam performs better than stochastic gradient descent (SGD) in our tasks. The convergence of zeroth order Adam-type optimizer is given in Chen et al.\cite{chen2019zo}. 
We will specify experimental settings, data descriptions, and QMO hyperparameters for each task.
In all settings, QMO hyperparameters were tuned to a narrow range and then all the reported combinations were tried for each starting sequence. Among all feasible solutions returned by QMO, we report the one having the best molecular score given the required constraints.
The stability analysis of QMO is studied in Section \ref{appendix_stability} of the Supplementary Material.

\subsection*{Machine Learning  Experimental Settings}
In our experiments, we run the QMO procedure based on the reported hyperparmeter values and report the results of the best molecule found in the search process. The procedure will return null (that is, an unsuccessful search) if the it fails to find a valid molecule sequence.

\paragraph{Benchmarks on QED and penalized logP}
The pre-trained encoder-decoder by Winter et al. \cite{winter2019learning} is used, with the latent dimension $d=512$. For the penalized logP optimization task, we use $Q=100$, $\beta=10$, $\alpha_0=2.5$, $\gamma_{\text{penalized logP}}=0.04$, and $T=80$. 
For the QED task, we use $Q=50$, $\beta=10$, $\alpha_0=0.05$, $\gamma_{\text{QED}}=4$, $T=20$, and report the best results among $50$ restarts. We find that for the QED task, using multiple restarts can further improve the performance (see Section \ref{appendix_stability} in the Supplementary Material for detailed discussion). For penalized logP, there is no reason to continue optimizing past 80 iterations as penalized logP can be increased almost arbitrarily without making the resulting molecule more useful for drug discovery \cite{zhou2019optimization} --- even under similarity constraints, as we find. Therefore, we set $T=80$ for the  penalized logP task.

\paragraph{Optimizing Existing Inhibitor Molecules for SARS-CoV-2 Main Protease}
The pre-trained encoder-decoder by Winter et al. \cite{winter2019learning} is used, with the latent dimension $d=512$. The hyperparameters of QMO are $Q=10$, $T=2000$, $\beta=\{10, 25\}$, $\alpha_0=\{0.1, 0.05\}$, and $\lambda_{\text{Tanimoto}}=\{1, 10\}$. 

\paragraph{Optimization of AMPs for Improved Toxicity}

The pre-trained predictors for toxicity and AMP by Das et al. \cite{das2020accelerating} are used, with the latent dimension $d=100$. 
The similarity between the original sequence $x_0$ and the improved sequence $x$ is computed using the global alignment function in Biopython, formally defined as $g_{\text{sim}}(x|x_0) = \textsf{global-alignment}(x,x_0) / \log (\textsf{length}(x_0))$,
where $\textsf{global-alignment}(x,x_0)$ is the value returned by the function pairwise2.align.globalds($x$, $x_0$, matlist.blosum62, -10, -1) and $\log ( \textsf{length}(x_0) )$ is the log value of $x_0$'s sequence length. Blosum62 is the weight matrix for estimating alignment score\cite{henikoff1992amino}, and -10/-1 is the penalty for  opening/continuing a gap.
The QMO parameters are $Q=100$, $\beta=\{1, 10\}$, $\alpha_0=\{0.1, 0.05, 0.01\}$, $\lambda_{\text{sim}}=0.01$, and $T=5000$. The toxicity property constraint is  set as $f_{\text{tox}}(x) \leq 0.1529$ and amp as $f_{\text{amp}}(x) \geq 0.9998$. Binary classification on this threshold gives 93.7\% accuracy for toxicity and 88.00\% for AMP prediction on a large peptide database \cite{das2020accelerating}.

\subsection*{Trajectory Visualization}
In Figure \ref{fig:QMO_loss_landscape} and Figure \ref{fig:QMO_diversity}, the optimization  trajectory by QMO is visualized by projection on two selected directors $v_x$ and $v_y$ originated from the starting embedding $z_0$. Specifically, in \eqref{fig:QMO_loss_landscape} we set $v_x = z^* - z_0$ and set $v_y$ as a unit-norm randomly generated vector that is orthogonal to $v_x$. The two-dimensional local grid in the embedding space are then sampled according to $z_{\text{grid}}(x,y)=z_0+x\cdot v_x+y\cdot \|z^*\|_2 \cdot u_y$, where $\|\cdot\|_2$ denotes the Euclidean distance and we sample $x$ and $y$ uniformly from $[-0.5,1.5]$ and $[-2,2]$, respectively. Note that by construction, $z_{\text{grid}}(0,0)=z_0$ and $z_{\text{grid}}(1,0)=z^*$. Then, we evaluate the Tanimoto similarity and binding affinity prediction of the grid and present their results in Figure \ref{fig:QMO_loss_landscape}. 
Similarly, in Figure \ref{fig:QMO_diversity} we set $v_x$ and $v_y$ to be two unit-norm randomly generated vectors, and set $z_{\text{grid}}(x,y)=z_0+x\cdot \|z_0\|_2 \cdot v_x+y\cdot \|z_0\|_2 \cdot u_y$, where $x$ and $y$ are sampled uniformly from $[-1.6,1.6]$.

\subsection*{Data and Reproducible Codes}
Data and codes for the benchmark molecule optimization tasks (QED and penalized logP) are available at \textcolor{blue}{\url{https://github.com/IBM/QMO}}\cite{samuel_hoffman_2021_5562908}. For other inquiries,
please contact Pin-Yu Chen <pin-yu.chen@ibm.com> and Payel Das <daspa@us.ibm.com>.





\section*{Author contributions statement}

S.\ C.\ Hoffman and V.\ Chenthamarakshan contributed to the SARS-CoV-2, QED, and penalized logP optimization experiments. K.\ Wadhawan contributed to the AMP experiment. S.\ C.\ Hoffman and P.\ Das contributed to the docking simulations. P.-Y.\ Chen contributed to QMO methodology design and property landscape analysis. V.\ Chenthamarakshan contributed to common substructure analysis.
All authors conceived and designed research, analyzed results, and contributed to paper writing. 

\section*{Materials and Correspondence} Please contact Dr. Pin-Yu Chen  (pin-yu.chen@ibm.com) and Dr. Payel Das (daspa@us.ibm.com)

\section*{Acknowledgements} 
The authors thank Prof. Tingjun Hou and Zhe Wang, for the help with binding free energy calculation using the FarPPI webserver. The authors also thank Bhanushee Sharma for her help in improving the presentation of the system plot (Figure 1).

\bibliography{guided_search_ref}






\clearpage
\section*{Supplementary Material}



\section{Background on Zeroth Order Optimization}
\label{sec_ZO_opt}

In contrast to first-order (i.e. gradient-based) optimization, zeroth-order (ZO) optimization uses function values evaluated at queried data points to approximate the gradient and perform gradient descent, which we call \textit{pseudo gradient descent} \cite{ghadimi2013stochastic}. It has been widely used in machine learning tasks when the function values are observable, while the gradient and other higher-order information are either infeasible to obtain or difficult to compute. A recent appealing application of ZO optimization
is on efficient generation of prediction-evasive adversarial examples from information-limited machine learning models,
known as \textit{black-box adversarial attacks}
\cite{CPY17zoo}.
For the purpose of evaluating
practical robustness,
the target model only provides model predictions (e.g., a prediction API) to an attacking algorithm  and does not reveal other information. 

A major benefit of ZO optimization is its adaptivity from gradient-based methods. 
Despite using gradient estimates, many ZO optimization algorithms enjoy the same iteration complexity to converge to a stationary solution as their first-order counterparts under similar conditions\cite{liu2020primer}. However, an additional multiplicative cost in a polynomial order of the problem dimension $d$ (usually $O(d)$ or $O(\sqrt{d})$) will appear in the rate of convergence, due to the nature of query-driven pseudo gradient descent. 


\section{Additional Algorithm Descriptions}
\label{apprndix_algo}

Algorithm \ref{algo_QMO_adam} describes the zeroth order version of the Adam optimizer \cite{kingma2014adam} used in our QMO implementation.

\begin{algorithm}[h]
\caption{Query-based Molecule Optimization --- Adam Variant (QMO-Adam)}
\label{algo_QMO_adam}
\begin{algorithmic}[1]
\State {\bfseries Input:} Trained Encoder-Decoder; Molecular property predictors $\{f_i\}_{i=1}^I$ and thresholds $\{\tau_i\}_{i=1}^I$; Molecular similarity metrics $\{g_j\}_{j=1}^J$ and thresholds $\{\eta_j\}_{j=1}^J$; Total search iteration $T$; Initial step size $\alpha_0$; Starting molecule sequence $x_0$; Reference sequence set $\cS$; \loss~function
\State {\bfseries Initialization:} $z^{(0)}=\enc(x_0)$; $\cZ_{\text{solution}}\leftarrow \{\varnothing \}$; $m^{(0)}, v^{(0)} \leftarrow 0$; $\mathrm{B}_1 \leftarrow 0.9$; $\mathrm{B}_2 \leftarrow 0.999$
\For{$t=0,1, \ldots,T-1$}
 \State \textcolor{blue}{\textsf{Gradient estimation:}}   $\nablahat \loss(z^{(t)}) = \frac{d}{\beta \cdot Q} \sum_{q=1}^Q \left[ \loss(z^{(t)}+\beta u^{(q)}) - \loss(z^{(t)}) \right]    \cdot u^{(q)}$
  \State \textcolor{blue}{\textsf{Adam update:}} 
  \State
  $m^{(t+1)} = \mathrm{B}_1 \cdot m^{(t)} + (1 - \mathrm{B}_1) \cdot \nablahat \loss(z^{(t)})$
  \State $v^{(t+1)} = \mathrm{B}_2 \cdot v^{(t)} + (1 - \mathrm{B}_2) \cdot \nablahat \loss(z^{(t)})^2$
  \State $\hat{m}^{(t+1)} = m^{(t+1)}/(1 - (\mathrm{B}_1)^{t+1})$
  \State $\hat{v}^{(t+1)} = v^{(t+1)}/(1 - (\mathrm{B}_2)^{t+1})$
  \State $z^{(t+1)} = z^{(t)} - \alpha_0 \cdot \frac{\hat{m}^{(t+1)}}{\sqrt{\hat{v}^{(t+1)}} + \epsilon}$
  \State \textcolor{blue}{\textsf{Molecular property and constraint verification}:}
 \If{solving \eqref{eq_unconstrained_search},  check
 {$f_i(\dec(z^{(t)})) \geq \tau_i$ for all $i \in [I]$  }}
 \State \textcolor{blue}{\textsf{Save valid molecule sequence:}} $\cZ_{\text{solution}} \leftarrow \cZ_{\text{solution}} \cup \{z^{(t)}\} $
 \EndIf
 
 \If{solving \eqref{eq_unconstrained_search_2}, check  {$g_j(\dec(z^{(t)})|\cS) \geq \eta_j$ for all $j \in [J]$  }}
 \State \textcolor{blue}{\textsf{Save valid molecule sequence:}} $\cZ_{\text{solution}} \leftarrow \cZ_{\text{solution}} \cup \{z^{(t)}\} $
 \EndIf
\EndFor
\Return $\cZ_{\text{solution}}$
\end{algorithmic}
\end{algorithm}

\clearpage
\section{More details on Model and Dataset}
\label{appendix_model_data}

\subsection{Pre-trained Encoder-Decoder Models}
\label{sup_pre-trained}

\paragraph{QED, Penalized logP, and SARS-CoV-2 Use Cases}
We used the continuous and data-driven descriptors (CDDD) model\cite{winter2019learning} to embed the SMILES representation of a molecule into a continuous latent space. The model consists of an encoder and decoder, each with three stacked gated recurrent unit (GRU) cells and a fully-connected layer with  hyperbolic tangent activation function. The latent representation value of each dimension is thus constrained within the range of $[-1,1]$. In our QMO implementation, we perform projection of the candidate embedding $z^{(t)}$ onto $[-1,1]$ for each iteration. The training of CDDD is also regularized by molecular property predictors.  The complete training of the CDDD model involves minimizing the reconstruction loss and property prediction loss. The decoder uses a left-to-right beam search for inference. The details of the datasets, models, and hyperparameters used to train the CDDD model are described in the supplementary material of Winter et al\cite{winter2019learning}.

\paragraph{Antimicrobial Peptide Use Case}
We used the pre-trained Wasserstein autoencoder (WAE) from Das et al \cite{das2020accelerating}.
In order to learn meaningful continuous representations from sequences in unsupervised fashion, variational autoencoder (VAE) family has been proven to be a principled approach \cite{kingma2013auto}. However, vanilla VAE models are prone to mode collapse \cite{bowman2015generating}. Advanced models such as $\beta$-VAE \cite{higgins2016beta} and WAE \cite{tolstikhin2017wasserstein} were proposed to address this issue, and 
WAE was found to encode the AMP space better\cite{das2020accelerating}.

The default WAE architecture involved bidirectional-GRU encoder and GRU decoder, where GRU stands for the gated recurrent unit. The encoder is a bi-directional GRU with hidden state size of 80. The latent capacity was set to be 100. We used the same parameter setting as in Das et al \cite{das2020accelerating}. We used the random feature approximation of the Gaussian kernel with kernel bandwidth  $\sigma=7$ as reported to be performing the best. The inclusion of latent space noise log variance regularization, denoted as $R(logVar)$, helped avoiding collapse to a deterministic encoder. Among different regularization weights used, $1e-3$ had the most desirable behavior on the reported performance metrics\cite{das2020accelerating}. For more details refer Das et al \cite{das2020accelerating} and Table  \ref{tab:vae_model_metrics} of the Supplementary Material. QMO performance comparison of different model variants based on $\beta-VAE$ and WAE models are discussed in Sections \ref{sup_encoder} and \ref{sup_latent_dim}.

\subsection{Datasets}

\paragraph{SARS-CoV-2 Use Case}

In Table \ref{tab:IC50}, we see experimental $\text{IC}_{50}$ values for a few of the COVID-19 inhibitor candidates examined in Table \ref{table_covid_targets}
compared with their predicted affinity values. These are provided for reference.

\begin{table}[h]
\centering
\caption{Experimental $\text{IC}_{50}$ values copied from literature\cite{jin2020structure,jeon2020identification} and converted to $\text{pIC}_{50}$ for comparison. Affinity predictions are $\text{pIC}_{50}=-\log_{10} (\text{IC}_{50})$. The results show that the prediction is close to the experimental value.}
\label{tab:IC50}
\begin{tabular}{@{}l|rr|r@{}}
\toprule
                   & \multicolumn{2}{c|}{Experimental} & \multicolumn{1}{c}{Predicted} \\
compound           & $\text{IC}_{50}$ (\si{\micro \meter}) & $\text{pIC}_{50}$ & $\text{pIC}_{50}$ \\ \midrule
Tideglusib         & 1.55   & 5.81 & 4.94     \\
Chloroquine        & 7.28   & 5.14 & 6.07     \\
Lopinavir          & 9.12   & 5.04 & 6.42     \\
Disulfiram         & 9.35   & 5.03 & 5.54     \\
Remdesivir         & 11.41  & 4.94 & 6.32     \\
Shikonin           & 15.75  & 4.80 & 5.11     \\
PX-12              & 21.39  & 4.67 & 5.01     \\
Cinanserin         & 124.93 & 3.90 & 4.43     \\
\bottomrule
\end{tabular}
\end{table}

\paragraph{Antimicrobial Peptide Use Case}

In our experiments, we used the labeled part of a large curated antimicrobial peptide (AMP) database in a recent AI-empowered antimicrobial discovery study \cite{das2020accelerating}. 
The AMP dataset has several attributes associated with peptides from which we used antimicrobial (AMP) and toxicity. The labeled dataset has only linear and monomeric sequences with no terminal modification and length up to 50 amino acids. The dataset contains  8683 AMP and 6536 non-AMP; and 3149 toxic and 16280 non-toxic sequences. For the starting AMP sequences, we consider sequences with up to length 15 and with property being both AMP and toxic.
We then filter sequences for which our toxicity classifier predictions match with ground truth and obtain 167 sequences. 


\section{Additional Data Analysis and Visualization}
\label{appendix_addtl_vis}

\subsection{Penalized log P Score Optimization}
\begin{figure}[h]
    \centering
    \includegraphics[width=0.6\textwidth]{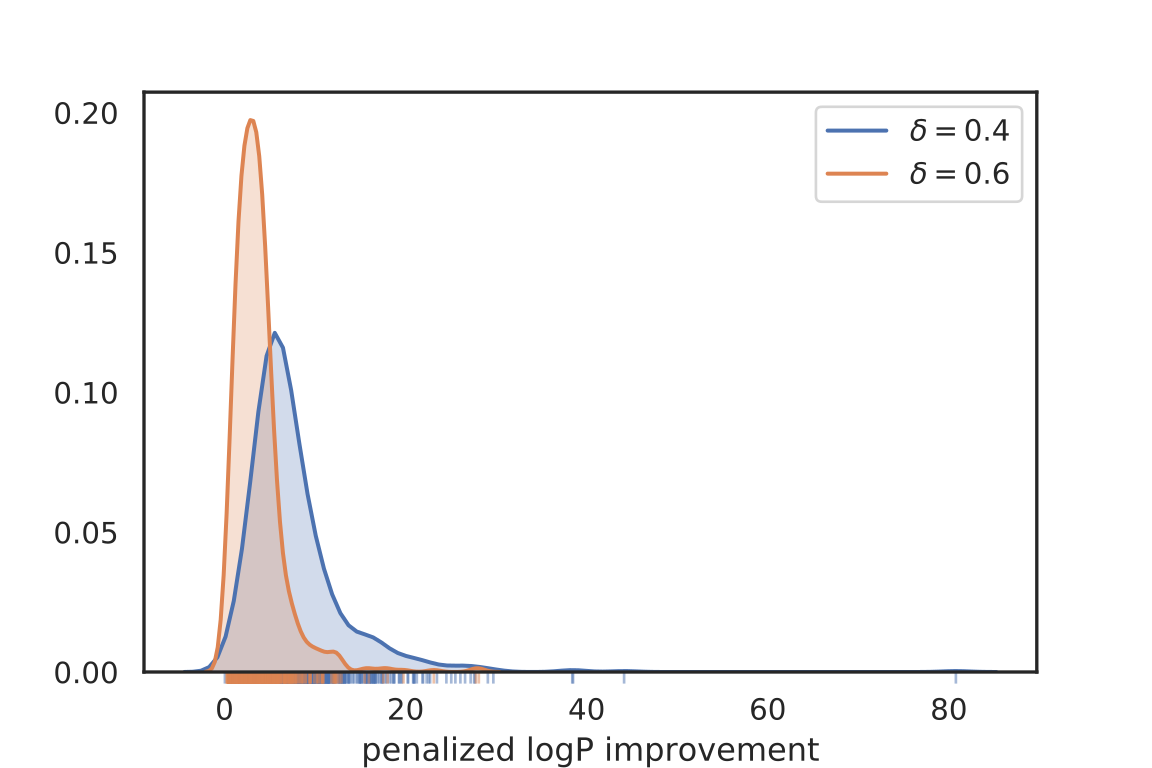}
    \caption{Distribution of improvement in penalized logP values after optimization by QMO on the 800 molecule set. A rug plot on the bottom shows the location of individual molecules for better visualization.}
    \label{fig:penalized_logP_dist}
\end{figure}

Figure \ref{fig:penalized_logP_dist} shows the distributions of improvement in penalized logP scores compared to the original molecules for the two similarity thresholds. The results show a long tail on the right-hand side, increasing the variance.

\subsection{SARS-CoV-2 Use Case}
\label{appendix_covid}

Table \ref{table_covid_targets} shows the results for improving binding affinity of 23 existing  SARS-CoV-2 M\textsuperscript{pro} inhibitor molecules. Affinity predictions, Tanimoto similarity, quantitative estimation of drug-likeness (QED) \cite{bickerton2012quantifying}, synthetic accessibility (SA) \cite{ertl2009estimation}, and the logarithm of partition coefficient (logP) properties are reported.
The results show that predicted affinity is improved past the threshold for every starting compound while attaining similarity of 0.64 on average. QED increases slightly, on average, showing QMO preserves and in some cases improves drug-likeness. SA increases only slightly, indicating synthesizability is still reasonable. Hydrophobicity (logP) decreases slightly meaning the molecules are more water-soluble. Finally, variants of all 11 M\textsuperscript{pro} inhibitors show better or comparable  binding free energy, when compared to that of the original one.

Table \ref{fig:covid_targets_2Dstructs} shows the original and improved versions of all 23 COVID-19 inhibitor candidates. Table \ref{tab:covid_targets_SMILES} shows the SMILES representations of these same molecules.

\begin{figure}[h]
    \centering
    \includegraphics[height=0.5\textheight,trim={300px 0 200px 0},clip]{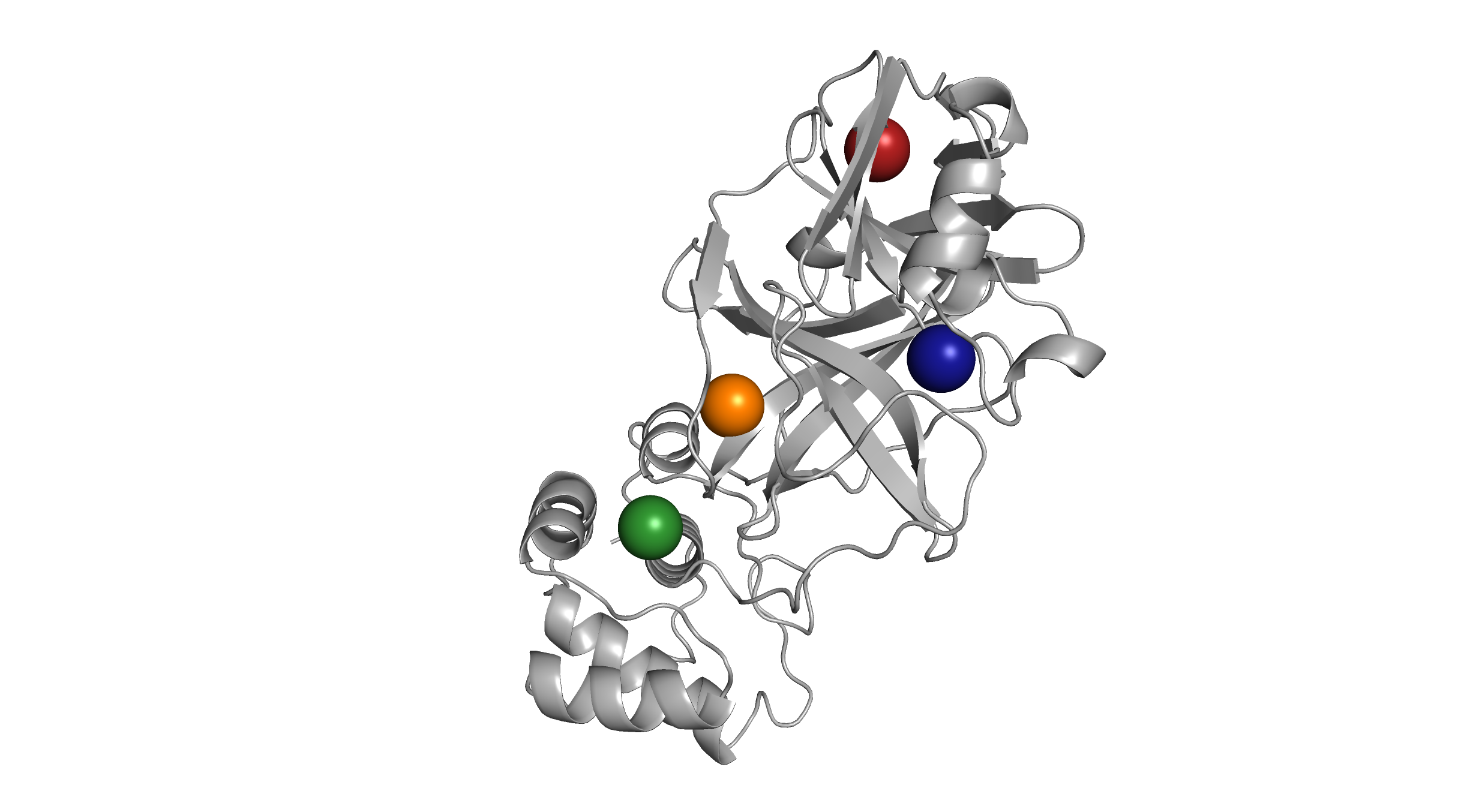}
    \caption{Locations of the binding pockets of M\textsuperscript{pro} from Table \ref{tab:docking_pockets}. The colored spheres represent the different pockets as follows: blue (0), orange (1), green (2), red (3).}
    \label{fig:mpro_pocket_locs}
\end{figure}

We also provide the extended results of docking analysis on the COVID-19 inhibitor candidates and their improved variants in Table \ref{tab:docking_pockets}.
We show MM/PBSA energy calculation results on the top 3 docking poses for each molecule.
In addition, we investigate if the top binding modes revealed in docking do correspond to any of binding pockets reported in  \cite{chenthamarakshan2020targetspecific}, which were estimated using PrankWeb\footnote{\url{http://prankweb.cz/}} and indexed by score (see Figure \ref{fig:mpro_pocket_locs} for the locations of these pockets). Note: pocket 0 corresponds to the substrate-binding pocket of M\textsuperscript{pro}.
If the pocket does not change between original and improved molecules, we can expect a similar mode of inhibition of the target which is desirable (in the cases where we know the experimentally validated binding pocket of the original drug, e.g. see Figure \ref{fig:dipyridamole}).

Finally, Table \ref{tab:dipyridamole_contacts} lists all of the residues that Dipyridamole contacts in it's original and optimized versions along with the number of heavy atom contacts. The improved version makes 73 contacts compared to 64 originally, in agreement with the improved MM/PBSA BFE estimate.

\newpage


\begin{table}[h]
\centering
\adjustbox{max width=0.97\textwidth}{
\begin{tabular}{@{}l|rr|c|rr|rr|rr|rr@{}}
\toprule
                   & \multicolumn{2}{c|}{\textcolor{blue}{Affinity}} & \textcolor{blue}{Similarity} & \multicolumn{2}{c|}{QED}  & \multicolumn{2}{c|}{SA}   & \multicolumn{2}{c|}{logP}  & \multicolumn{2}{c}{BFE}       \\
compound           & orig.          & imp.         &            & orig.        & imp.       & orig.        & imp.       & orig.        & imp.        & orig.         & imp.          \\ \midrule
Dipyridamole       & 3.94           & 7.59         & 0.58       & 0.31         & 0.74       & 2.99         & 2.95       & -0.02        & 1.34        & -11.49 (-21.39) & -25.65 \\
Favipiravir        & 4.32           & 8.44         & 0.46       & 0.55         & 0.54       & 2.90         & 2.79       & -0.99        & -1.48       & -0.77 (-7.91)          & -10.93 \\
Cinanserin         & 4.43           & 8.02         & 0.79       & 0.44         & 0.22       & 2.07         & 2.41       & 4.38         & 1.73        & -11.51 (-15.61) & -11.92 \\
Tideglusib         & 4.94           & 7.71         & 0.81       & 0.58         & 0.57       & 2.28         & 2.35       & 3.26         & 3.40        & ** & -13.31 \\
Bromhexine         & 5.00           & 7.53         & 0.67       & 0.78         & 0.66       & 2.38         & 2.42       & 4.56         & 3.50        & -20.41          & -17.89 \\
PX-12              & 5.01           & 7.51         & 0.47       & 0.74         & 0.79       & 3.98         & 4.24       & 2.95         & 3.55        & *        & *        \\
Ebselen            & 5.09           & 7.55         & 0.42       & 0.63         & 0.65       & 2.05         & 2.44       & 3.05         & 1.90        & -11.86 & -10.56          \\
Shikonin           & 5.11           & 7.71         & 0.42       & 0.58         & 0.71       & 3.41         & 4.06       & 2.12         & 1.82        & *        & *        \\
Disulfiram         & 5.54           & 7.51         & 0.57       & 0.57         & 0.36       & 3.12         & 3.66       & 3.62         & 5.37        & -20.43          & -9.81 (-17.98)          \\
Entecavir          & 5.55           & 7.68         & 0.39       & 0.53         & 0.40       & 4.09         & 4.78       & -0.83        & -1.29       & *        & *        \\
Hydroxychloroquine & 5.85           & 7.51         & 0.42       & 0.73         & 0.59       & 2.79         & 3.39       & 3.78         & 2.67        & *        & *        \\
Chloroquine        & 6.07           & 7.58         & 0.66       & 0.76         & 0.79       & 2.67         & 2.69       & 4.81         & 4.30        & *        & *        \\
O6K                & 6.18           & 7.70         & 0.62       & 0.25         & 0.71       & 4.12         & 3.56       & 2.22         & 0.75        & *        & *        \\
Remdesivir         & 6.32           & 7.77         & 0.61       & 0.16         & 0.45       & 4.82         & 4.73       & 2.31         & 1.00        & *        & *        \\
Umifenovir         & 6.36           & 7.60         & 0.73       & 0.38         & 0.42       & 2.68         & 2.63       & 5.18         & 5.12        & -16.08 & -20.87 \\
Lopinavir          & 6.42           & 8.03         & 0.76       & 0.20         & 0.46       & 3.90         & 3.66       & 4.33         & 3.60        & *        & *        \\
Ambroxol           & 6.46           & 7.97         & 0.64       & 0.71         & 0.75       & 2.50         & 3.22       & 3.19         & 2.17        & * & *        \\
GS-441524          & 6.56           & 7.97         & 0.73       & 0.50         & 0.83       & 4.38         & 4.91       & -1.86        & 0.44        & *        & *        \\
Nelfinavir         & 6.63           & 8.22         & 0.78       & 0.33         & 0.40       & 4.04         & 4.09       & 4.75         & 3.71        & *        & *        \\
Quercetin          & 6.70           & 7.56         & 0.66       & 0.43         & 0.51       & 2.54         & 2.36       & 1.99         & 2.28        & -12.71 & -8.32 (-8.96) \\
N3                 & 6.83           & 7.98         & 0.93       & 0.12         & 0.19       & 4.29         & 3.99       & 2.08         & 1.94        & *        & *        \\
Curcumin           & 6.86           & 7.62         & 0.89       & 0.55         & 0.55       & 2.43         & 2.45       & 3.37         & 3.37        & -7.31 & -3.67 (-15.94) \\
Kaempferol         & 6.90           & 7.72         & 0.67       & 0.55         & 0.64       & 2.37         & 2.36       & 2.28         & 3.23        & -11.86 & -13.48 \\ \midrule
Average            & 5.79           & 7.76         & 0.64       & 0.49         & 0.56       & 3.17         & 3.31       & 2.63         & 2.36        &  & \\\bottomrule
\end{tabular}
}
\caption{Results for improving binding affinity of 23 existing  SARS-CoV-2 M\textsuperscript{pro} inhibitor molecules. 
Original (orig.) and improved (imp.) values and their similarity are shown.
The affinity threshold is set to $\ge7.5$ ($\text{pIC}_{50}$) while maximizing the Tanimoto similarity. Only the properties highlighted in \textcolor{blue}{blue} are used in QMO. Relevant molecular properties --- QED, SA, and logP --- are also reported. The ``BFE'' column shows the binding free energy results obtained by MM/PBSA rescoring of the docking pose with the M\textsuperscript{pro} substrate binding pocket (lower is better) using the webserver of Fast Amber Rescoring for PPI Inhibitors\protect\footnotemark with default forcefield (AMBER GAFF2 and ff14sb  for the ligand and the protein protonated at pH=6.5, respectively) and parameter choices. More favorable BFE for alternative pocket is reported within parentheses. 
Chiral molecules (indicated with *) were excluded. Tideglusib was not possible to model with the webserver (**).}
\label{table_covid_targets}
\end{table}
\footnotetext{\url{http://cadd.zju.edu.cn/farppi/}}

\newpage

\begin{center}
\begin{longtable}{p{0.1\textwidth}cc|p{0.1\textwidth}cc}
& original & improved && original & improved \\\endfirsthead
\multicolumn{6}{l}{\bfseries \tablename\ \thetable{} continued}\\[10pt]
& original & improved && original & improved \\\endhead
Dipyridamole & \parbox[c]{0.175\textwidth}{\includegraphics[width=0.175\textwidth]{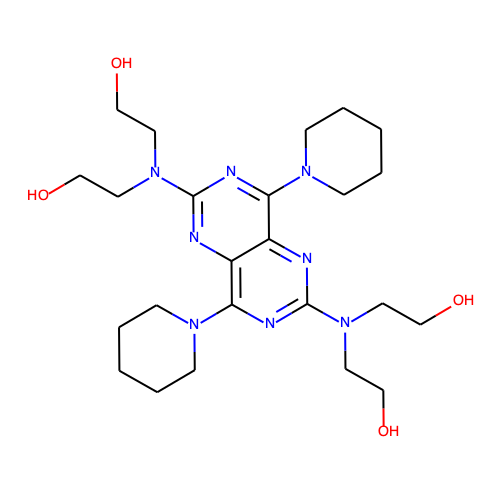}} & \parbox[c]{0.175\textwidth}{\includegraphics[width=0.175\textwidth]{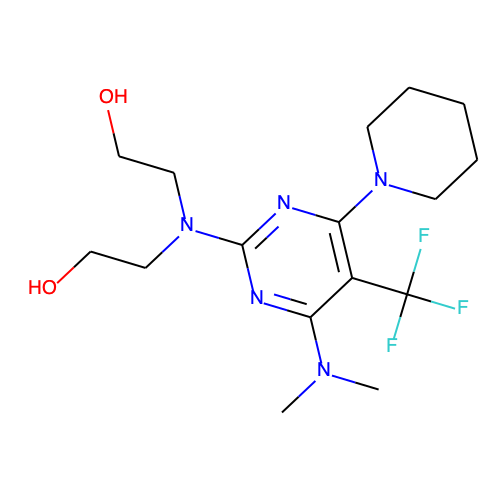}} &
Favipiravir & \parbox[c]{0.175\textwidth}{\includegraphics[width=0.175\textwidth]{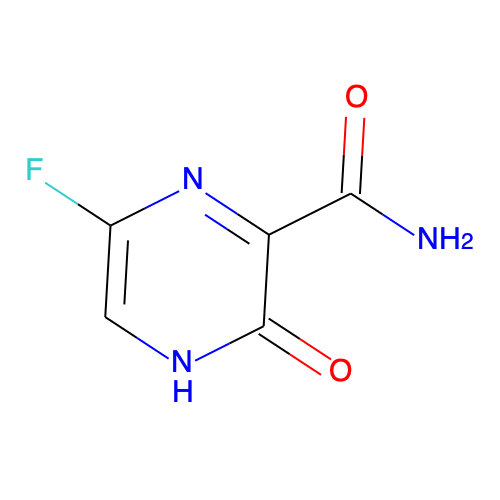}} & \parbox[c]{0.175\textwidth}{\includegraphics[width=0.175\textwidth]{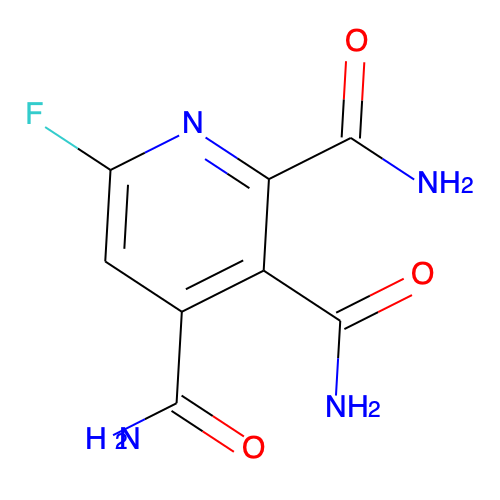}} \\*
Cinanserin & \parbox[c]{0.175\textwidth}{\includegraphics[width=0.175\textwidth]{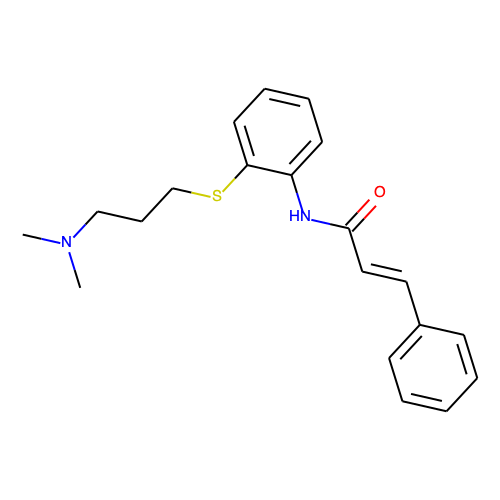}} & \parbox[c]{0.175\textwidth}{\includegraphics[width=0.175\textwidth]{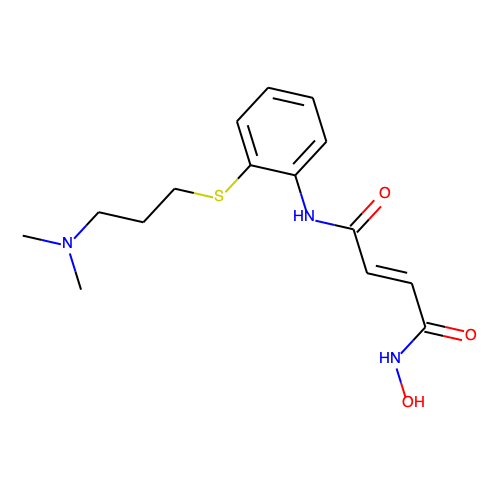}} &
Tideglusib & \parbox[c]{0.175\textwidth}{\includegraphics[width=0.175\textwidth]{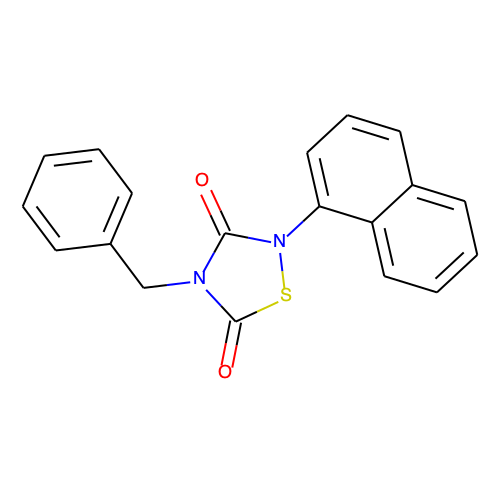}} & \parbox[c]{0.175\textwidth}{\includegraphics[width=0.175\textwidth]{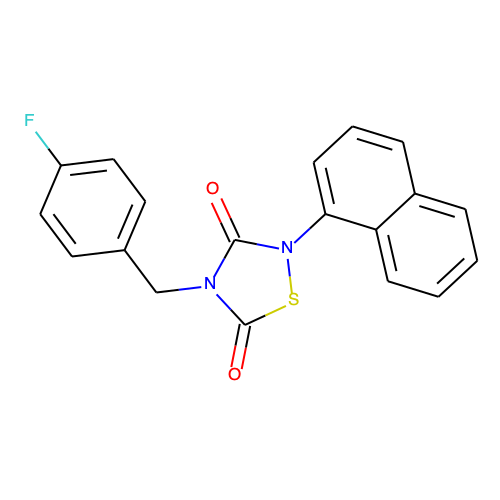}} \\*
Bromhexine & \parbox[c]{0.175\textwidth}{\includegraphics[width=0.175\textwidth]{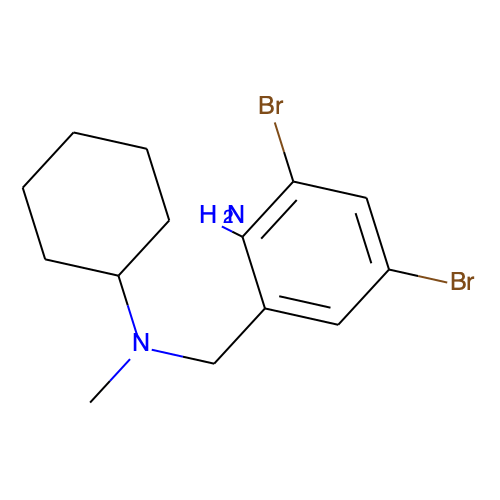}} & \parbox[c]{0.175\textwidth}{\includegraphics[width=0.175\textwidth]{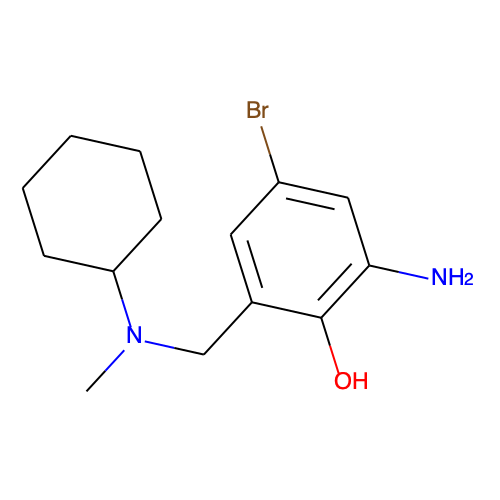}} &
PX-12 & \parbox[c]{0.175\textwidth}{\includegraphics[width=0.175\textwidth]{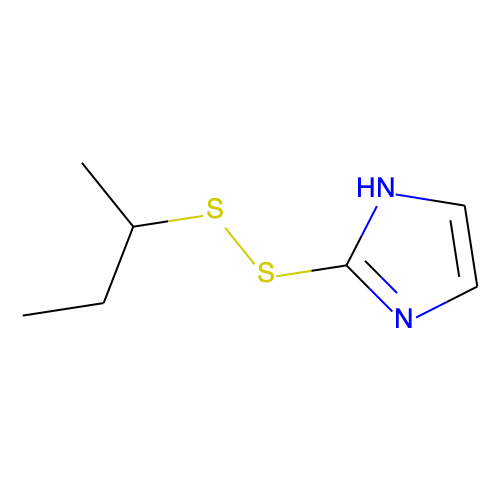}} & \parbox[c]{0.175\textwidth}{\includegraphics[width=0.175\textwidth]{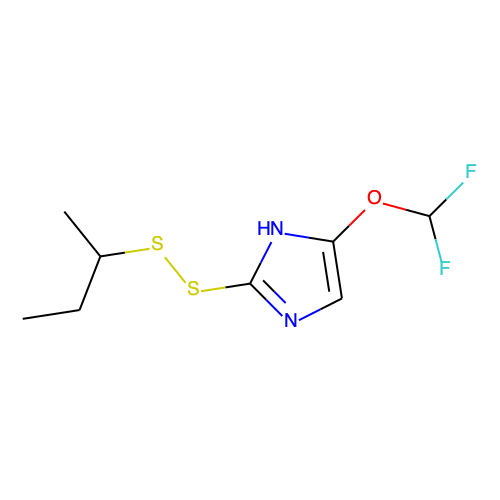}} \\*
Ebselen & \parbox[c]{0.175\textwidth}{\includegraphics[width=0.175\textwidth]{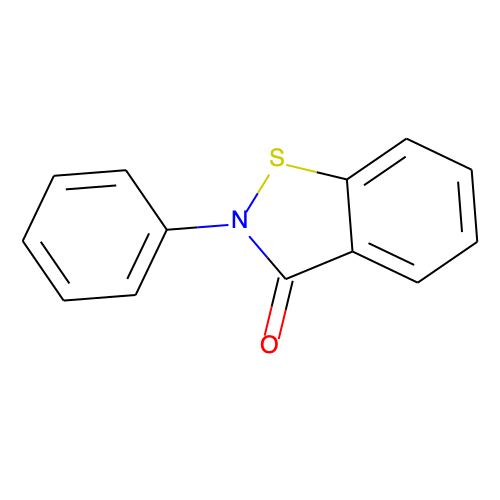}} & \parbox[c]{0.175\textwidth}{\includegraphics[width=0.175\textwidth]{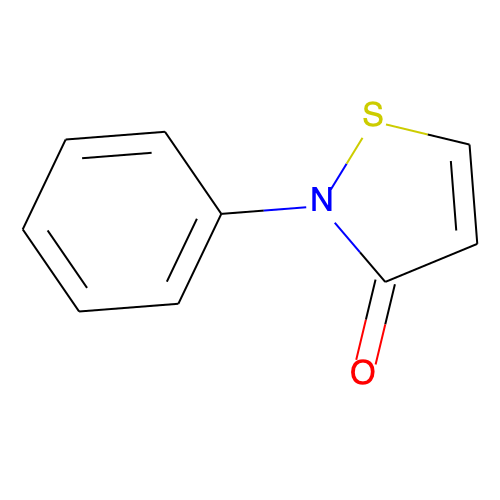}} &
Shikonin & \parbox[c]{0.175\textwidth}{\includegraphics[width=0.175\textwidth]{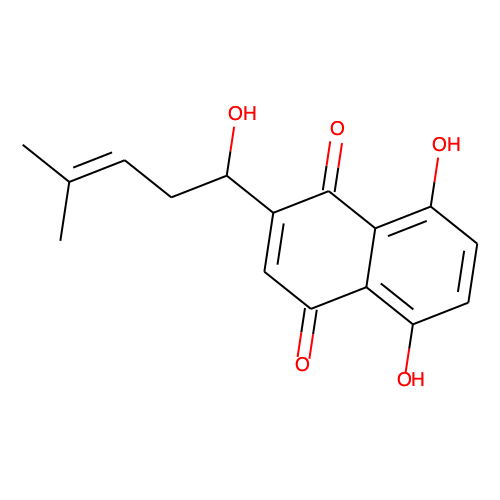}} & \parbox[c]{0.175\textwidth}{\includegraphics[width=0.175\textwidth]{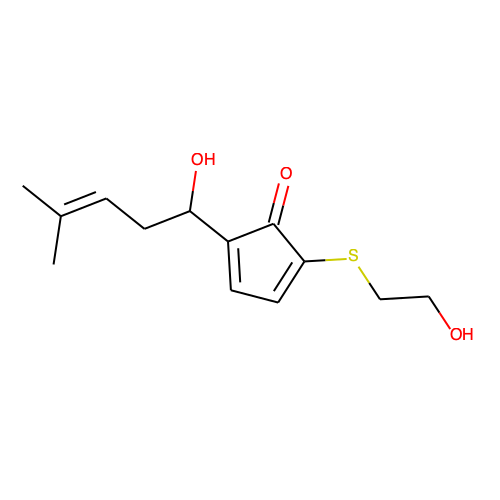}} \\*
Disulfiram & \parbox[c]{0.175\textwidth}{\includegraphics[width=0.175\textwidth]{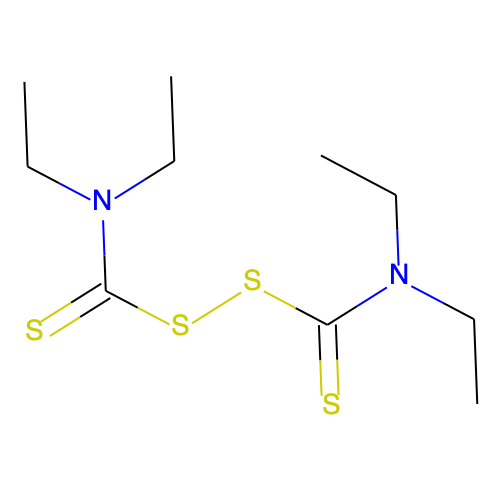}} & \parbox[c]{0.175\textwidth}{\includegraphics[width=0.175\textwidth]{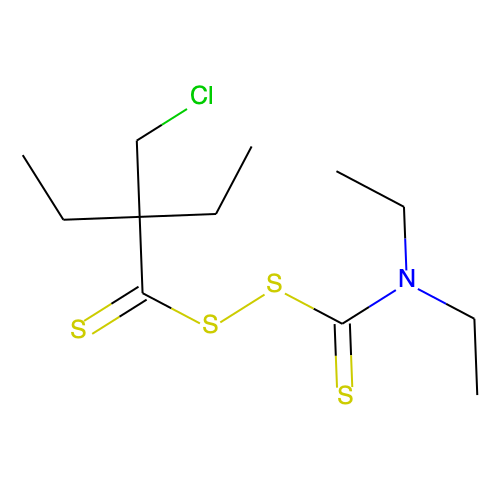}} &
Entecavir & \parbox[c]{0.175\textwidth}{\includegraphics[width=0.175\textwidth]{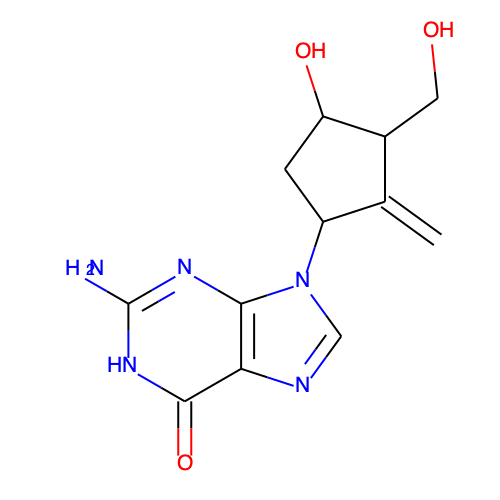}} & \parbox[c]{0.175\textwidth}{\includegraphics[width=0.175\textwidth]{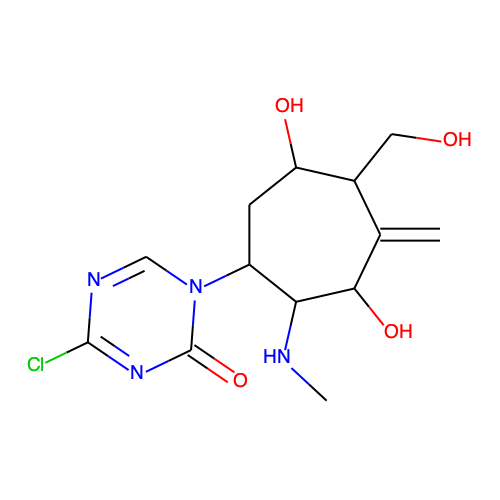}} \\
Hydroxy-\newline chloroquine & \parbox[c]{0.175\textwidth}{\includegraphics[width=0.175\textwidth]{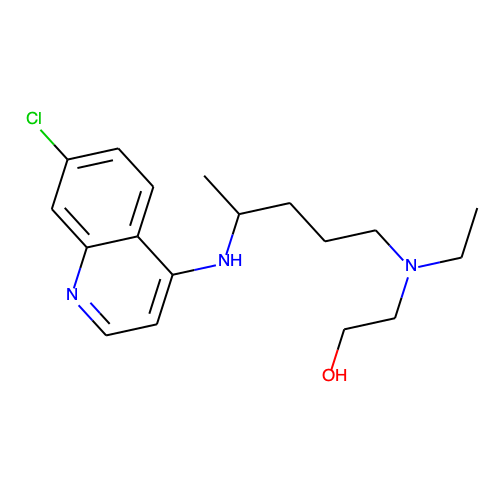}} & \parbox[c]{0.175\textwidth}{\includegraphics[width=0.175\textwidth]{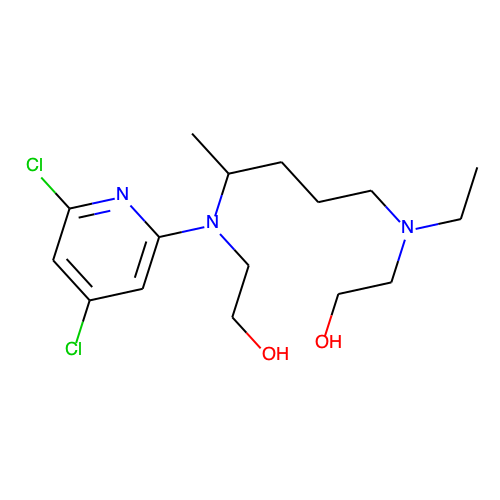}} &
Chloroquine & \parbox[c]{0.175\textwidth}{\includegraphics[width=0.175\textwidth]{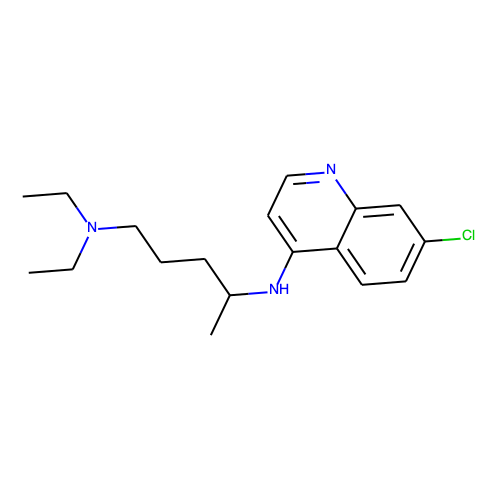}} & \parbox[c]{0.175\textwidth}{\includegraphics[width=0.175\textwidth]{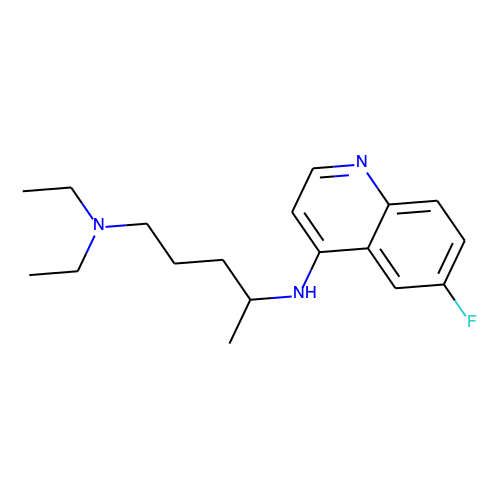}} \\
O6K & \parbox[c]{0.175\textwidth}{\includegraphics[width=0.175\textwidth]{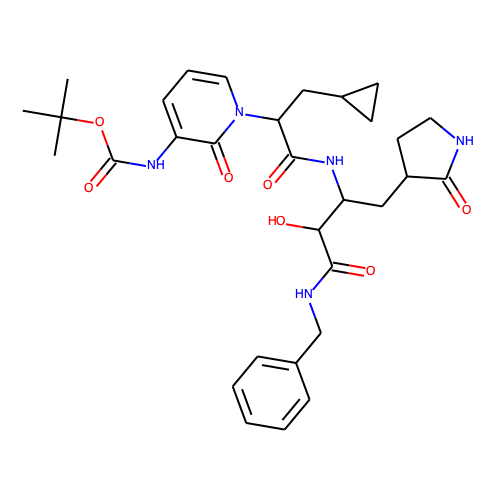}} & \parbox[c]{0.175\textwidth}{\includegraphics[width=0.175\textwidth]{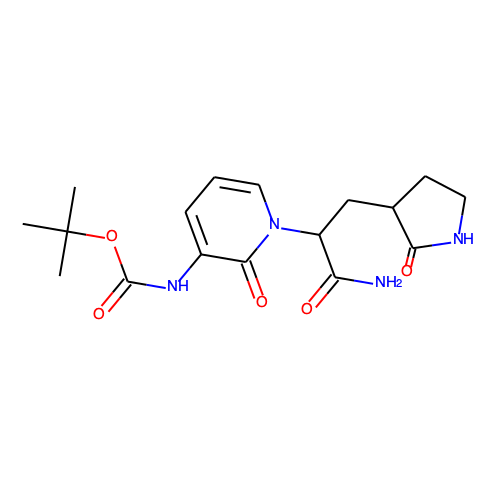}} &
Remdesivir & \parbox[c]{0.175\textwidth}{\includegraphics[width=0.175\textwidth]{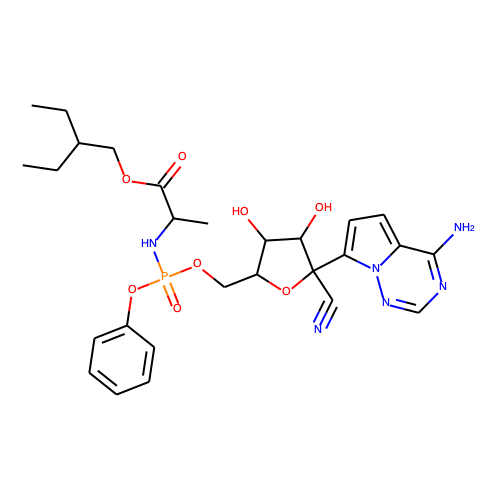}} & \parbox[c]{0.175\textwidth}{\includegraphics[width=0.175\textwidth]{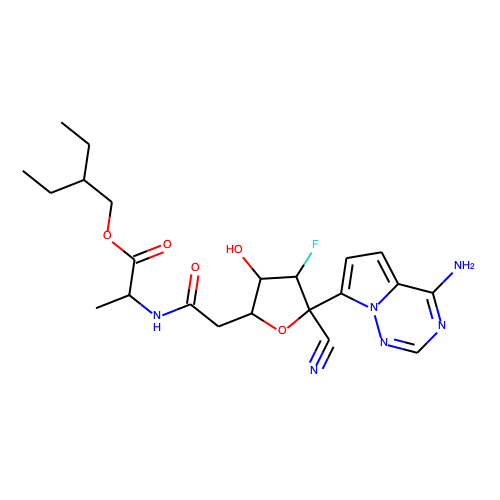}} \\
Umifenovir & \parbox[c]{0.175\textwidth}{\includegraphics[width=0.175\textwidth]{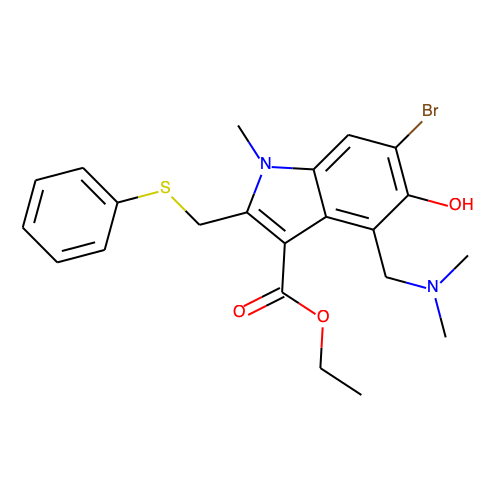}} & \parbox[c]{0.175\textwidth}{\includegraphics[width=0.175\textwidth]{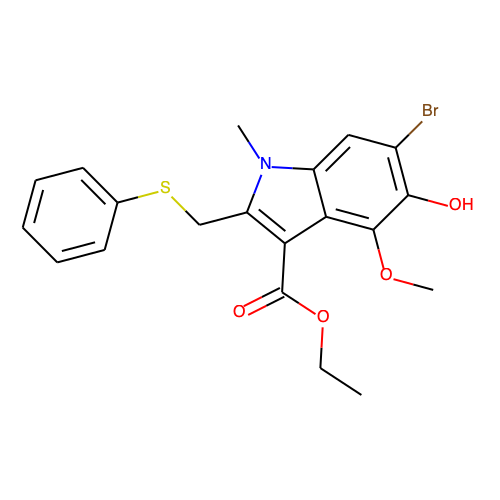}} &
Lopinavir & \parbox[c]{0.175\textwidth}{\includegraphics[width=0.175\textwidth]{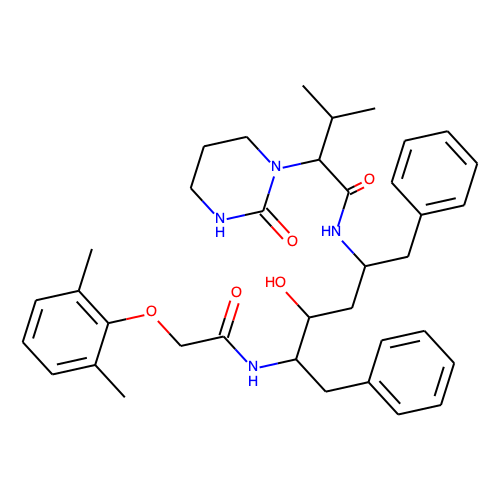}} & \parbox[c]{0.175\textwidth}{\includegraphics[width=0.175\textwidth]{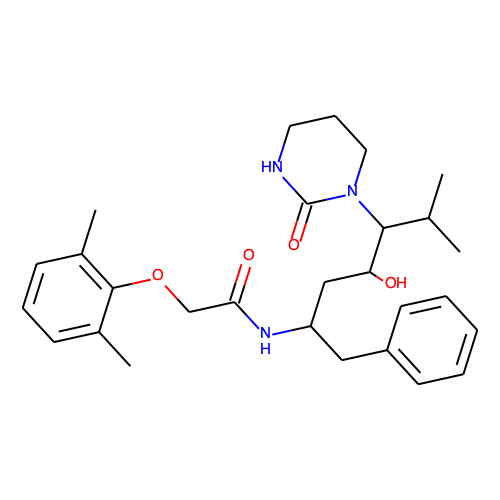}} \\
Ambroxol & \parbox[c]{0.175\textwidth}{\includegraphics[width=0.175\textwidth]{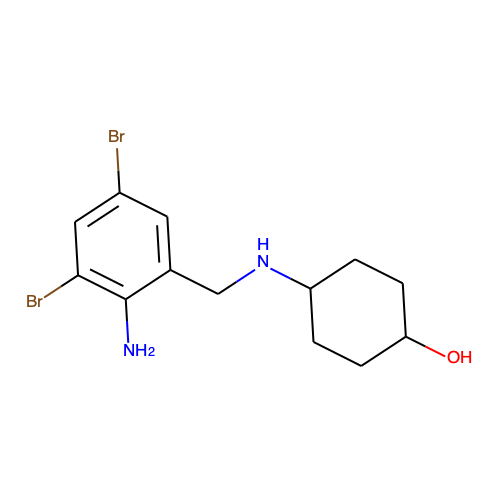}} & \parbox[c]{0.175\textwidth}{\includegraphics[width=0.175\textwidth]{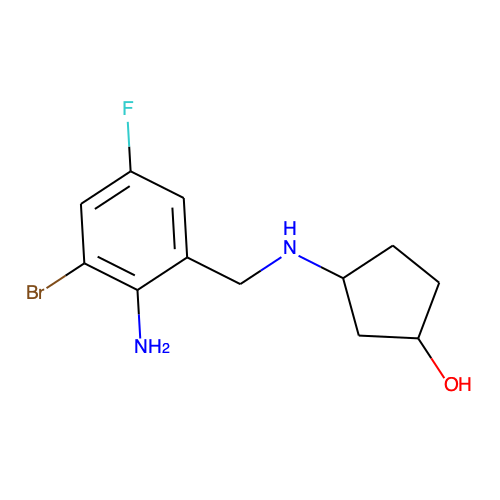}} &
GS-441524 & \parbox[c]{0.175\textwidth}{\includegraphics[width=0.175\textwidth]{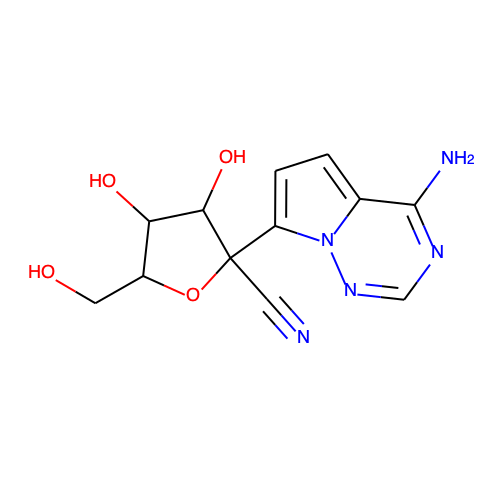}} & \parbox[c]{0.175\textwidth}{\includegraphics[width=0.175\textwidth]{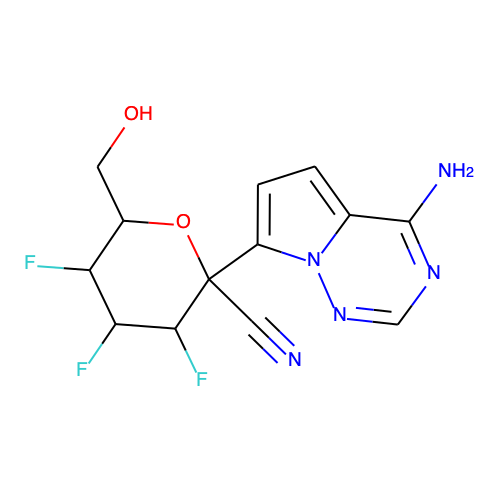}} \\*
Nelfinavir & \parbox[c]{0.175\textwidth}{\includegraphics[width=0.175\textwidth]{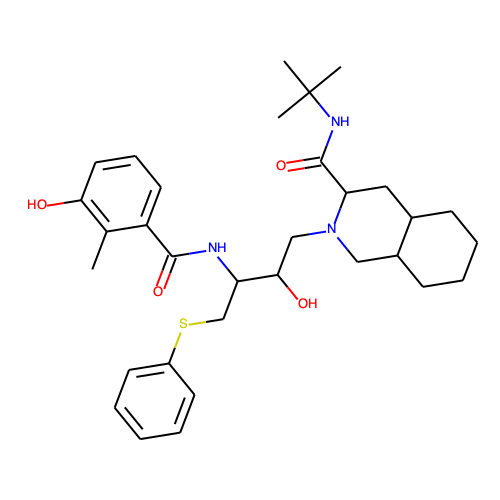}} & \parbox[c]{0.175\textwidth}{\includegraphics[width=0.175\textwidth]{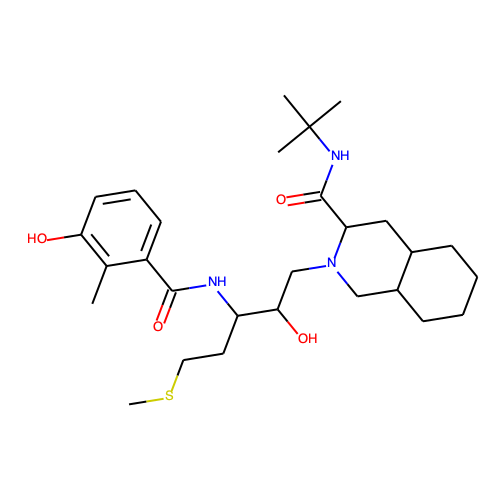}} &
Quercetin & \parbox[c]{0.175\textwidth}{\includegraphics[width=0.175\textwidth]{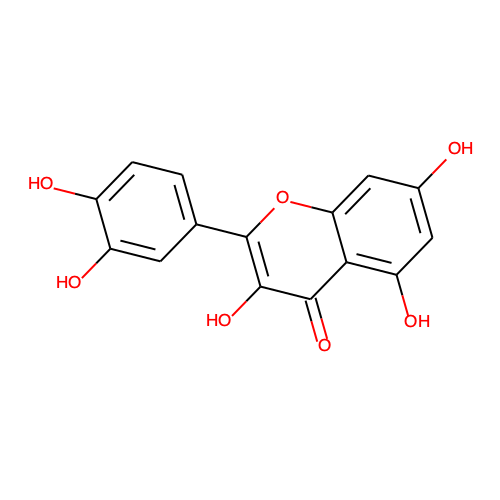}} & \parbox[c]{0.175\textwidth}{\includegraphics[width=0.175\textwidth]{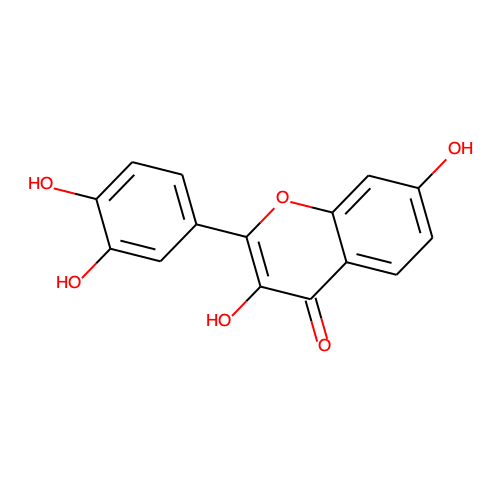}} \\*
N3 & \parbox[c]{0.175\textwidth}{\includegraphics[width=0.175\textwidth]{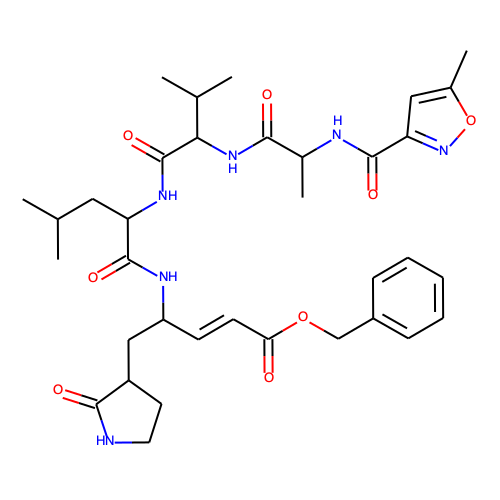}} & \parbox[c]{0.175\textwidth}{\includegraphics[width=0.175\textwidth]{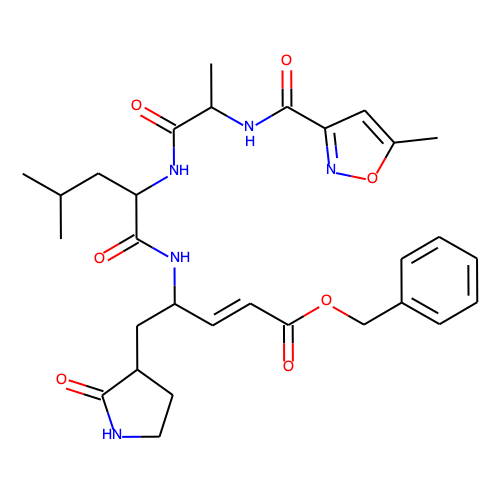}} &
Curcumin & \parbox[c]{0.175\textwidth}{\includegraphics[width=0.175\textwidth]{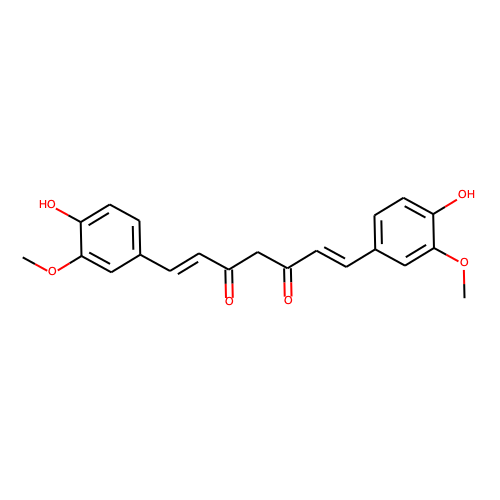}} & \parbox[c]{0.175\textwidth}{\includegraphics[width=0.175\textwidth]{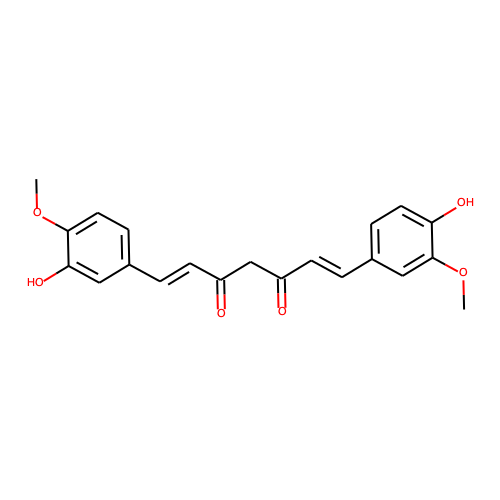}} \\*
Kaempferol & \parbox[c]{0.175\textwidth}{\includegraphics[width=0.175\textwidth]{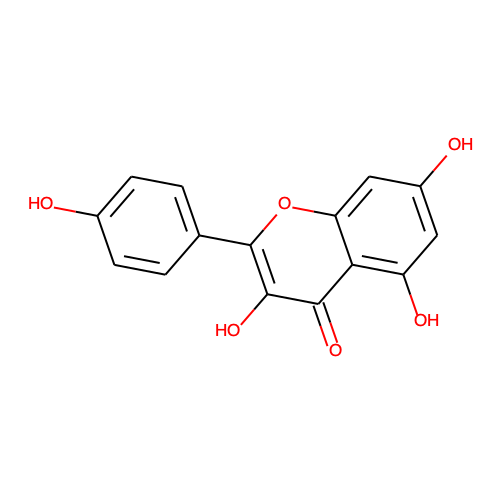}} & \parbox[c]{0.175\textwidth}{\includegraphics[width=0.175\textwidth]{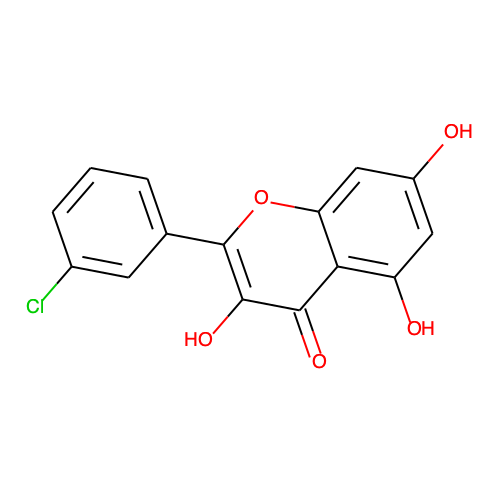}} &&&\\*
\caption{Results of QMO optimization of existing inhibitor molecules for COVID-19. For each pair of columns on either side of the dividing line, the left column shows the 2D structure of the original molecules and the right column shows the 2D structure of the QMO-optimized molecules which started from the initial state to the left with the original inhibitor name to the left of both.}
\label{fig:covid_targets_2Dstructs}
\end{longtable}
\end{center}

\newpage

\begin{table}[t!]
\centering
\adjustbox{max width=0.99\textwidth, max totalheight=0.89\textheight}{
\begin{tabular}{@{}lcl@{}}
\toprule
compound                             &          & smiles                                                                                 \\
\midrule
\multirow{2}{*}{Dipyridamole}        & original & OCCN(CCO)c1nc(N2CCCCC2)c2nc(N(CCO)CCO)nc(N3CCCCC3)c2n1                                 \\
                                     & improved & CN(C)c1nc(N(CCO)CCO)nc(N2CCCCC2)c1C(F)(F)F                                             \\[8pt]
\multirow{2}{*}{Favipiravir}         & original & NC(=O)c1nc(F)c[nH]c1=O                                                                 \\
                                     & improved & NC(=O)c1cc(F)nc(C(N)=O)c1C(N)=O                                                        \\[8pt]
\multirow{2}{*}{Cinanserin}          & original & CN(C)CCCSc1ccccc1NC(=O)C=Cc1ccccc1                                                     \\
                                     & improved & CN(C)CCCSc1ccccc1NC(=O)C=CC(=O)NO                                                      \\[8pt]
\multirow{2}{*}{Tideglusib}          & original & O=c1sn(-c2cccc3ccccc23)c(=O)n1Cc1ccccc1                                                \\
                                     & improved & O=c1sn(-c2cccc3ccccc23)c(=O)n1Cc1ccc(F)cc1                                             \\[8pt]
\multirow{2}{*}{Bromhexine}          & original & CN(Cc1cc(Br)cc(Br)c1N)C1CCCCC1                                                         \\
                                     & improved & CN(Cc1cc(Br)cc(N)c1O)C1CCCCC1                                                          \\[8pt]
\multirow{2}{*}{PX-12}               & original & CCC(C)SSc1ncc[nH]1                                                                     \\
                                     & improved & CCC(C)SSc1ncc(OC(F)F)[nH]1                                                             \\[8pt]
\multirow{2}{*}{Ebselen\footnotemark}& original & O=c1c2ccccc2sn1-c1ccccc1                                                               \\
                                     & improved & O=c1ccsn1-c1ccccc1                                                                     \\[8pt]
\multirow{2}{*}{Shikonin}            & original & CC(C)=CCC(O)C1=CC(=O)c2c(O)ccc(O)c2C1=O                                                \\
                                     & improved & CC(C)=CCC(O)C1=CC=C(SCCO)C1=O                                                          \\[8pt]
\multirow{2}{*}{Disulfiram}          & original & CCN(CC)C(=S)SSC(=S)N(CC)CC                                                             \\
                                     & improved & CCN(CC)C(=S)SSC(=S)C(CC)(CC)CCl                                                        \\[8pt]
\multirow{2}{*}{Entecavir}           & original & C=C1C(CO)C(O)CC1n1cnc2c(=O)[nH]c(N)nc21                                                \\
                                     & improved & C=C1C(O)C(NC)C(n2cnc(Cl)nc2=O)CC(O)C1CO                                                \\[8pt]
\multirow{2}{*}{Hydroxychloroquine}  & original & CCN(CCO)CCCC(C)Nc1ccnc2cc(Cl)ccc12                                                     \\
                                     & improved & CCN(CCO)CCCC(C)N(CCO)c1cc(Cl)cc(Cl)n1                                                  \\[8pt]
\multirow{2}{*}{Chloroquine}         & original & CCN(CC)CCCC(C)Nc1ccnc2cc(Cl)ccc12                                                      \\
                                     & improved & CCN(CC)CCCC(C)Nc1ccnc2ccc(F)cc12                                                       \\[8pt]
\multirow{2}{*}{O6K}                 & original & CC(C)(C)OC(=O)Nc1cccn(C(CC2CC2)C(=O)NC(CC2CCNC2=O)C(O)C(=O)NCc2ccccc2)c1=O             \\
                                     & improved & CC(C)(C)OC(=O)Nc1cccn(C(CC2CCNC2=O)C(N)=O)c1=O                                         \\[8pt]
\multirow{2}{*}{Remdesivir}          & original & CCC(CC)COC(=O)C(C)NP(=O)(OCC1OC(C\#N)(c2ccc3c(N)ncnn23)C(O)C1O)Oc1ccccc1                \\
                                     & improved & CCC(CC)COC(=O)C(C)NC(=O)CC1OC(C\#N)(c2ccc3c(N)ncnn23)C(F)C1O                            \\[8pt]
\multirow{2}{*}{umifenovir}          & original & CCOC(=O)c1c(CSc2ccccc2)n(C)c2cc(Br)c(O)c(CN(C)C)c12                                    \\
                                     & improved & CCOC(=O)c1c(CSc2ccccc2)n(C)c2cc(Br)c(O)c(OC)c12                                        \\[8pt]
\multirow{2}{*}{lopinavir}           & original & Cc1cccc(C)c1OCC(=O)NC(Cc1ccccc1)C(O)CC(Cc1ccccc1)NC(=O)C(C(C)C)N1CCCNC1=O              \\
                                     & improved & Cc1cccc(C)c1OCC(=O)NC(Cc1ccccc1)CC(O)C(C(C)C)N1CCCNC1=O                                \\[8pt]
\multirow{2}{*}{Ambroxol}            & original & Nc1c(Br)cc(Br)cc1CNC1CCC(O)CC1                                                         \\
                                     & improved & Nc1c(Br)cc(F)cc1CNC1CCC(O)C1                                                           \\[8pt]
\multirow{2}{*}{GS-441524}           & original & N\#CC1(c2ccc3c(N)ncnn23)OC(CO)C(O)C1O                                                   \\
                                     & improved & N\#CC1(c2ccc3c(N)ncnn23)OC(CO)C(F)C(F)C1F                                               \\[8pt]
\multirow{2}{*}{Nelfinavir}          & original & Cc1c(O)cccc1C(=O)NC(CSc1ccccc1)C(O)CN1CC2CCCCC2CC1C(=O)NC(C)(C)C                       \\
                                     & improved & CSCCC(NC(=O)c1cccc(O)c1C)C(O)CN1CC2CCCCC2CC1C(=O)NC(C)(C)C                             \\[8pt]
\multirow{2}{*}{Quercetin}           & original & O=c1c(O)c(-c2ccc(O)c(O)c2)oc2cc(O)cc(O)c12                                             \\
                                     & improved & O=c1c(O)c(-c2ccc(O)c(O)c2)oc2cc(O)ccc12                                                \\[8pt]
\multirow{2}{*}{N3}                  & original & Cc1cc(C(=O)NC(C)C(=O)NC(C(=O)NC(CC(C)C)C(=O)NC(C=CC(=O)OCc2ccccc2)CC2CCNC2=O)C(C)C)no1 \\
                                     & improved & Cc1cc(C(=O)NC(C)C(=O)NC(CC(C)C)C(=O)NC(C=CC(=O)OCc2ccccc2)CC2CCNC2=O)no1               \\[8pt]
\multirow{2}{*}{Curcumin}            & original & COc1cc(C=CC(=O)CC(=O)C=Cc2ccc(O)c(OC)c2)ccc1O                                          \\
                                     & improved & COc1ccc(C=CC(=O)CC(=O)C=Cc2ccc(O)c(OC)c2)cc1O                                          \\[8pt]
\multirow{2}{*}{Kaempferol}          & original & O=c1c(O)c(-c2ccc(O)cc2)oc2cc(O)cc(O)c12                                                \\
                                     & improved & O=c1c(O)c(-c2cccc(Cl)c2)oc2cc(O)cc(O)c12                                               \\
\bottomrule
\end{tabular}

}
\caption{SMILES representations of original and improved (QMO-optimized) inhibitor molecules.}
\label{tab:covid_targets_SMILES}
\end{table}
\footnotetext{The selenium atom in Ebselen is rare for drug molecules and cannot be handled by the encoder/decoder so it is substituted for sulfur before beginning optimization as in \cite{jin2020structure}.}

\begin{table}[t]
\centering
\adjustbox{max width=0.99\textwidth}{
\begin{tabular}{@{}l|rrr|rrr|ccc|ccc@{}}
\toprule
             & \multicolumn{6}{c|}{Binding Free Energy} & \multicolumn{6}{c}{Binding Pocket} \\
             & \multicolumn{3}{c|}{orig. pose} & \multicolumn{3}{c|}{imp. pose} & \multicolumn{3}{c|}{orig. pose} & \multicolumn{3}{c}{imp. pose} \\
compound     & 1 & 2 & 3 & 1 & 2 & 3 & 1 & 2 & 3 & 1 & 2 & 3 \\
\midrule
Bromhexine   & -11.89                     & -14.01                     & \textcolor{red}{-20.41} & -13.86                        & -15.72                       & \textcolor{red}{-17.89}    & 1 & 0 & 0 & 1 & 1 & 0 \\
Cinanserin   & -5.95                      & \textcolor{red}{-11.55} & -15.61                     & \textcolor{red}{-11.92}    & -0.64                        & -8.66                         & 0 & 0 & 1 & 0 & 0 & 0 \\
Curcumin     & \textcolor{red}{-7.31}  & -5.36                      & -1.31                      & \textcolor{red}{-3.67}     & -2.01                        & \textcolor{blue}{-15.94} & 0 & 0 & 0 & 0 & 0 & 1 \\
Dipyridamole & -21.39                     & -17.89                     & \textcolor{red}{-11.49} & \textcolor{red}{-25.65}    & -10.89                       & -14.11                        & 2 & 2 & 0 & 0 & 0 & 0 \\
Disulfiram   & -20.39                     & \textcolor{red}{-20.43} & -19.73                     & \textcolor{blue}{-17.98} & \textcolor{red}{-9.81}    & -9.99                         & 0 & 0 & 0 & 1 & 0 & 1 \\
Ebselen      & -11.16                     & -11.69                     & \textcolor{red}{-11.86} & \textcolor{red}{-10.56}    &                              &                               & 1 & 0 & 0 & 0 & 2 & 3 \\
Favipiravir  & -7.91                      & -4.51                      & \textcolor{red}{-0.77}  & -7.04                         & \textcolor{red}{-10.93}   & -10.86                        & 1 & 2 & 0 & 1 & 0 & 1 \\
Kaempferol   & -7.42                      & \textcolor{red}{-11.86} & -11.18                     & \textcolor{red}{-13.48}    & -6.52                        & -11.45                        & 1 & 0 & 0 & 0 & 0 & 0 \\
Quercetin    & \textcolor{red}{-12.71} & -9.68                      & -10.08                     & \textcolor{red}{-8.32}     & \textcolor{blue}{-8.96} & 9.25                          & 0 & 0 & 0 & 0 & 3 & 0 \\
Tideglusib   & **                     &                            &                            & -13.31                        &                              &                               & 0 & 1 & 0 & 0 & 0 & 1 \\
Umifenovir   & \textcolor{red}{-16.08} & -7.3                       & -16.03                     & -13.75                        & -20.11                       & \textcolor{red}{-20.87}    & 0 & 0 & 0 & 0 & 0 & 0 \\
\bottomrule
\end{tabular}
}
\caption{Extended docking analysis: For each molecule, the top 3 docking poses (ranked by docking score) are shown with MM/PBSA binding free energy (kcal/mol) and corresponding binding pocket number. The best energy scores are highlighted: in \textcolor{red}{red} for the best energy that docks in pocket 0 or in \textcolor{blue}{blue} if the lowest energy over the top 3 poses corresponds to a different pocket. Binding pockets are the same as identified in prior studies\cite{chenthamarakshan2020targetspecific}. As noted previously, Tideglusib (original) was not possible to model with AMBER forcefield  (**).}
\label{tab:docking_pockets}
\end{table}

\begin{table}[h]
\centering
\begin{tabular}{@{}r|cc@{}}
    \toprule
    Residue Num. & Original & QMO-optimized \\
    \midrule
    25  & \checkmark &            \\
    27  & \checkmark &            \\
    41  & \checkmark & \checkmark \\
    46  & \checkmark &            \\
    49  & \checkmark & \checkmark \\
    140 & \checkmark &            \\
    141 & \checkmark &            \\
    142 & \checkmark & \checkmark \\
    143 & \checkmark &            \\
    144 & \checkmark &            \\
    145 & \checkmark & \checkmark \\
    163 & \checkmark &            \\
    164 & \checkmark & \checkmark \\
    165 & \checkmark & \checkmark \\
    166 & \checkmark & \checkmark \\
    167 &            & \checkmark \\
    168 &            & \checkmark \\
    186 &            & \checkmark \\
    187 &            & \checkmark \\
    188 &            & \checkmark \\
    189 & \checkmark & \checkmark \\
    190 &            & \checkmark \\
    191 &            & \checkmark \\
    192 &            & \checkmark \\
    \midrule
    Total contacts & 64 & 73 \\ 
    \bottomrule
\end{tabular}
\caption{Full set of contacts between SARS-CoV-2 M\textsuperscript{pro} and Dipyridamole, original and QMO-optimized variant. Residues that contact either of the molecule variants are listed by number. The bottom row shows the total number of heavy atom contacts with a 5 \AA\ cutoff.}
\label{tab:dipyridamole_contacts}
\end{table}

\clearpage
\subsection{AMP Optimization}

\begin{figure}[htb]
\centering
  \begin{subfigure}[b]{0.48\textwidth}
    \centering
    \includegraphics[width=\textwidth]{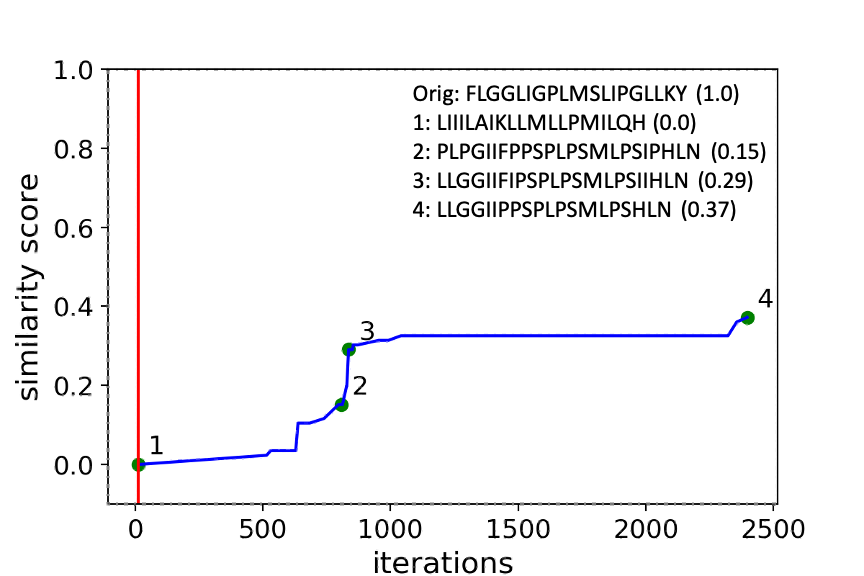}
    \caption{\texttt{FLGGLIGPLMSLIPGLLKY}}
  \end{subfigure}
  \begin{subfigure}[b]{0.48\textwidth}
    \centering
    \includegraphics[width=\textwidth]{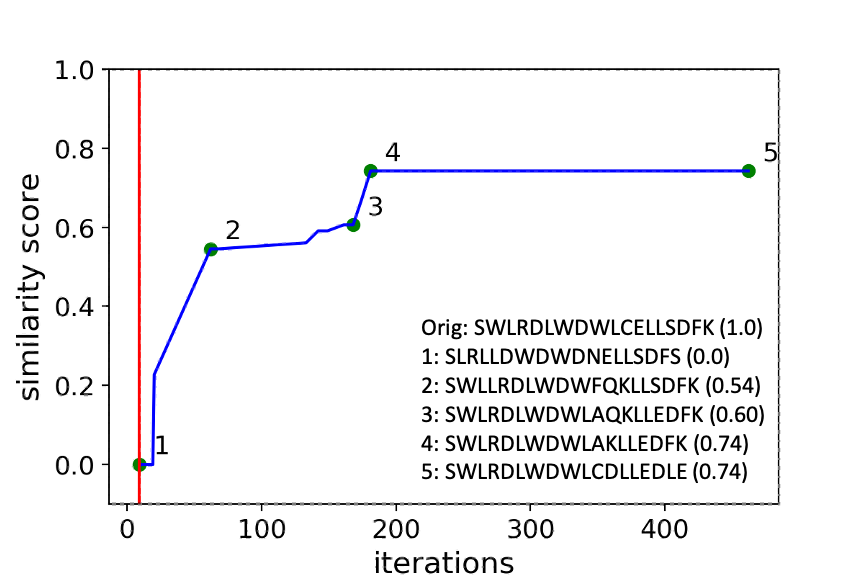}
    \caption{\texttt{SWLRDLWDWLCELLSDFK}}
  \end{subfigure}
  \begin{subfigure}[b]{0.48\textwidth}
    \centering
    \includegraphics[width=\textwidth]{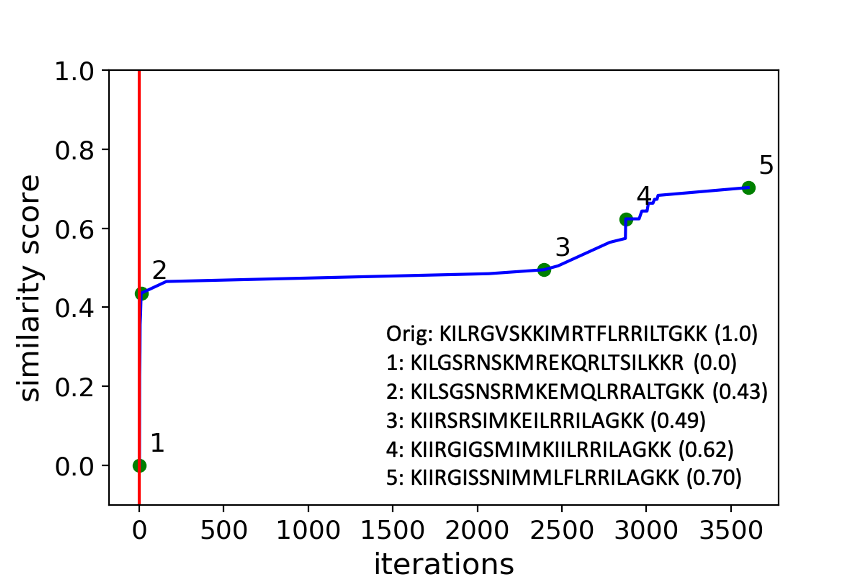}
    \caption{\texttt{KILRGVSKKIMRTFLRRILTGKK}}
  \end{subfigure}
  \begin{subfigure}[b]{0.48\textwidth}
    \centering
    \includegraphics[width=\textwidth]{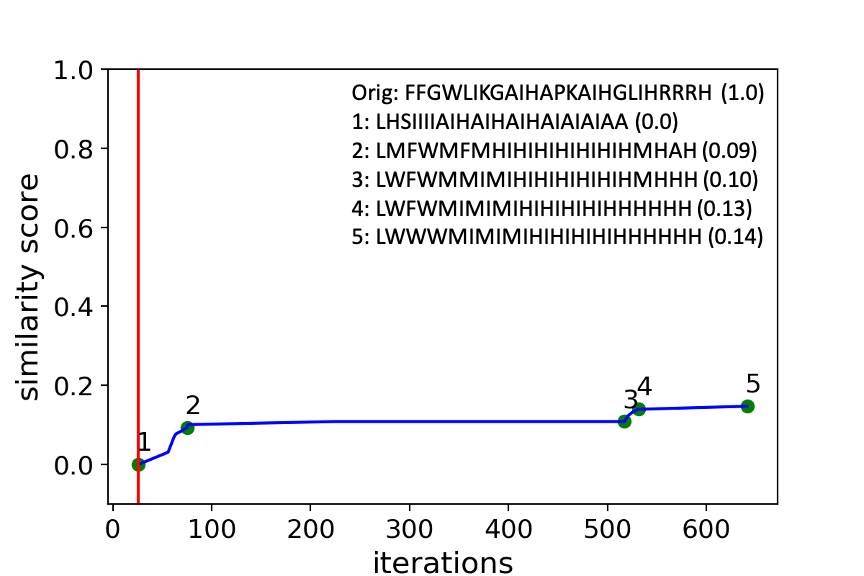}
    \caption{\texttt{FFGWLIKGAIHAPKAIHGLIHRRRH}}
  \end{subfigure}
\caption{Similarity (rescaled to 0-1 range with self-similarity as upper bound) of successful optimization runs for four peptide sequences. The curve shows the similarity of best-found valid molecule (highest similarity and predicted as non-toxic AMP) with respect to iteration counts in QMO.
The original sequence and few selected improved sequences are shown in the legend, along with their similarity score relative to the original sequence. The red line indicates the iteration of the first success (first time finding a qualified molecule).}
\label{fig:dist_impr}
\end{figure}

Figure \ref{fig:dist_impr} shows the similarity (normalized by self-similarity of the starting sequence) as QMO continues past the first success and the evolution of the sequences of selected iterations.  The curve shows the similarity of best-found valid molecule (highest similarity and predicted as non-toxic AMP) with respect to iteration counts in QMO. We can see that QMO keeps on improving similarity of best-found candidate while maintaining low toxicity and amp property as the optimization process continues. 
For these experiments we used $Q=100$, $\beta={1,10}$, $\alpha_0=\{0.1, 0.05, 0.01\}$, $\lambda_{\text{sim}}=0.01$, and $T=5000$.

Table \ref{tab:ext-classifiers} summarizes the results of AMP optimization as displayed in Figure \ref{fig:AMP}. Table \ref{tab:bio_properties} shows several physicochemical properties of the 109 original and QMO-optimized sequences displayed in Figure \ref{fig:bio_properties}.

Existing machine learning-based AMP classifiers report a wide range of accuracy. For example, the reported accuracy is 66.8\% for iAMP Pred \cite{meher2017predicting},  79\% for DBAASP-SP \cite{vishnepolsky2018predictive} that relies on local density-based sampling using physico-chemical features, and 94\%  for Witten \textit{et al.} method \cite{witten2019deep} that uses a convolutional neural net trained directly on a large  corpus of peptide sequences. Our sequence-level LSTM model shows a comparable 88\% accuracy.
 
 Among the available toxicity predictors, many focuses on predicting hemolytic nature. For example, HLPpred-fuse \cite{hasan2020hlppred} fuses predicted probabilities from six different classifiers to the final model. In contrast, HAPPENN \cite{timmons2020happenn} uses a neural net model and reports  84\% accuracy. Ref. \cite{Plisson2020MachineLD}  uses a gradient boosting based model and obtains a 95\% accuracy, similar to the in-house sequence-level toxicity classifier. We emphasize that the goal of this study is not to provide a new antimicrobial or toxicity prediction method that outperforms  existing machine learning-based  classifiers. Rather our goal is to leverage feedback from a set of reliable predictors to guide searches in the latent space. It should be noted that, comparing different AMP (as well as toxicity) prediction models is non-trivial, as different methods widely vary by training  dataset size, sequence length, different definition of positive and negative sequences, and other data curation criteria.

\begin{table}[ht]
    \centering
    \begin{tabular}{p{4.8cm}|p{1.4cm}|p{2.8cm}|p{2.8cm}|p{3.2cm}}
        \toprule
        AMP and Toxicity Classifiers & Reported Accuracy (\%) &  Prediction Accuracy (\%) on Starting Sequences 
        & Prediction Rate (\%) on QMO-optimized Sequences 
        & Improvement (\# of toxins (\textcolor{red}{T}) in starting sequences $\rightarrow$ \# of non-toxins (\textcolor{blue}{NT}) in optimized sequences \\ 
        \hline   
        AMP classifier\cite{das2020accelerating} used in QMO  & 88.00 & 100 & 100 &  109 \textcolor{blue}{AMP}$\rightarrow$109 \textcolor{blue}{AMP}\\
        iAMP-2L (AMP) \cite{xiao2013iamp} &  92.23 & 91.74 & 72.47 & 100 \textcolor{blue}{AMP}$\rightarrow$78 \textcolor{blue}{AMP} \\
        CAMP-RF (AMP) \cite{thomas2010camp} & 87.57 & 80.73 & 64.22 & 88 \textcolor{blue}{AMP}$\rightarrow$67 \textcolor{blue}{AMP} \\
        Witten E.Coli (AMP) \cite{witten2019deep} & 94.30 & 63.32 & 66.05 & 69 \textcolor{blue}{AMP}$\rightarrow$55 \textcolor{blue}{AMP} \\ 
        Witten S.aureus (AMP) \cite{witten2019deep} & 94.30 & 64.22 & 63.30 & 70 \textcolor{blue}{AMP}$\rightarrow$55 \textcolor{blue}{AMP} \\ 
        \hline
        Toxicity classifier\cite{das2020accelerating} used in QMO  & 93.7 & 100 & 0 & 109 \textcolor{red}{T} $\rightarrow$ 109 \textcolor{blue}{NT}\\
        HLPpred-Fuse (toxicity) \cite{hasan2020hlppred} & 97 & 61.46 & 44.03 & 67 \textcolor{red}{T} $\rightarrow$ 25 \textcolor{blue}{NT} \\ 
        HAPPENN (toxicity) \cite{timmons2020happenn} & 85.7   & 42.20 & 19.26 & 46 \textcolor{red}{T} $\rightarrow$ 31 \textcolor{blue}{NT} \\ 
        \hline
        iAMP-2L + HAPPENN (AMP + Non-toxic) & -- & 50.45 (AMP + Non-toxic) & 56.88 (AMP + Non-toxic) & 45 (AMP, \textcolor{red}{T}) $\rightarrow$ 22 (AMP, \textcolor{blue}{NT})\\
         \bottomrule
    \end{tabular}
    \caption{Reported accuracy, prediction rate, and property improvement for 109 pairs of known AMP  and QMO-optimized sequences 
    based on different AMP and Toxicity classifiers. The 109 starting sequences are experimentally verified toxic AMPs and are correctly predicted by the AMP and toxicity classifiers used in QMO.  
    The external classifiers have varying prediction accuracy as they may yield incorrect predictions  on some of starting sequences. The prediction rate on QMO-optimized sequences is defined as the fraction of AMP and/or toxin predictions.
    About 56.88\% of QMO-optimized sequences are predicted as non-toxic AMPs by iAMP-2L + HAPPENN, showing high agreement with the classifiers used in QMO.}
    \label{tab:ext-classifiers}
\end{table}

\begin{table}[t]
    \centering
    \begin{tabular}{l|rr}
    \toprule
    Property & Known AMP & QMO optimized Variant  \\ \hline
    Charge & $4.4873 \pm 2.7109$ & $4.1552 \pm 2.8018$ \\
    Charge Density & $0.0019 \pm 0.0011$ &  $0.0018 \pm 0.0012$ \\
    Aliphatic Index & $101.280 \pm 41.371$ & $92.955 \pm 42.996$ \\
    Aromaticity & $0.1265 \pm 0.0790$ & $0.1094 \pm 0.0879$ \\
    Hydrophobicity & $0.0132 \pm 0.3557$ & $-0.0289 \pm 0.3387$ \\
    Hydrophobic Moment & $0.4400 \pm 0.1936$ & $0.2509 \pm 0.1431$ \\
    Hydrophobic Ratio & $0.4436 \pm 0.1369$ & $0.4068 \pm 0.1423$ \\
    \bottomrule
    \end{tabular}
    \caption{Physicochemical properties, such as charge, charge density, aliphatic index, aromaticity, hydrophobicity, hydrophobic moment, hydrophobic ratio, instability index, estimated on 109 known AMPs and their QMO optimized variants.}
    \label{tab:bio_properties}
\end{table}

\clearpage
\section{Stability Analysis of QMO}
\label{appendix_stability}
In this section, we selected two tasks to study the stability of QMO, in terms of run time comparison and the effect on the number $Q$ of per-iteration random directions used in QMO.

\subsection{QED Optimization}

The total runtime for the QED task with $T=20$ iterations and 50 random restarts was approximately 487 CPU/GPU hours or, spread over 32 cores, 15.2 hours of wall time. We complete 15 random restarts at a success rate of 85.9\% in roughly 8 hours of wall time for 32 cores, the same time reported in \cite{you2018graph}. We ran all our experiments on machines with Intel Xeon E5-2600 CPUs and NVIDIA K80 GPUs.

\begin{figure}[h]
    \centering
    \begin{subfigure}[b]{0.48\textwidth}
        \includegraphics[width=\textwidth]{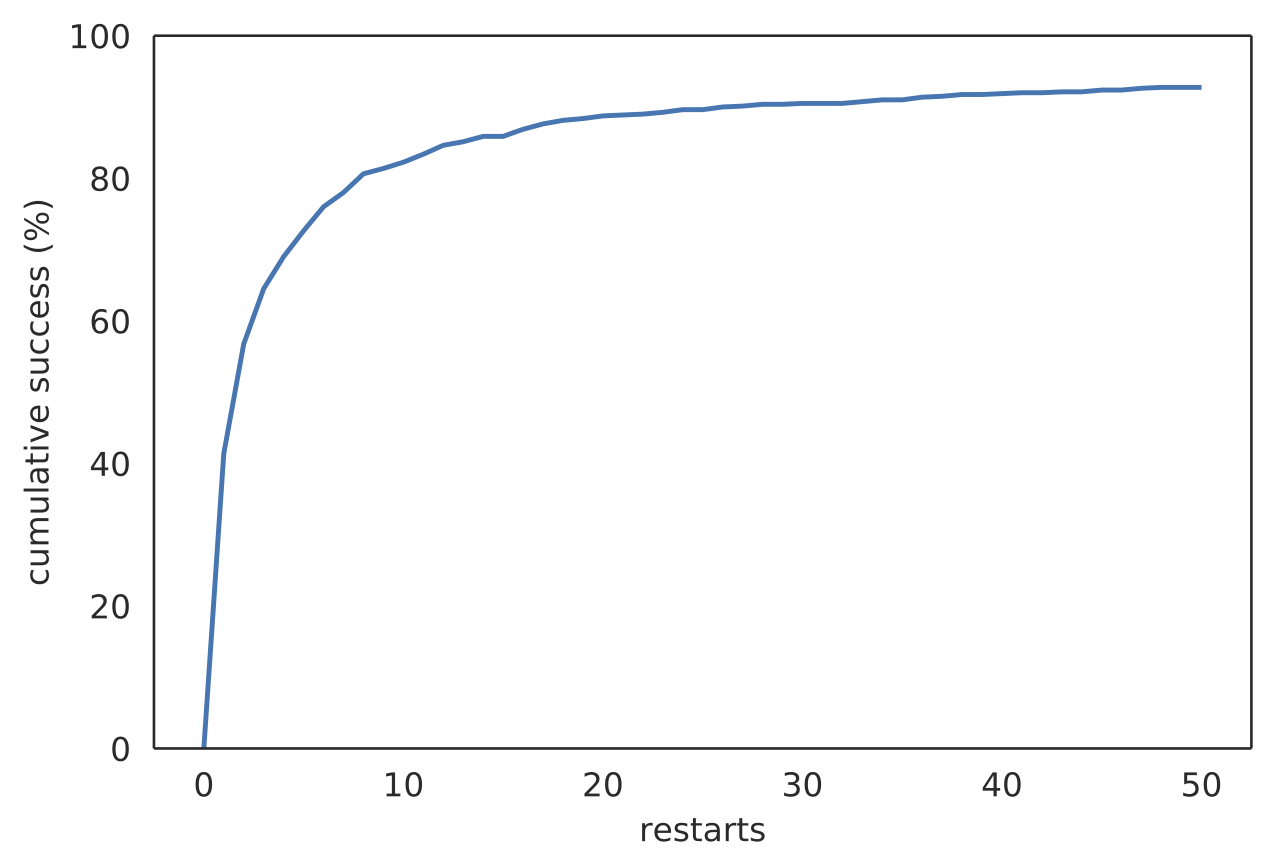}
        \caption{20 steps per restart (50x20).}
        \label{fig:QED_50x20}
    \end{subfigure}
    \begin{subfigure}[b]{0.48\textwidth}
        \includegraphics[width=\textwidth]{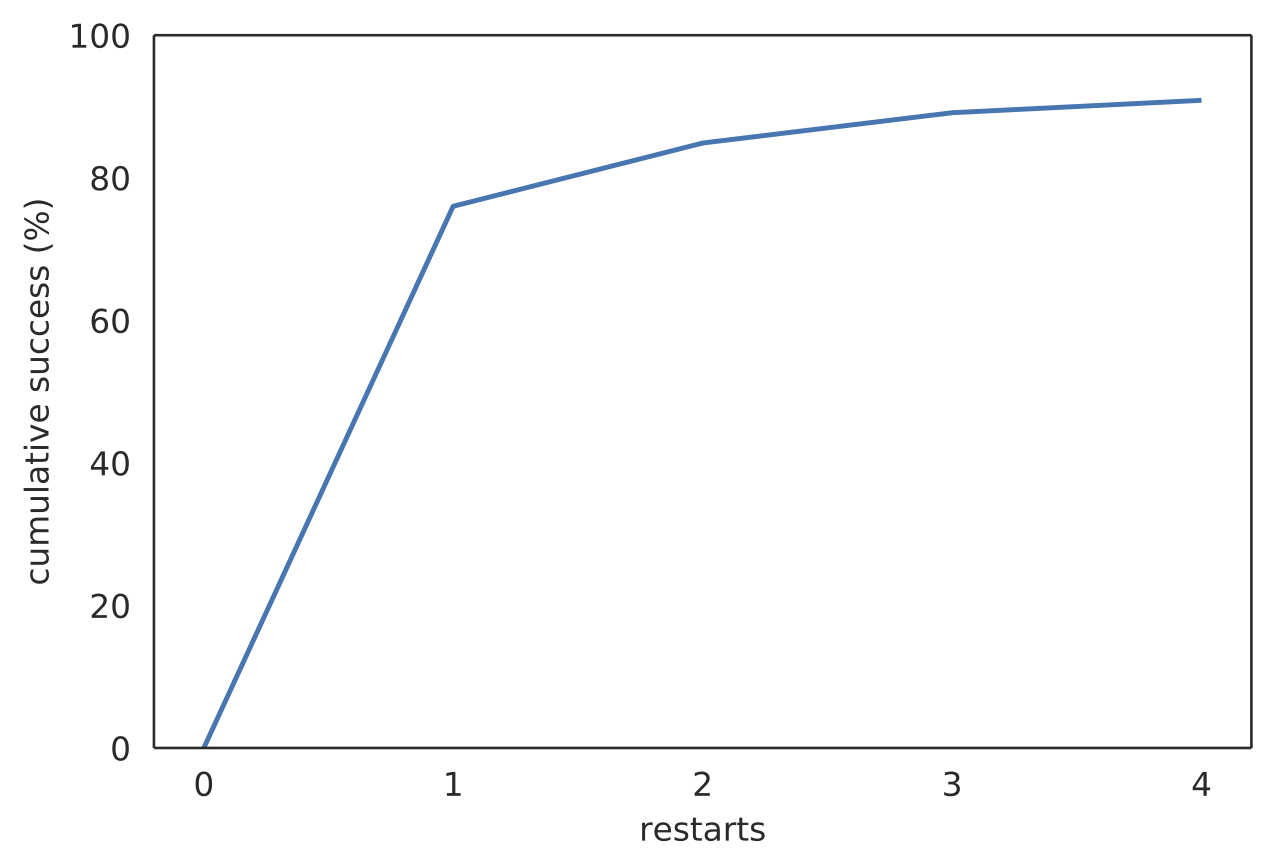}
        \caption{300 steps per restart (4x300).}
        \label{fig:QED_300x4}
    \end{subfigure}
    \caption{Cumulative success rate as a function of number of random restarts with 20 or 300 steps per restart for the QED task.}
    \label{fig:QED_success_curves}
\end{figure}

\begin{figure}[h]
    \centering
    \includegraphics[width=0.6\textwidth]{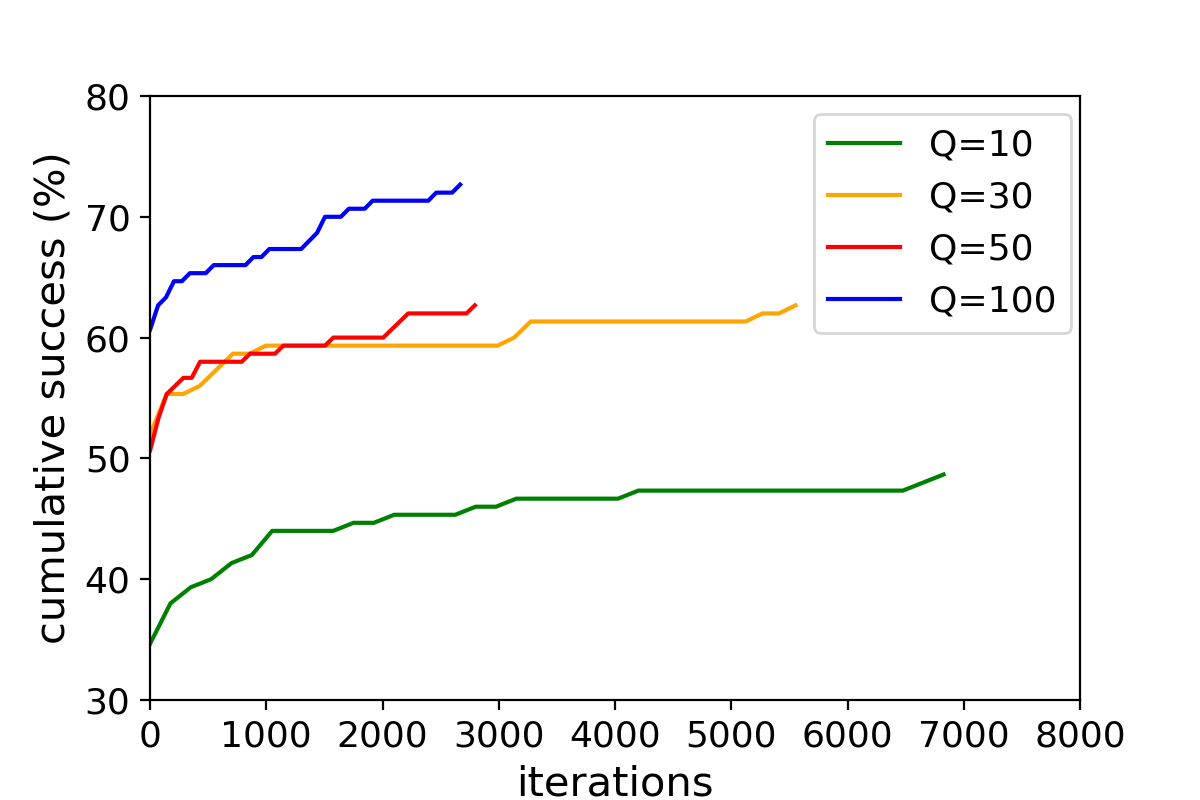}
    \caption{Cumulative success rate for the AMP task as a function of number of iterations with different per-iteration random sampling query value $Q$.}
    \label{fig:AMP_diff_Q}
\end{figure}

In examining the effect of multiple restarts, we show two configurations for running QMO on the QED task in Figure \ref{fig:QED_success_curves}. Restarts happen if the algorithm is unable to find a successful candidate after the allotted number of steps. For this task, the algorithm stops after the first successful candidate is found. After 50 restarts at 20 steps (Figure \ref{fig:QED_50x20}), we reach a success rate of 92.8\% whereas after 4 restarts at 300 steps (Figure \ref{fig:QED_300x4}), we only achieve 90.9\% success despite taking about as long --- 447 hours (4x300) compared to 487 hours (50x20). We conclude that for relatively easy tasks (single-constraint optimization) using a small number of steps $T$, restarting with a new random seed can be very effective since at early iterations the (effective) step size is relatively large and the guided search may tend to overshoot.

\subsection{AMP Optimization}

Figure \ref{fig:AMP_diff_Q} compares the effect of the per-iteration random direction sampling number $Q$ in QMO on the cumulative success of the AMP experiments. We observe that setting larger $Q$ value can improve the cumulative success rate but at the cost of increased number of property evaluations. 
The performance becomes similar once the $Q$ value is sufficiently large, suggesting the stability of QMO. 



\section{Additional Analysis on Diversity, Novelty, and Optimization Objectives}
\label{appen_diversity}

\subsection{Diversity and Novelty Metrics when Varying Similarity Constraints}
\label{sup_diversity}


\begin{table}[h]
\centering
\begin{tabular}{@{}l|cccccc@{}}
\toprule
Metric      & $\delta=0.8$  & $\delta=0.6$  & $\delta=0.4$  & $\delta=0.2$   & $\delta=0.1$   & $\delta=0.0$   \\ \midrule
Improvement & $0.82\pm1.29$ & $3.60\pm2.67$ & $7.49\pm3.29$ & $11.66\pm3.94$ & $14.66\pm4.50$ & $16.59\pm5.59$ \\
Int. Div.   & 0.87          & 0.86          & 0.86          & 0.84           & 0.81           & 0.78           \\
Unique@100  & 0.07          & 0.38          & 0.79          & 0.99           & 0.99           & 0.95           \\ 
\bottomrule
\end{tabular}
\caption{Performance metrics on the penalized logP improvement task on a random subset of 80 molecules using QMO when varying the Tanimoto similarity threshold $\delta$. Rows from top to bottom: penalized logP improvement (mean $\pm$ standard deviation), internal diversity (of the final, improved 80 molecules), average novelty (after running experiment 100 times with new random seeds and averaging over all 80 molecules).}
\label{tab:logP_sim_range}
\end{table}

Following the same experiment setup,
Table \ref{tab:logP_sim_range} shows the performance gains of QMO on the logP task on 80 randomly sampled lead molecules as the similarity constraint is relaxed.
Internal diversity is computed as the average Tanimoto similarity between all pairs of molecules in the set of 80 (1 run). The internal diversity of the original molecules is 0.89.
It is observed that the internal diversity decreases as the threshold decreases.  This is because the algorithm is incentivized to find more similar substructures which dominantly improve logP and this gets easier with a lower similarity threshold. On the other hand, after running QMO 100 times we find the fraction of unique final molecules for each initial point, averaged over all 80 initial molecules, increases dramatically as the similarity constraint is relaxed. In other words, QMO is able to generate novel molecules when given different random seeds for the same starting point if the search space is expanded.


\subsection{Different Optimization Objectives}
\label{sup_objective}


\begin{table}[h]
\centering
\begin{tabular}{@{}ll|cccccc@{}}
\toprule
Objective & & Similarity & QED & SA & \textcolor{blue}{Affinity} & logP & Int. Div \\ \midrule
\multirow{2}{*}{\textcolor{blue}{SA}} & original & \multirow{2}{*}{0.097} & 0.495 & 3.165 & 5.786 & 2.632 & 0.851 \\
   & improved &       & 0.613 & \textbf{1.108} & \textbf{7.556} & 3.423 & 0.473 \\ \midrule
\multirow{2}{*}{\textcolor{blue}{QED}} & original & \multirow{2}{*}{0.110} & 0.495 & 3.165 & 5.786 & 2.632 & 0.851 \\
    & improved &       & \textbf{0.946} & 3.439 & \textbf{7.905} & 2.835 & 0.823\\ \bottomrule
\end{tabular}
\caption{Average QMO performance starting from  23 SARS-CoV-2 inhibitor lead-molecules on M\textsuperscript{pro}-affinity $\ge7.5$, min-SA and max-QED objectives for the top and bottom rows respectively.}
\label{tab:mpro_qed_sa}
\end{table}

QMO works with a variety of optimization objectives, including those which do not require a similarity constraint. In these cases, the choice of starting sequence is less important since the optimizer is able to move arbitrarily far from the initial point (see Section \ref{sup_rand_z}). In Table \ref{tab:mpro_qed_sa}, we show the results on two tasks without similarity constraints: minimizing SA while maintaining binding affinity with M\textsuperscript{pro} $\ge7.5$ and maximizing QED while maintaining the same binding affinity threshold. In the SA minimization task, we note that the internal diversity of the improved molecules is low and we observe that the exact same final molecule is often found when starting from different initial points. In this case, the task is relatively simple and the optimizer is able to find the same minimum point regardless of where it starts. The QED maximization task shows more diverse results. For full results, see Tables \ref{tab:Mpro_QED} and \ref{tab:Mpro_SA}.

\subsection{Randomly Generated Lead Sequence}
\label{sup_rand_z}


\begin{table}[h]
\centering
\begin{tabular}{@{}ll|ccccc@{}}
\toprule
Objective & & Similarity & QED & SA & \textcolor{blue}{Affinity} & logP \\ \midrule
\multirow{2}{*}{\textcolor{blue}{SA}} & original & \multirow{2}{*}{0.081} & 0.574 & 3.752 & 6.034 & 1.882 \\
   & improved &       & 0.644 & \textbf{1.166} & \textbf{7.688} & 3.598 \\ \midrule
\multirow{2}{*}{\textcolor{blue}{QED}} & original & \multirow{2}{*}{0.096} & 0.574 & 3.752 & 6.034 & 1.882 \\
    & improved &       & \textbf{0.947} & 3.345 & \textbf{7.718} & 2.885 \\ \bottomrule
\end{tabular}
\caption{Average QMO performance starting from 23 randomly generated molecules on M\textsuperscript{pro}-affinity $\ge7.5$, min-SA and max-QED objectives for the top and bottom rows respectively.}
\label{tab:mpro_rand}
\end{table}

In addition to starting from known lead-molecules for SARS-CoV-2 inhibition, we also experimented with starting from molecules drawn uniformly randomly from the latent space. The same set of 23 random molecules are used to reproduce the experiments in Section \ref{sup_objective}. The summarized results in Table \ref{tab:mpro_rand} show that the utility of QMO is not restricted to just searching the space near a known lead molecule and can be used for random generation as well. QMO is able to consistently achieve similar results to Section \ref{sup_objective} without starting from a predetermined initial point. Full results can be found in Tables \ref{tab:Mpro_rand_SA} and \ref{tab:Mpro_rand_QED} below.


\section{Ablation Study of QMO}
\label{appen_ablation}

\subsection{The Effect of Encoder-Decoder}
\label{sup_encoder}


Here we compare QMO performance using
different $\beta$-VAE\cite{higgins2016beta} and the default WAE\cite{tolstikhin2017wasserstein} models studied in Das et al \cite{das2020accelerating} for the AMP task.
For $\beta$-VAE, we used three models by changing the Kullback–Leibler (KL) annealing  term $\beta$ from $0$ to $0.03$, $0$ to $0.3$, and $0$ to $1.0$. For WAE, we used random feature approximation of Gaussian kernel with kernel bandwidth $\sigma=7$ as this was found working to be the best model\cite{das2020accelerating}. Additionally, the latent space noise log variance regularization, $R(logVar)$, was imposed to prevent encoder from mode collapsing.  Table \ref{tab:vae_model_metrics} summarizes the performance of these models following Das et al\cite{das2020accelerating}.

\begin{table}[h]
\centering
\begin{tabular}{l|l|rrrr}
\toprule
\multicolumn{2}{c}{Architecture} & PPL & BLEU & Recon & Encoder variance  \\ \hline
\multicolumn{2}{l}{$\beta$-VAE (1.0)}                   &	3.82   &   3e-3   & 2.764 & -3e-4   \\ 
\multicolumn{2}{l}{$\beta$-VAE (0.3)}                  &	2.70   &   2e-3   & 2.765 & -4e-4  \\
\multicolumn{2}{l}{$\beta$-VAE (0.03)}                  &	15.14   &   0.499   & 1.070 & -0.608 \\
\cmidrule{1-6} 
\multicolumn{2}{l}{WAE, latent dim$=100$, $ \sigma=7$, $R(logVar)=1e-2$}      &	15.16 & 0.665  & 0.685 & -0.396 \\
\multicolumn{2}{l}{WAE (our-model), latent dim$=100$, $\sigma=7$, $R(logVar)=1e-3$}          & 12.75 & 0.824  & 0.195 &  -4.015  \\ 
\multicolumn{2}{l}{WAE, latent dim$=50$, $\sigma=7$, $R(logVar)=1e-3$}          & 14.31 & 0.713  & 0.623 &  -6.28 \\ 
\multicolumn{2}{l}{WAE, latent dim$=120$, $\sigma=7$, $R(logVar)=1e-3$}          & 12.72 & 0.900  & 0.188 &  -3.45 \\ 

 \bottomrule
\end{tabular}
\caption{Performance of various autoencoder schemes on the AMP dataset\cite{das2020accelerating}.  PPL means perplexity (closer to 13.26 is desired as discussed in  Das et.al \cite{das2020accelerating}), BLEU (bilingual evaluation understudy) measures the similarity between original and reconstructed sequences (higher is better), Recon means the reconstruction error (lower is better). Large negative values of encoder variance indicate that the encoder is collapsed to become deterministic. } 
\label{tab:vae_model_metrics}
\end{table}

Figure \ref{fig:bvae_wae_cmp} shows the QMO success rate plot for these model variants. All QMO hyperparamters are fixed among all set of experiments with only change of encoder-decoder model. It is shown that QMO performance increases  with an improved underlying autoencoder model. The cumulative success rate aligns well with the performance metric evaluations reported in Table \ref{tab:vae_model_metrics}. Our default WAE model performs better than variants of $\beta$-VAE as it has lower reconstruction error and higher BLEU score. The results show that a better encoder-decoder can lead to improved QMO performance. We also note that setting a larger latent dimension may improve the performance of the encoder-decoder at the price of requiring more queries per iteration for balancing gradient estimation.  

\begin{table}[t]
\begin{minipage}[t]{.49\textwidth}
    \centering
    \includegraphics[width=1\textwidth]{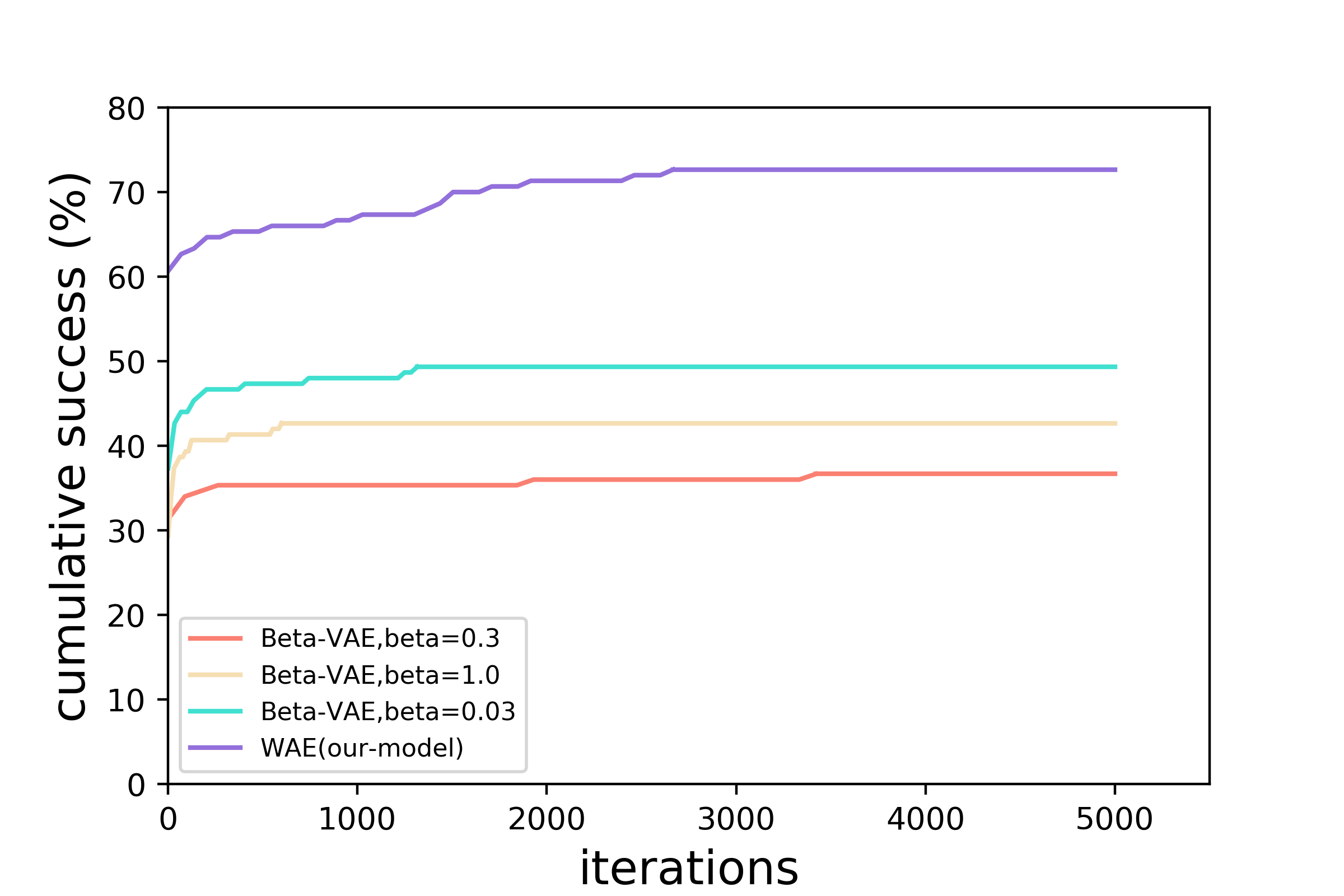}
    \captionof{figure}{Cumulative success rate for the AMP task as a function of number of iterations with different autoencoders.}
    \label{fig:bvae_wae_cmp}
\end{minipage}
\hfill
\begin{minipage}[t]{.49\textwidth}
    \centering
    \includegraphics[width=1\textwidth]{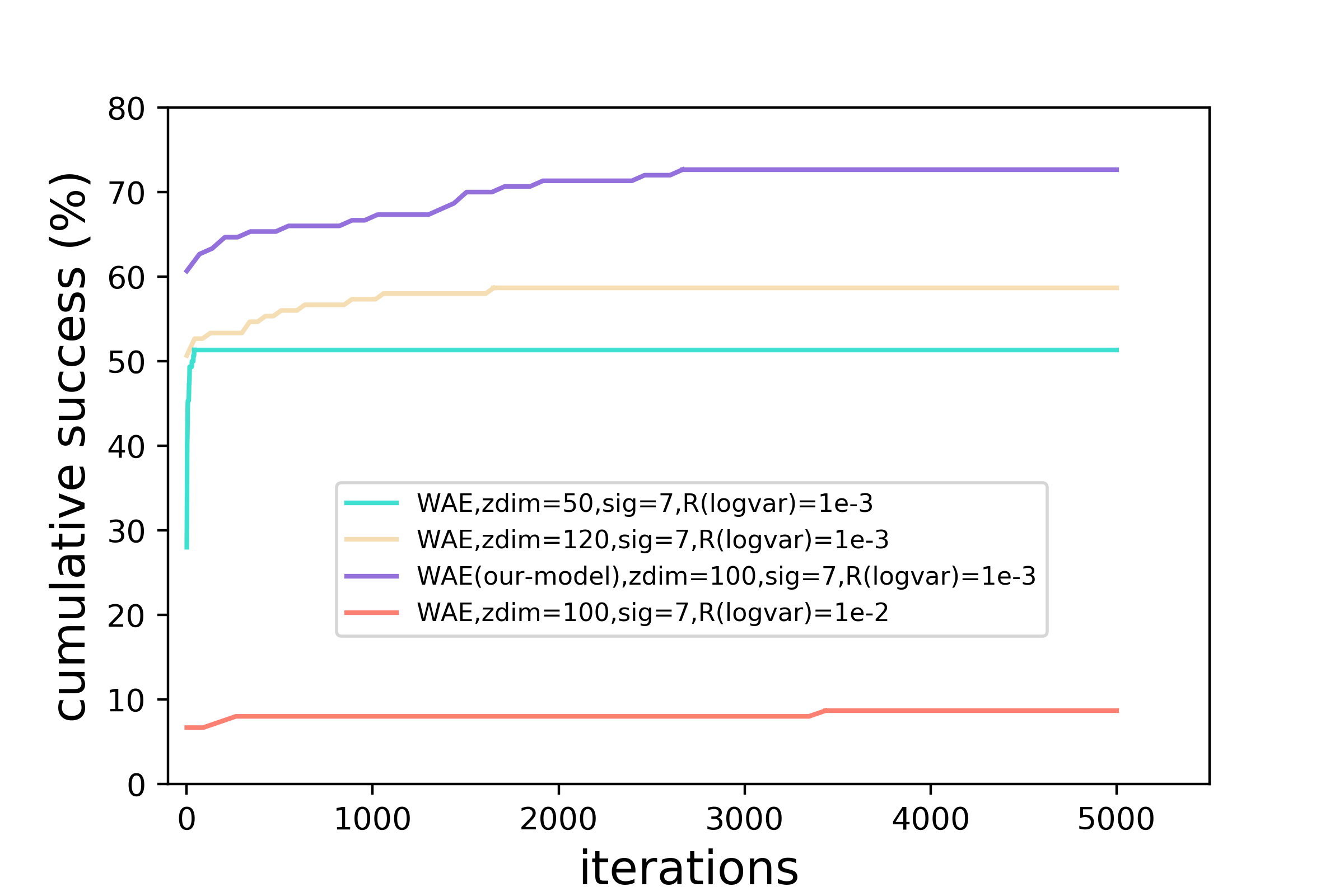}
     \captionof{figure}{Cumulative success rate for the AMP task as a function of number of iterations with WAE model variants.}
    \label{fig:zdim_wae_cmp}
\end{minipage}
\end{table}


\subsection{The Effect of Latent Dimension in Encoder-Decoder}
\label{sup_latent_dim}

Given the same number of queries per iteration, Figure \ref{fig:clf_wae_cmp} compares the QMO performance on the AMP task using WAE models with different latent dimensions as listed in Table \ref{tab:vae_model_metrics}.
One can observe that models having low BLEU scores and large reconstruction errors (e.g., setting the latent dimension to be 50 or changing log variance $R(logVar) = 1e-2$) lead to low success rate due to poor molecule representation learning of the encoder-decoder. Increasing the latent dimension may slightly improve the encoder-decoder performance, but it may not improve the QMO performance since more queries per iteration are needed for zeroth-order gradient descent to reach the same level of gradient estimation accuracy\cite{liu2020primer}.


\subsection{Sequence v.s. Latent Space Classifier}
\label{sup_classifier}

We test the impact of switching from sequence-level property predictors to latent-space (z-space) predictors on QMO optimization for AMP task. For the latent-space classifier, we use logistic regression with inverse regularization strength $C = 1.0$ and 200 Limited-memory BFGS (L-BFGS) optimization iterations to train on the latent representations from the same underlying encoder-decoder. Datasets for AMP and Toxicity attributes are taken from Das et al \cite{das2020accelerating}. Test accuracies of the latent-space classifiers are $80.27\%$ for AMP and $87.65\%$ for toxicity.

\begin{figure}[t]
    \centering
    \includegraphics[width=0.5\textwidth]{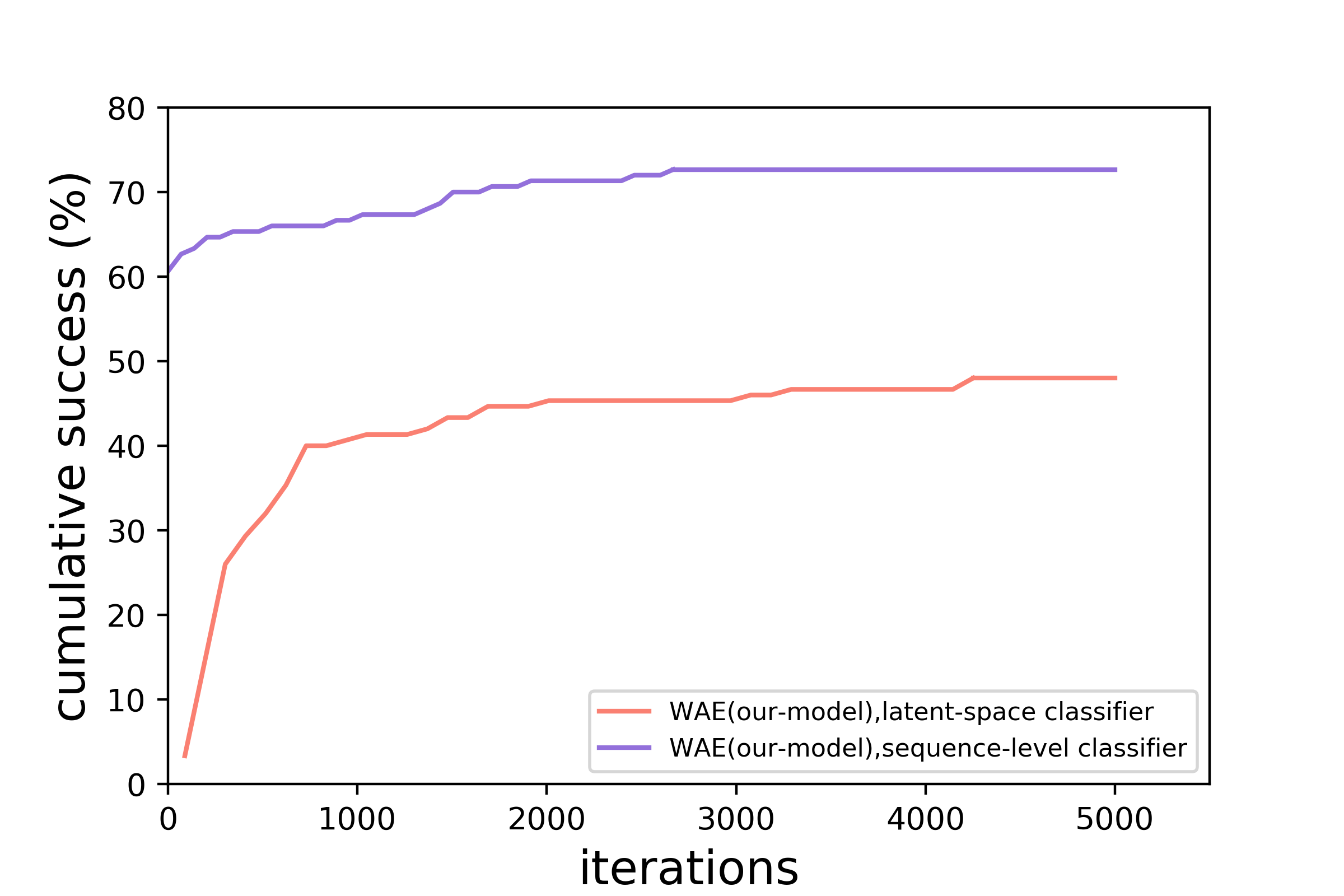}
    \caption{Cumulative success rate for the AMP task as a function of number of iterations with different classifiers (sequence-level v.s. latent-space).}
    \label{fig:clf_wae_cmp}
\end{figure}

Figure \ref{fig:clf_wae_cmp} shows the performance of QMO using sequence-level and latent-space classifiers. All hyperparameter settings are the same except that different classifiers are used for property prediction.
We found that QMO with sequence-level classifiers performs better. The final success rate of sequence-level classifiers is $109/150$ while that of latent-space classifiers is $92/150$. Moreover, in the latter case 20 out of 
92 lead sequences get improved candidate same as original ones, which attributes to  latent-space classifiers mis-classifying those lead sequences. Excluding the initially mis-classified sequences, the success rate of QMO using latent-space classifiers comes down to $72/150$. The reason about the performance difference can be explained from the fact that
since latent representations are information-compressed proxies of molecules, using latent-space classifiers can give worse results for guided optimization when compared to using sequence-level classifiers.

\subsection{The effect of Search Radius in QMO}
\label{sup_radius}


\begin{table}[h]
\begin{minipage}[t]{.49\textwidth}
    \centering
    \includegraphics[width=0.8\textwidth]{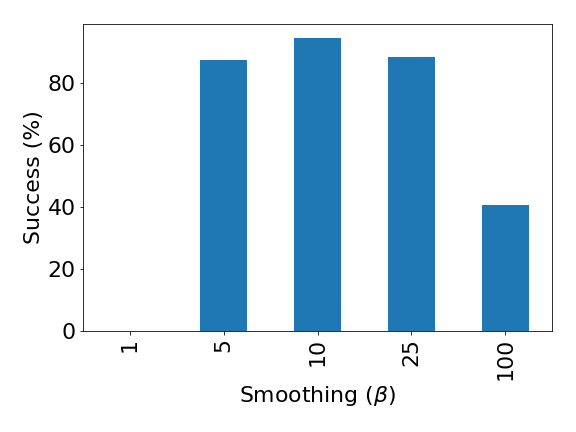}
    \captionof{figure}{Success rate on drug likeness (QED) task vs.\ $\beta$ value in QMO for the QED optimization task. }
    \label{fig:QED_beta}
\end{minipage}
\hfill
\begin{minipage}[t]{.49\textwidth}
\vspace{-40mm}
\centering
\begin{tabular}[h]{l|l}
\toprule
Method
 & Success (\%) \\ \midrule
 QMO-SPSA \cite{} & 80.1 \\ \hline
  \textbf{QMO} & \textbf{88.4} \\ 
 \bottomrule
\end{tabular}
\caption{Performance of drug likeness (QED) task with Tanimoto similarity constraint $\delta=0.4$. QMO-SPSA means solving the QMO loss function using the simultaneous perturbation stochastic approximation (SPSA) method\cite{spall1998overview}.}
\label{tab:ZOO}
\end{minipage}
\end{table}


In order to demonstrate the effect of the search radius/smoothing parameter in the QMO algorithm, $\beta$, we show results from the QED task with varying $\beta$ values in Figure \ref{fig:QED_beta}. At $\beta=1$ and below, QMO was unable to improve QED above the threshold for any molecules. This is because the search radius is too small to approximate an accurate gradient so the search is essentially random. At $\beta=100$ and above, all of the unsuccessful molecules were due to the decoder model returning invalid molecules. In this case, the search radius is too large and the resultant $z$ points are at the edges of the latent space. Between these extremes, the effect is milder and we found $\beta=10$ to work well in many cases with this decoder model.

\subsection{The Effect of Zeroth-Order Optimizer}
\label{sup_SPSA}



Table \ref{tab:ZOO} compares the results on the QED optimization task between our QMO zeroth-order optimizer and the simultaneous perturbation stochastic approximation (SPSA) method \cite{spall1998overview}. QMO-SPSA is identical to Algorithm \ref{algo_QMO_adam} except the gradient estimation step is replaced with SPSA: $\nablahat \loss(z^{(t)}) = \frac{\loss(z^{(t)}+\beta \cdot u) - \loss(z^{(t)}-\beta \cdot u)}{2 \cdot \beta \cdot u}$. QMO shows a significant (10.4\%) increase in success rate over this baseline while requiring an equivalent number of estimator queries.

For SPSA, we use a fixed perturbation coefficient, equivalent to $\beta$. In this case, we run QMO with $Q=10$ and the other parameters the same as in the main paper. We also run SPSA with $(Q+1)/2$ times as many steps for parity in the number of estimator queries. Hence, we use $T=110$ for SPSA.

 \subsection{QMO v.s. MSO}
 \label{sup_MSO}

\begin{figure}[h]
 \centering
    \includegraphics[width=0.4\textwidth]{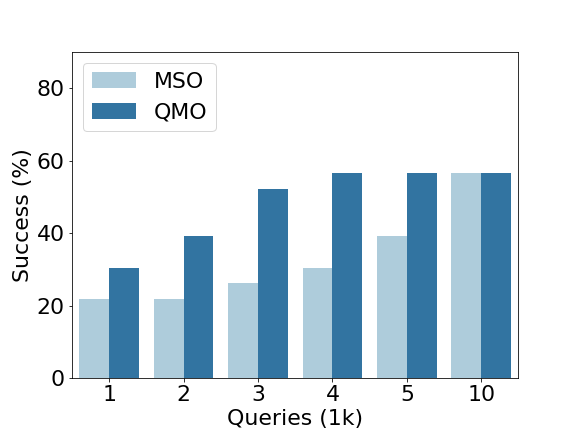}
    \captionof{figure}{Comparison of molecular optimization methods at equivalent numbers of loss function queries on a M\textsuperscript{pro}-affinity $\ge 7.5$ and similarity $\ge 0.6$ task. We use the same 23 lead-molecules from Section \ref{appendix_covid}.}
    \label{fig:mso_queries}
\end{figure}

Figure \ref{fig:mso_queries} demonstrates one advantage of our algorithm over the molecule swarm optimization (MSO)\cite{winter2019efficient} in terms of query efficiency: QMO has a higher success rate with low numbers of queries. In other words, QMO is able to find success more quickly than MSO. The task has two thresholds --- binding affinity with SARS-CoV-2 M\textsuperscript{pro}~$\ge 7.5$ and Tanimoto similarity to initial molecule~$\ge 0.6$ --- and the algorithm is successful as long as both criteria are met. For this task, we limit the number of queries allowed to the respective optimization objective function at intervals of 1000. To be specific, for MSO we use binding affinity weight $=5$, particles $=50$, and steps $=[19, 39, 59, 79, 99, 199]$ and for QMO we use $\alpha_0=0.02$, $\beta=10$, $Q=9$, $\lambda=1$, and $T=[100, 200, 300, 400, 500, 1000]$.

\subsection{Summary}
The key insights concluded from the corresponding QMO results in Section \ref{appen_ablation} are (i) encoder-decoders with lower
reconstruction errors lead to better optimization performance; (ii) using sequence-level classifier gives a higher success rate than latent-space classifier; (iii) setting the search radius too small or too large degrades the optimization performance; and (iv) QMO outperforms other gradient-free optimizers including particle swarm optimization and simultaneous perturbation
stochastic approximation.

\clearpage

\begin{center}
\begin{longtable}{ll|p{0.2\textwidth}ccccc}
\toprule
compound & & Molecule & \textcolor{blue}{Affinity} & Similarity & \textcolor{blue}{QED} & SA & logP \\\endfirsthead
\multicolumn{6}{l}{\bfseries \tablename\ \thetable{} continued}\\[10pt]
compound & & Molecule & \textcolor{blue}{Affinity} & Similarity & \textcolor{blue}{QED} & SA & logP \\\endhead
\midrule
\multirow{2}{*}[-0.35in]{Ambroxol} & improved & \parbox[c]{0.125\textwidth}{\includegraphics[width=0.125\textwidth]{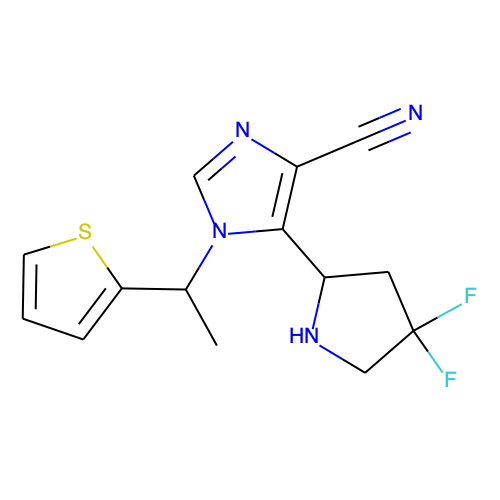}} &      7.55 &   \multirow{2}{*}[-0.35in]{0.05} &  0.95 &  4.23 &  3.10 \\*
          & original & \parbox[c]{0.125\textwidth}{\includegraphics[width=0.125\textwidth]{SVGs/Ambroxol_original.png}} &      6.46 &   &  0.71 &  2.50 &  3.19 \\
\hline
\multirow{2}{*}[-0.35in]{Bromhexine} & improved & \parbox[c]{0.125\textwidth}{\includegraphics[width=0.125\textwidth]{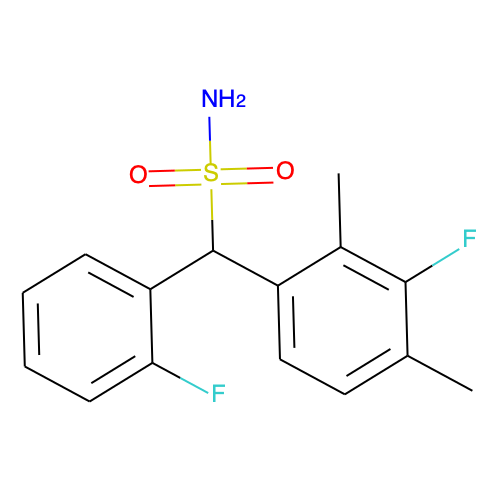}} &      8.18 &   \multirow{2}{*}[-0.35in]{0.06} &  0.95 &  3.05 &  2.96 \\*
          & original & \parbox[c]{0.125\textwidth}{\includegraphics[width=0.125\textwidth]{SVGs/Bromhexine_original.png}} &      5.00 &   &  0.78 &  2.38 &  4.56 \\
\hline
\multirow{2}{*}[-0.35in]{Chloroquine} & improved & \parbox[c]{0.125\textwidth}{\includegraphics[width=0.125\textwidth]{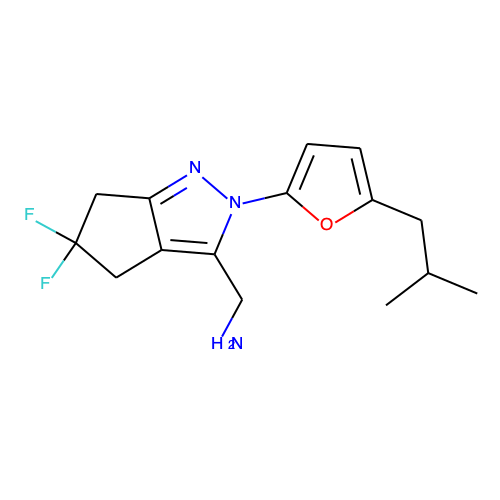}} &      7.90 &   \multirow{2}{*}[-0.35in]{0.11} &  0.94 &  3.63 &  2.86 \\*
          & original & \parbox[c]{0.125\textwidth}{\includegraphics[width=0.125\textwidth]{SVGs/Chloroquine_original.png}} &      6.07 &    &  0.76 &  2.67 &  4.81 \\
\hline
\multirow{2}{*}[-0.35in]{Cinanserin} & improved & \parbox[c]{0.125\textwidth}{\includegraphics[width=0.125\textwidth]{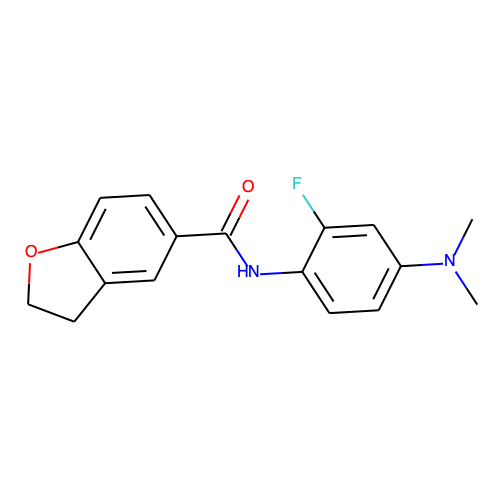}} &      8.01 &   \multirow{2}{*}[-0.35in]{0.16} &  0.95 &  2.03 &  3.08 \\*
          & original & \parbox[c]{0.125\textwidth}{\includegraphics[width=0.125\textwidth]{SVGs/Cinanserin_original.png}} &      4.43 &    &  0.44 &  2.07 &  4.38 \\
\hline
\multirow{2}{*}[-0.35in]{Curcumin} & improved & \parbox[c]{0.125\textwidth}{\includegraphics[width=0.125\textwidth]{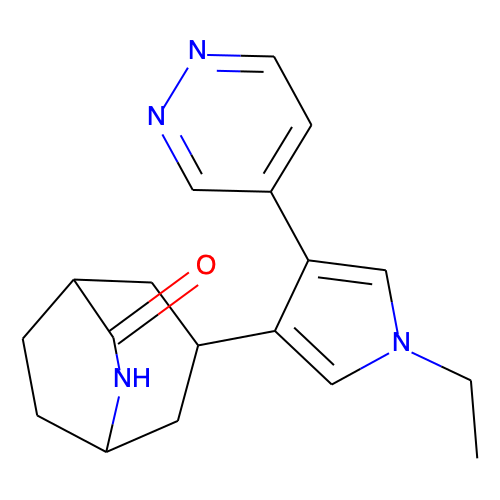}} &      8.58 &   \multirow{2}{*}[-0.35in]{0.10} &  0.95 &  4.89 &  2.74 \\*
          & original & \parbox[c]{0.125\textwidth}{\includegraphics[width=0.125\textwidth]{SVGs/Curcumin_original.png}} &      6.86 &    &  0.55 &  2.43 &  3.37 \\
\hline
\multirow{2}{*}[-0.35in]{Dipyridamole} & improved & \parbox[c]{0.125\textwidth}{\includegraphics[width=0.125\textwidth]{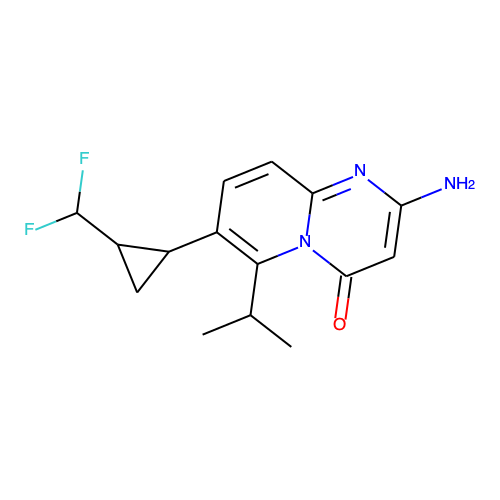}} &      7.94 &   \multirow{2}{*}[-0.35in]{0.07} &  0.95 &  4.00 &  2.77 \\*
          & original & \parbox[c]{0.125\textwidth}{\includegraphics[width=0.125\textwidth]{SVGs/Dipyridamole_original.png}} &      3.94 &    &  0.31 &  2.99 & -0.02 \\
\hline
\multirow{2}{*}[-0.35in]{Disulfiram} & improved & \parbox[c]{0.125\textwidth}{\includegraphics[width=0.125\textwidth]{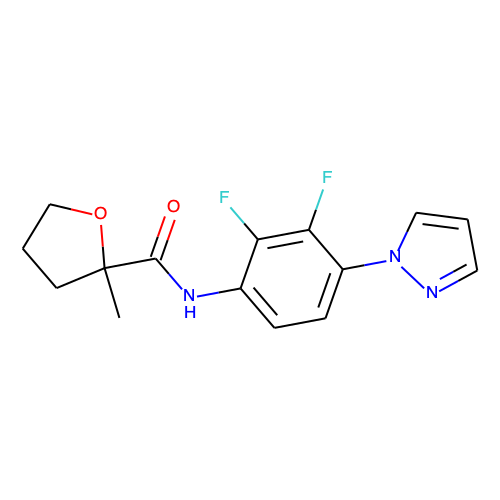}} &      8.01 &   \multirow{2}{*}[-0.35in]{0.03} &  0.95 &  3.20 &  2.66 \\*
          & original & \parbox[c]{0.125\textwidth}{\includegraphics[width=0.125\textwidth]{SVGs/Disulfiram_original.png}} &      5.54 &    &  0.57 &  3.12 &  3.62 \\
\hline
\multirow{2}{*}[-0.35in]{Ebselen} & improved & \parbox[c]{0.125\textwidth}{\includegraphics[width=0.125\textwidth]{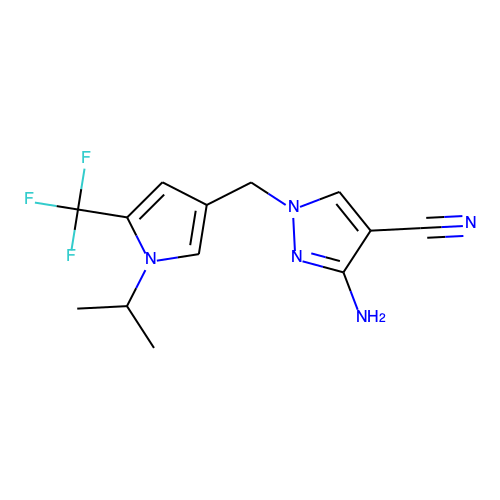}} &      7.95 &   \multirow{2}{*}[-0.35in]{0.05} &  0.95 &  3.22 &  2.79 \\*
          & original & \parbox[c]{0.125\textwidth}{\includegraphics[width=0.125\textwidth]{SVGs/Ebselen_original.png}} &      5.09 &    &  0.63 &  2.05 &  3.05 \\
\hline
\multirow{2}{*}[-0.35in]{Entecavir} & improved & \parbox[c]{0.125\textwidth}{\includegraphics[width=0.125\textwidth]{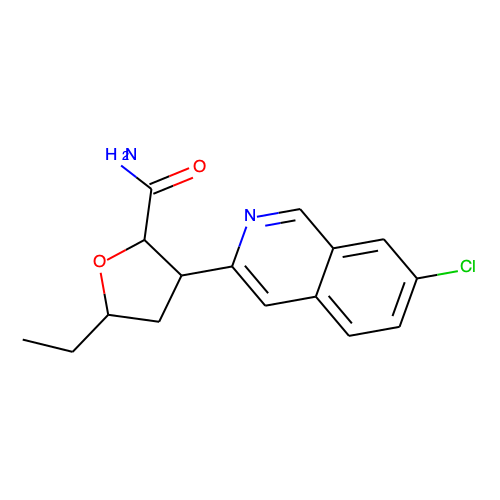}} &      7.57 &   \multirow{2}{*}[-0.35in]{0.15} &  0.95 &  3.71 &  3.02 \\*
          & original & \parbox[c]{0.125\textwidth}{\includegraphics[width=0.125\textwidth]{SVGs/Entecavir_original.png}} &      5.55 &    &  0.53 &  4.09 & -0.83 \\
\hline
\multirow{2}{*}[-0.35in]{Favipiravir} & improved & \parbox[c]{0.125\textwidth}{\includegraphics[width=0.125\textwidth]{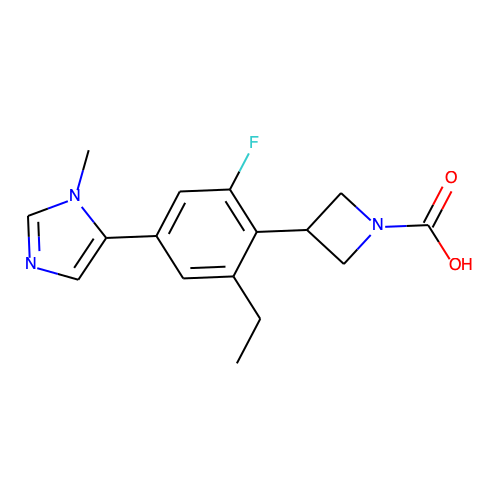}} &      8.23 &   \multirow{2}{*}[-0.35in]{0.12} &  0.95 &  2.83 &  2.87 \\*
          & original & \parbox[c]{0.125\textwidth}{\includegraphics[width=0.125\textwidth]{SVGs/Favipiravir_original.png}} &      4.32 &    &  0.55 &  2.90 & -0.99 \\
\hline
\multirow{2}{*}[-0.35in]{GS-441524} & improved & \parbox[c]{0.125\textwidth}{\includegraphics[width=0.125\textwidth]{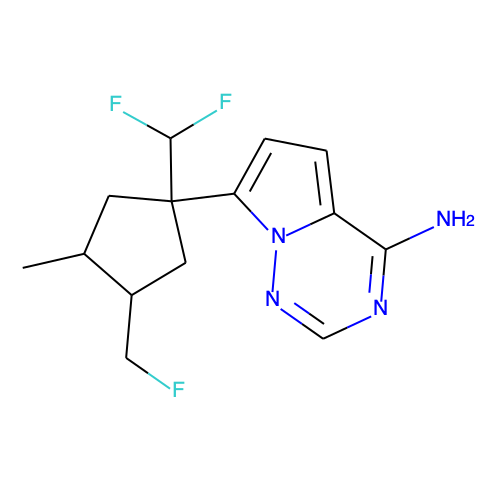}} &      7.52 &   \multirow{2}{*}[-0.35in]{0.37} &  0.95 &  4.54 &  2.83 \\*
          & original & \parbox[c]{0.125\textwidth}{\includegraphics[width=0.125\textwidth]{SVGs/GS-441524_original.png}} &      6.56 &    &  0.50 &  4.38 & -1.86 \\
\hline
\multirow{2}{*}[-0.35in]{Hydroxychloroquine} & improved & \parbox[c]{0.125\textwidth}{\includegraphics[width=0.125\textwidth]{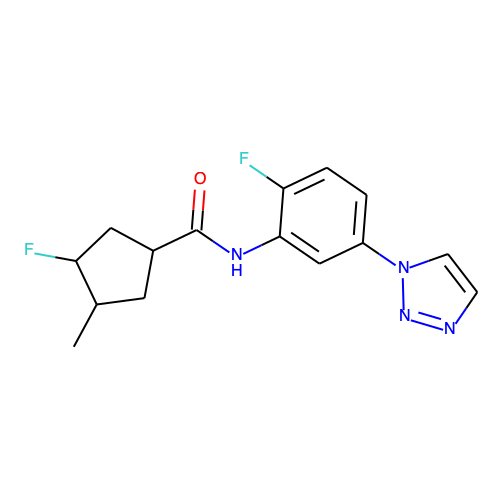}} &      7.71 &   \multirow{2}{*}[-0.35in]{0.13} &  0.95 &  3.67 &  2.73 \\*
          & original & \parbox[c]{0.125\textwidth}{\includegraphics[width=0.125\textwidth]{SVGs/Hydroxychloroquine_original.png}} &      5.85 &    &  0.73 &  2.79 &  3.78 \\
\hline
\multirow{2}{*}[-0.35in]{Kaempferol} & improved & \parbox[c]{0.125\textwidth}{\includegraphics[width=0.125\textwidth]{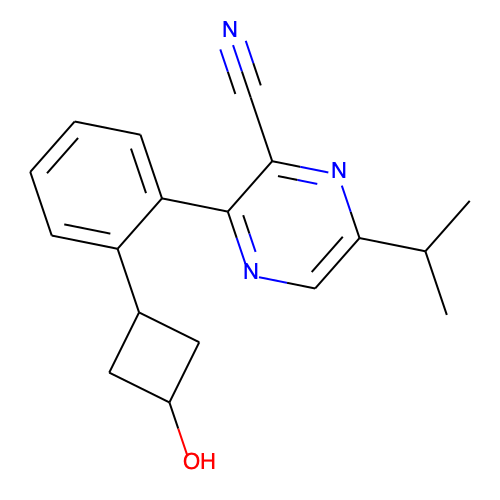}} &      7.96 &   \multirow{2}{*}[-0.35in]{0.07} &  0.94 &  2.81 &  3.38 \\*
          & original & \parbox[c]{0.125\textwidth}{\includegraphics[width=0.125\textwidth]{SVGs/Kaempferol_original.png}} &      6.90 &    &  0.55 &  2.37 &  2.28 \\
\hline
\multirow{2}{*}[-0.35in]{Lopinavir} & improved & \parbox[c]{0.125\textwidth}{\includegraphics[width=0.125\textwidth]{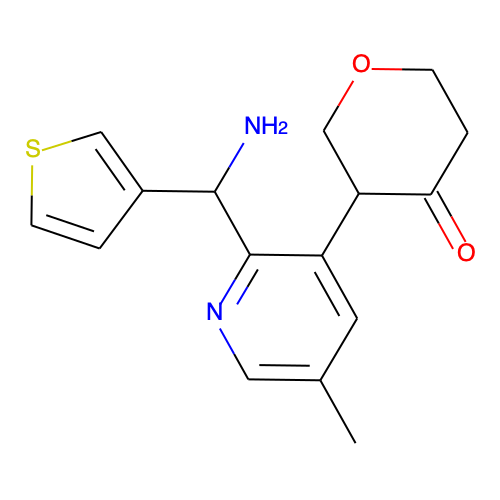}} &      7.75 &   \multirow{2}{*}[-0.35in]{0.10} &  0.95 &  3.86 &  2.57 \\*
          & original & \parbox[c]{0.125\textwidth}{\includegraphics[width=0.125\textwidth]{SVGs/Lopinavir_original.png}} &      6.42 &    &  0.20 &  3.90 &  4.33 \\
\hline
\multirow{2}{*}[-0.35in]{N3} & improved & \parbox[c]{0.125\textwidth}{\includegraphics[width=0.125\textwidth]{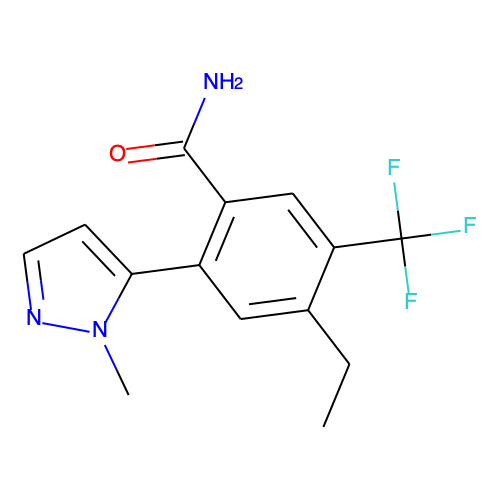}} &      7.68 &   \multirow{2}{*}[-0.35in]{0.12} &  0.95 &  2.66 &  2.77 \\*
          & original & \parbox[c]{0.125\textwidth}{\includegraphics[width=0.125\textwidth]{SVGs/N3_original.png}} &      6.83 &    &  0.12 &  4.29 &  2.08 \\
\hline
\multirow{2}{*}[-0.35in]{Nelfinavir} & improved & \parbox[c]{0.125\textwidth}{\includegraphics[width=0.125\textwidth]{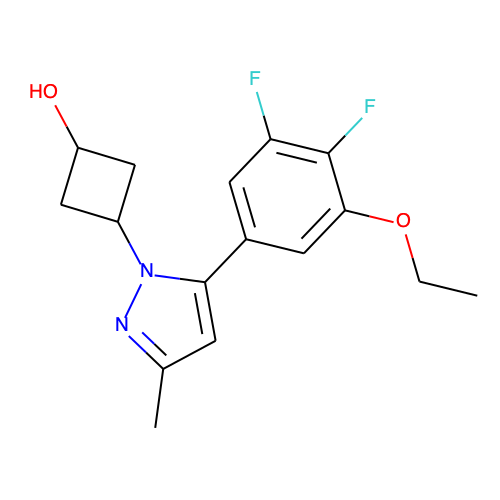}} &      7.93 &   \multirow{2}{*}[-0.35in]{0.10} &  0.94 &  2.63 &  3.23 \\*
          & original & \parbox[c]{0.125\textwidth}{\includegraphics[width=0.125\textwidth]{SVGs/Nelfinavir_original.png}} &      6.63 &    &  0.33 &  4.04 &  4.75 \\
\hline
\multirow{2}{*}[-0.35in]{O6K} & improved & \parbox[c]{0.125\textwidth}{\includegraphics[width=0.125\textwidth]{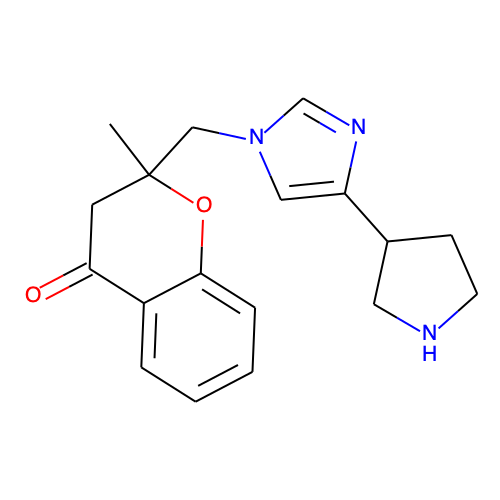}} &      7.57 &   \multirow{2}{*}[-0.35in]{0.14} &  0.95 &  3.77 &  2.38 \\*
          & original & \parbox[c]{0.125\textwidth}{\includegraphics[width=0.125\textwidth]{SVGs/O6K_original.png}} &      6.18 &    &  0.25 &  4.12 &  2.22 \\
\hline
\multirow{2}{*}[-0.35in]{PX-12} & improved & \parbox[c]{0.125\textwidth}{\includegraphics[width=0.125\textwidth]{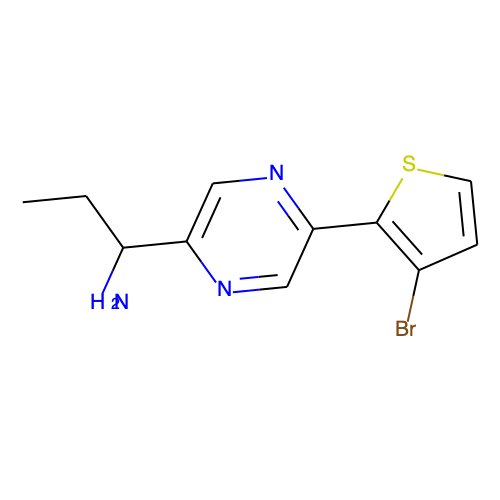}} &      7.90 &   \multirow{2}{*}[-0.35in]{0.16} &  0.94 &  3.39 &  3.38 \\*
          & original & \parbox[c]{0.125\textwidth}{\includegraphics[width=0.125\textwidth]{SVGs/PX-12_original.png}} &      5.01 &    &  0.74 &  3.98 &  2.95 \\
\hline
\multirow{2}{*}[-0.35in]{Quercetin} & improved & \parbox[c]{0.125\textwidth}{\includegraphics[width=0.125\textwidth]{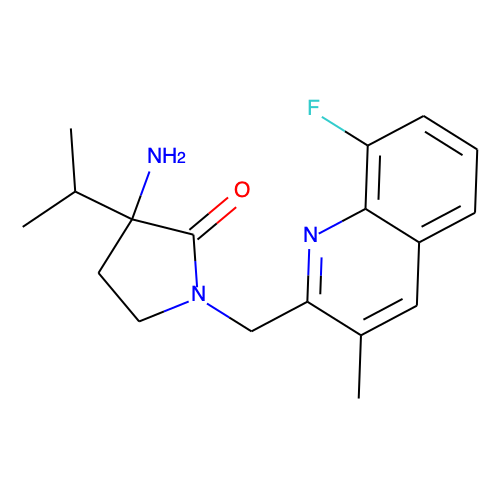}} &      8.10 &   \multirow{2}{*}[-0.35in]{0.08} &  0.95 &  3.44 &  2.77 \\*
          & original & \parbox[c]{0.125\textwidth}{\includegraphics[width=0.125\textwidth]{SVGs/Quercetin_original.png}} &      6.70 &    &  0.43 &  2.54 &  1.99 \\
\hline
\multirow{2}{*}[-0.35in]{Remdesivir} & improved & \parbox[c]{0.125\textwidth}{\includegraphics[width=0.125\textwidth]{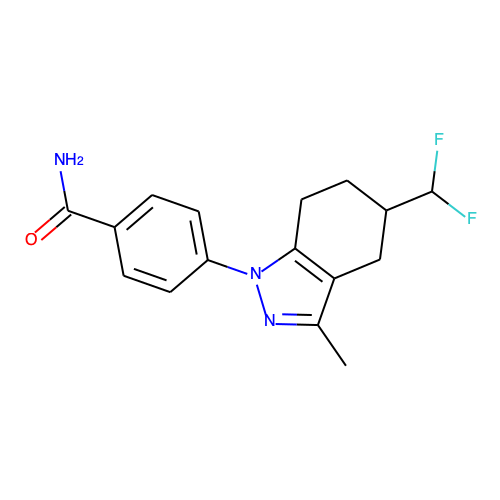}} &      8.00 &   \multirow{2}{*}[-0.35in]{0.13} &  0.95 &  3.08 &  2.65 \\*
          & original & \parbox[c]{0.125\textwidth}{\includegraphics[width=0.125\textwidth]{SVGs/Remdesivir_original.png}} &      6.32 &    &  0.16 &  4.82 &  2.31 \\
\hline
\multirow{2}{*}[-0.35in]{Shikonin} & improved & \parbox[c]{0.125\textwidth}{\includegraphics[width=0.125\textwidth]{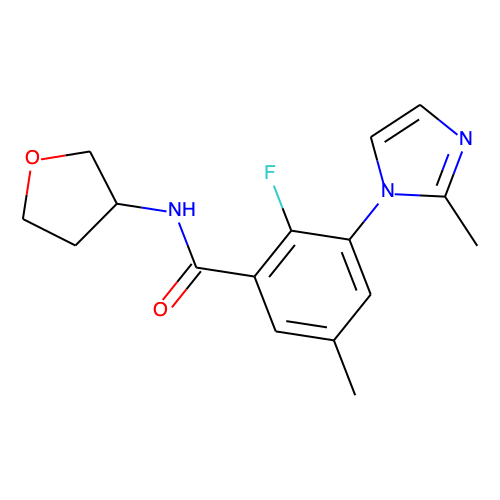}} &      7.84 &   \multirow{2}{*}[-0.35in]{0.06} &  0.94 &  3.04 &  2.15 \\*
          & original & \parbox[c]{0.125\textwidth}{\includegraphics[width=0.125\textwidth]{SVGs/Shikonin_original.png}} &      5.11 &    &  0.58 &  3.41 &  2.12 \\
\hline
\multirow{2}{*}[-0.35in]{Tideglusib} & improved & \parbox[c]{0.125\textwidth}{\includegraphics[width=0.125\textwidth]{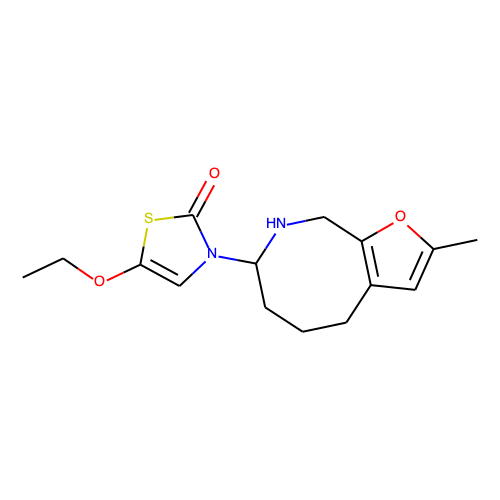}} &      8.21 &   \multirow{2}{*}[-0.35in]{0.10} &  0.95 &  4.12 &  2.83 \\*
          & original & \parbox[c]{0.125\textwidth}{\includegraphics[width=0.125\textwidth]{SVGs/Tideglusib_original.png}} &      4.94 &    &  0.58 &  2.28 &  3.26 \\
\hline
\multirow{2}{*}[-0.35in]{Umifenovir} & improved & \parbox[c]{0.125\textwidth}{\includegraphics[width=0.125\textwidth]{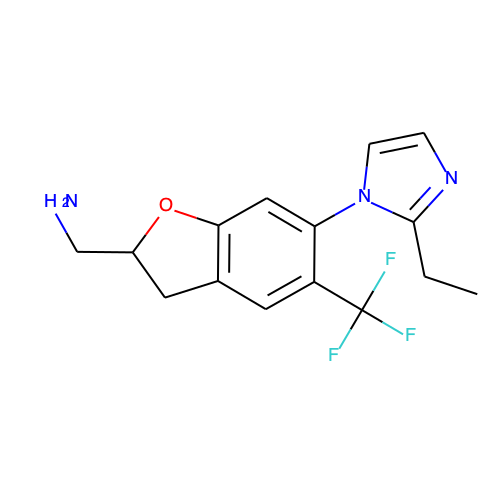}} &      7.74 &   \multirow{2}{*}[-0.35in]{0.08} &  0.95 &  3.29 &  2.72 \\*
          & original & \parbox[c]{0.125\textwidth}{\includegraphics[width=0.125\textwidth]{SVGs/Umifenovir_original.png}} &      6.36 &    &  0.38 &  2.68 &  5.18 \\
\hline
\multirow{2}{*}{Average} & improved & & \textbf{7.91} & \multirow{2}{*}{0.11} & \textbf{0.95} & 3.44 & 2.84 \\
 & original & & 5.79 & & 0.50 & 3.17 & 2.63 \\
\bottomrule
\caption{Results of QMO optimization starting from 23 SARS-CoV-2 inhibitor lead-molecules with M\textsuperscript{pro}-affinity $\ge7.5$, max-QED objective.}
\label{tab:Mpro_QED}
\end{longtable}
\end{center}

\newpage

\begin{center}
\begin{longtable}{ll|p{0.2\textwidth}ccccc}
\toprule
compound & & Molecule & \textcolor{blue}{Affinity} & Similarity & QED & \textcolor{blue}{SA} & logP \\\endfirsthead
\multicolumn{6}{l}{\bfseries \tablename\ \thetable{} continued}\\[10pt]
compound & & Molecule & \textcolor{blue}{Affinity} & Similarity & QED & \textcolor{blue}{SA} & logP \\\endhead
\midrule
\multirow{2}{*}[-0.35in]{Ambroxol} & improved &            \parbox[c]{0.125\textwidth}{\includegraphics[width=0.125\textwidth]{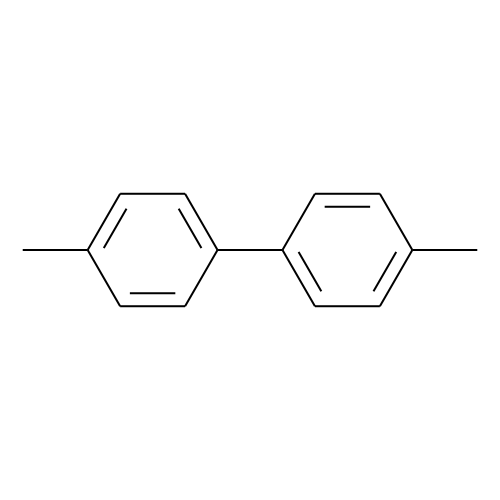}} &      8.12 &  \multirow{2}{*}[-0.35in]{0.05} & 0.63 & 1.00 &  3.97 \\*
           & original &               \parbox[c]{0.125\textwidth}{\includegraphics[width=0.125\textwidth]{SVGs/Ambroxol_original.png}} &      6.46 &                                 & 0.71 & 2.50 &  3.19 \\
\hline
\multirow{2}{*}[-0.35in]{Bromhexine} & improved &          \parbox[c]{0.125\textwidth}{\includegraphics[width=0.125\textwidth]{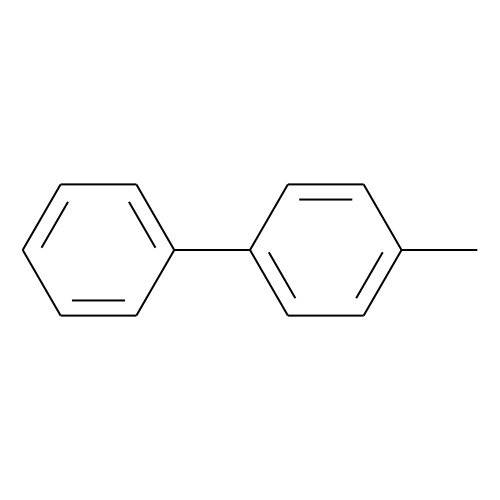}} &      7.51 &  \multirow{2}{*}[-0.35in]{0.07} & 0.61 & 1.07 &  3.66 \\*
           & original &             \parbox[c]{0.125\textwidth}{\includegraphics[width=0.125\textwidth]{SVGs/Bromhexine_original.png}} &      5.00 &                                 & 0.78 & 2.38 &  4.56 \\
\hline
\multirow{2}{*}[-0.35in]{Chloroquine} & improved &         \parbox[c]{0.125\textwidth}{\includegraphics[width=0.125\textwidth]{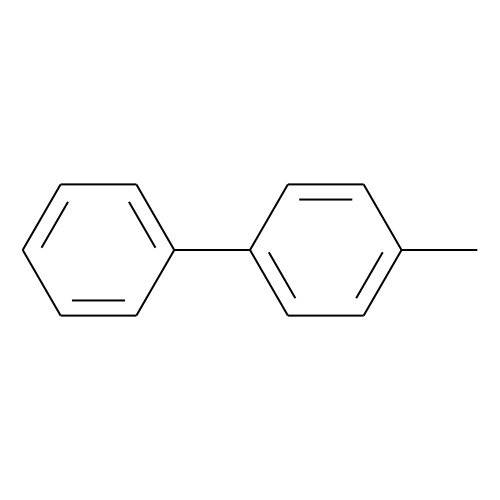}} &      7.51 &  \multirow{2}{*}[-0.35in]{0.07} & 0.61 & 1.07 &  3.66 \\*
           & original &            \parbox[c]{0.125\textwidth}{\includegraphics[width=0.125\textwidth]{SVGs/Chloroquine_original.png}} &      6.07 &                                 & 0.76 & 2.67 &  4.81 \\
\hline
\multirow{2}{*}[-0.35in]{Cinanserin} & improved &          \parbox[c]{0.125\textwidth}{\includegraphics[width=0.125\textwidth]{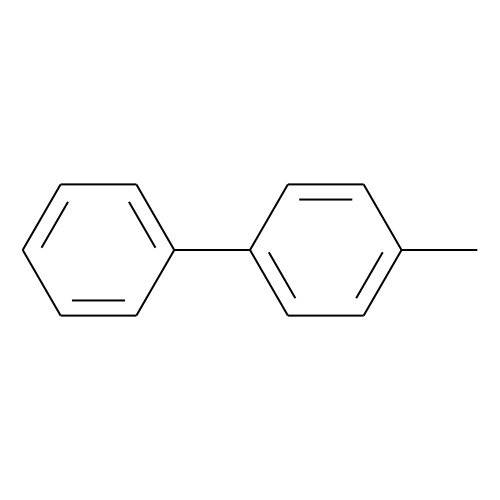}} &      7.51 &  \multirow{2}{*}[-0.35in]{0.14} & 0.61 & 1.07 &  3.66 \\*
           & original &             \parbox[c]{0.125\textwidth}{\includegraphics[width=0.125\textwidth]{SVGs/Cinanserin_original.png}} &      4.43 &                                 & 0.44 & 2.07 &  4.38 \\
\hline
\multirow{2}{*}[-0.35in]{Curcumin} & improved &            \parbox[c]{0.125\textwidth}{\includegraphics[width=0.125\textwidth]{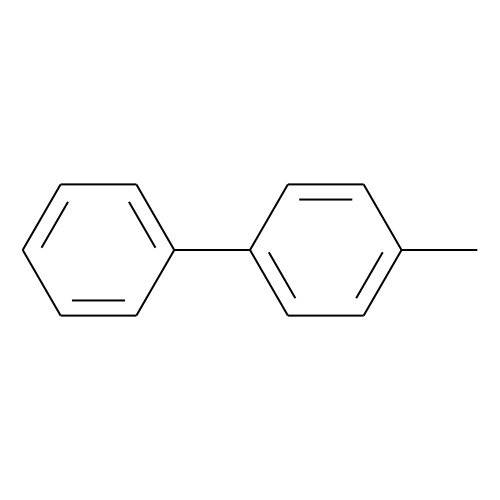}} &      7.51 &  \multirow{2}{*}[-0.35in]{0.10} & 0.61 & 1.07 &  3.66 \\*
           & original &               \parbox[c]{0.125\textwidth}{\includegraphics[width=0.125\textwidth]{SVGs/Curcumin_original.png}} &      6.86 &                                 & 0.55 & 2.43 &  3.37 \\
\hline
\multirow{2}{*}[-0.35in]{Dipyridamole} & improved &        \parbox[c]{0.125\textwidth}{\includegraphics[width=0.125\textwidth]{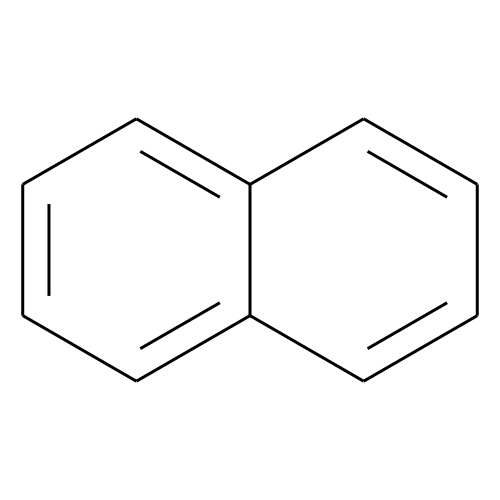}} &      8.16 &  \multirow{2}{*}[-0.35in]{0.06} & 0.51 & 1.00 &  2.84 \\*
           & original &           \parbox[c]{0.125\textwidth}{\includegraphics[width=0.125\textwidth]{SVGs/Dipyridamole_original.png}} &      3.94 &                                 & 0.31 & 2.99 & -0.02 \\
\hline
\multirow{2}{*}[-0.35in]{Disulfiram} & improved &          \parbox[c]{0.125\textwidth}{\includegraphics[width=0.125\textwidth]{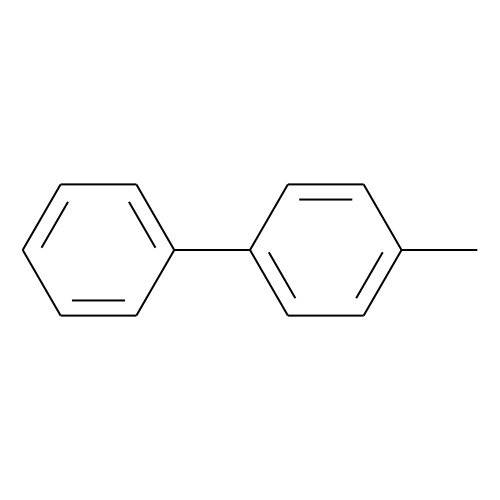}} &      7.51 &  \multirow{2}{*}[-0.35in]{0.03} & 0.61 & 1.07 &  3.66 \\*
           & original &             \parbox[c]{0.125\textwidth}{\includegraphics[width=0.125\textwidth]{SVGs/Disulfiram_original.png}} &      5.54 &                                 & 0.57 & 3.12 &  3.62 \\
\hline
\multirow{2}{*}[-0.35in]{Ebselen} & improved &           \parbox[c]{0.125\textwidth}{\includegraphics[width=0.125\textwidth]{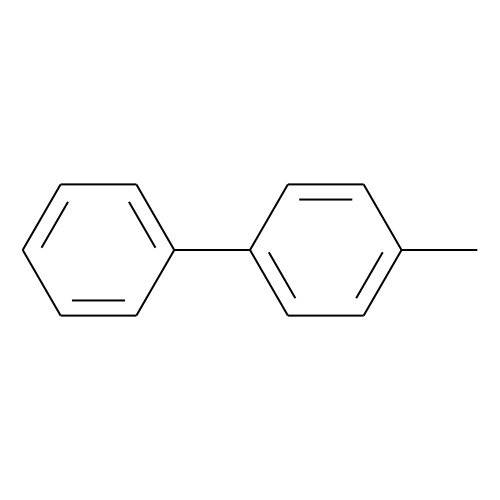}} &      7.51 &  \multirow{2}{*}[-0.35in]{0.18} & 0.61 & 1.07 &  3.66 \\*
           & original &              \parbox[c]{0.125\textwidth}{\includegraphics[width=0.125\textwidth]{SVGs/Ebselen_original.png}} &      5.09 &                                 & 0.63 & 2.05 &  3.05 \\
\hline
\multirow{2}{*}[-0.35in]{Entecavir} & improved &           \parbox[c]{0.125\textwidth}{\includegraphics[width=0.125\textwidth]{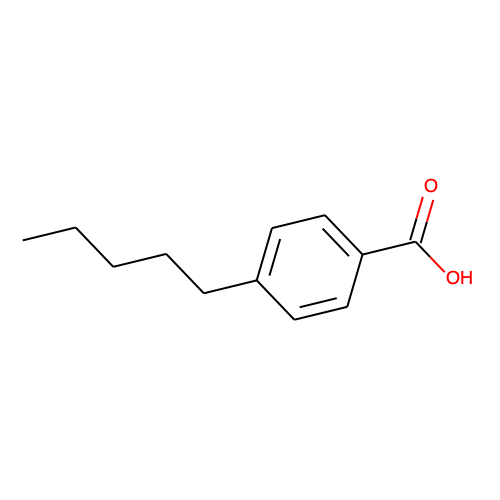}} &      7.63 &  \multirow{2}{*}[-0.35in]{0.07} & 0.73 & 1.38 &  3.12 \\*
           & original &              \parbox[c]{0.125\textwidth}{\includegraphics[width=0.125\textwidth]{SVGs/Entecavir_original.png}} &      5.55 &                                 & 0.53 & 4.09 & -0.83 \\
\hline
\multirow{2}{*}[-0.35in]{Favipiravir} & improved &         \parbox[c]{0.125\textwidth}{\includegraphics[width=0.125\textwidth]{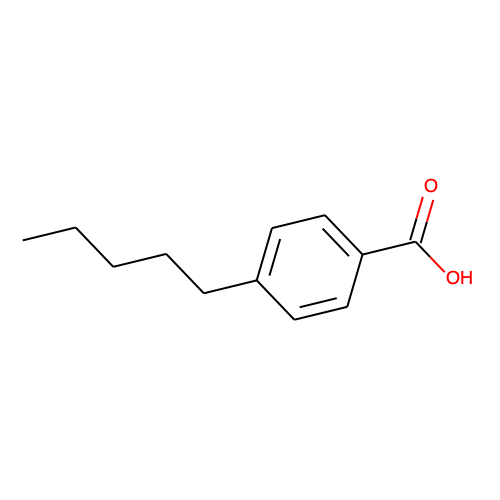}} &      7.63 &  \multirow{2}{*}[-0.35in]{0.11} & 0.73 & 1.38 &  3.12 \\*
           & original &            \parbox[c]{0.125\textwidth}{\includegraphics[width=0.125\textwidth]{SVGs/Favipiravir_original.png}} &      4.32 &                                 & 0.55 & 2.90 & -0.99 \\
\hline
\multirow{2}{*}[-0.35in]{GS-441524} & improved &           \parbox[c]{0.125\textwidth}{\includegraphics[width=0.125\textwidth]{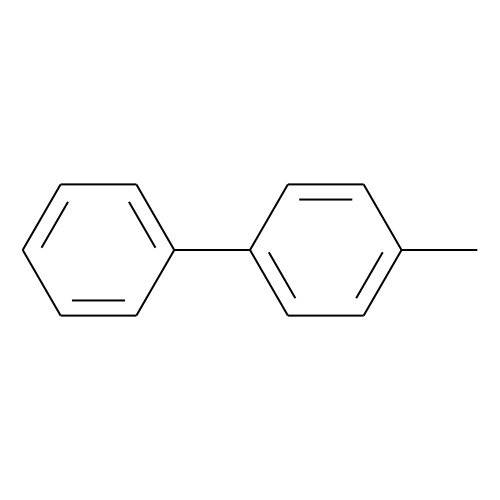}} &      7.51 &  \multirow{2}{*}[-0.35in]{0.05} & 0.61 & 1.07 &  3.66 \\*
           & original &              \parbox[c]{0.125\textwidth}{\includegraphics[width=0.125\textwidth]{SVGs/GS-441524_original.png}} &      6.56 &                                 & 0.50 & 4.38 & -1.86 \\
\hline
\multirow{2}{*}[-0.35in]{Hydroxychloroquine} & improved &  \parbox[c]{0.125\textwidth}{\includegraphics[width=0.125\textwidth]{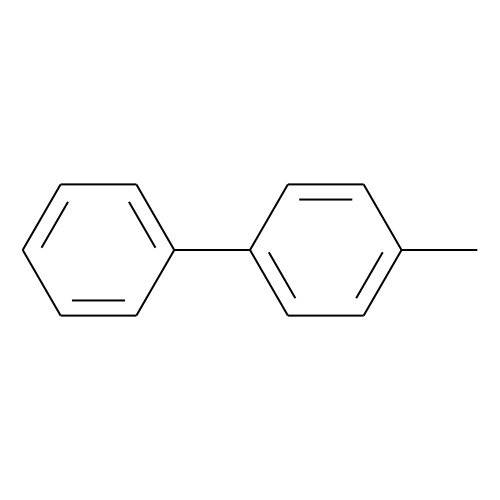}} &      7.51 &  \multirow{2}{*}[-0.35in]{0.07} & 0.61 & 1.07 &  3.66 \\*
           & original &     \parbox[c]{0.125\textwidth}{\includegraphics[width=0.125\textwidth]{SVGs/Hydroxychloroquine_original.png}} &      5.85 &                                 & 0.73 & 2.79 &  3.78 \\
\hline
\multirow{2}{*}[-0.35in]{Kaempferol} & improved &          \parbox[c]{0.125\textwidth}{\includegraphics[width=0.125\textwidth]{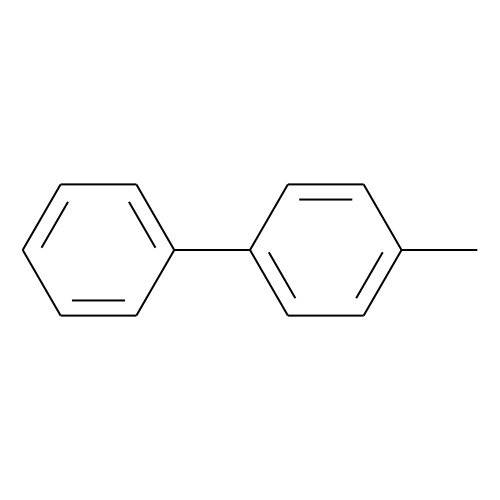}} &      7.51 &  \multirow{2}{*}[-0.35in]{0.12} & 0.61 & 1.07 &  3.66 \\*
           & original &             \parbox[c]{0.125\textwidth}{\includegraphics[width=0.125\textwidth]{SVGs/Kaempferol_original.png}} &      6.90 &                                 & 0.55 & 2.37 &  2.28 \\
\hline
\multirow{2}{*}[-0.35in]{Lopinavir} & improved &           \parbox[c]{0.125\textwidth}{\includegraphics[width=0.125\textwidth]{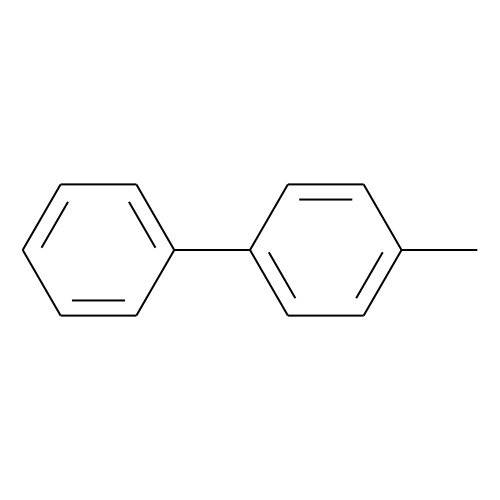}} &      7.51 &  \multirow{2}{*}[-0.35in]{0.11} & 0.61 & 1.07 &  3.66 \\*
           & original &              \parbox[c]{0.125\textwidth}{\includegraphics[width=0.125\textwidth]{SVGs/Lopinavir_original.png}} &      6.42 &                                 & 0.20 & 3.90 &  4.33 \\
\hline
\multirow{2}{*}[-0.35in]{N3} & improved &                  \parbox[c]{0.125\textwidth}{\includegraphics[width=0.125\textwidth]{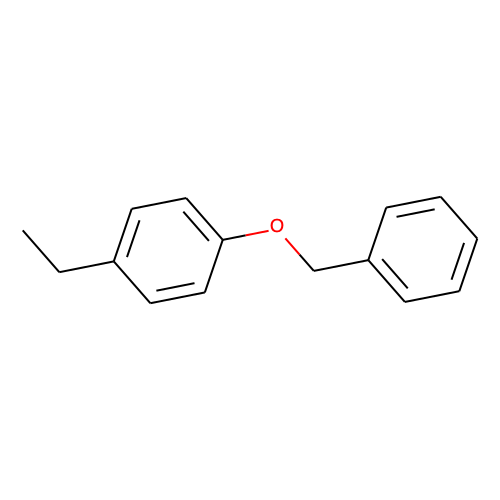}} &      8.21 &  \multirow{2}{*}[-0.35in]{0.14} & 0.75 & 1.25 &  3.83 \\*
           & original &                     \parbox[c]{0.125\textwidth}{\includegraphics[width=0.125\textwidth]{SVGs/N3_original.png}} &      6.83 &                                 & 0.12 & 4.29 &  2.08 \\
\hline
\multirow{2}{*}[-0.35in]{Nelfinavir} & improved &          \parbox[c]{0.125\textwidth}{\includegraphics[width=0.125\textwidth]{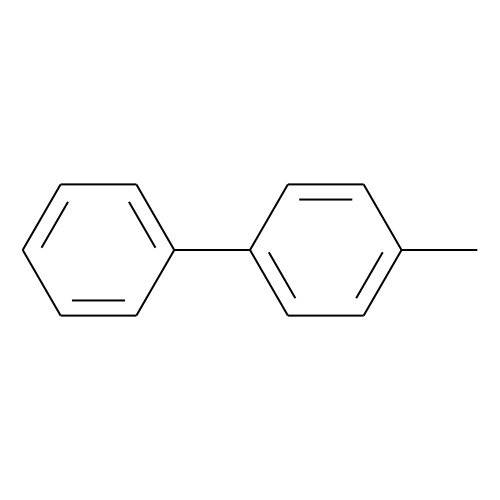}} &      7.51 &  \multirow{2}{*}[-0.35in]{0.11} & 0.61 & 1.07 &  3.66 \\*
           & original &             \parbox[c]{0.125\textwidth}{\includegraphics[width=0.125\textwidth]{SVGs/Nelfinavir_original.png}} &      6.63 &                                 & 0.33 & 4.04 &  4.75 \\
\hline
\multirow{2}{*}[-0.35in]{O6K} & improved &                 \parbox[c]{0.125\textwidth}{\includegraphics[width=0.125\textwidth]{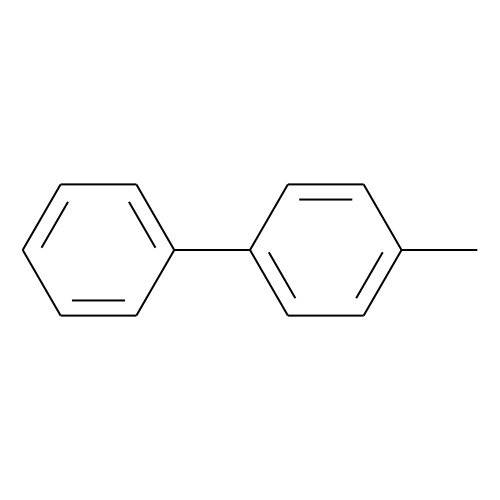}} &      7.51 &  \multirow{2}{*}[-0.35in]{0.09} & 0.61 & 1.07 &  3.66 \\*
           & original &                    \parbox[c]{0.125\textwidth}{\includegraphics[width=0.125\textwidth]{SVGs/O6K_original.png}} &      6.18 &                                 & 0.25 & 4.12 &  2.22 \\
\hline
\multirow{2}{*}[-0.35in]{PX-12} & improved &               \parbox[c]{0.125\textwidth}{\includegraphics[width=0.125\textwidth]{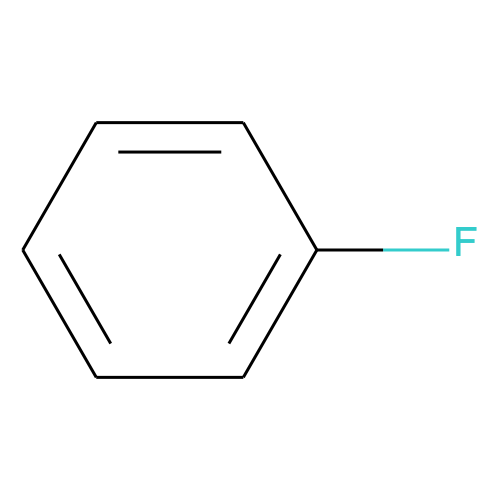}} &      3.90 &  \multirow{2}{*}[-0.35in]{0.05} & 0.46 & 1.00 &  1.83 \\*
           & original &                  \parbox[c]{0.125\textwidth}{\includegraphics[width=0.125\textwidth]{SVGs/PX-12_original.png}} &      5.01 &                                 & 0.74 & 3.98 &  2.95 \\
\hline
\multirow{2}{*}[-0.35in]{Quercetin} & improved &           \parbox[c]{0.125\textwidth}{\includegraphics[width=0.125\textwidth]{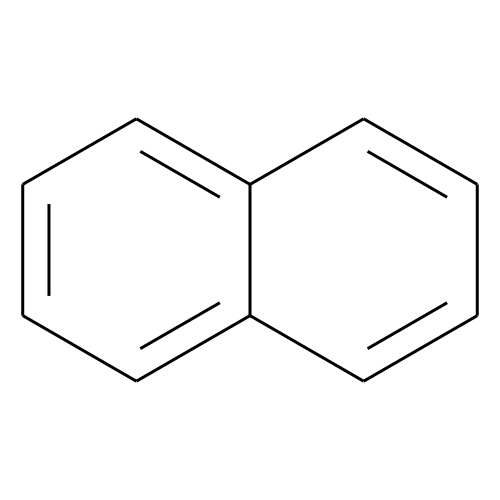}} &      8.16 &  \multirow{2}{*}[-0.35in]{0.08} & 0.51 & 1.00 &  2.84 \\*
           & original &              \parbox[c]{0.125\textwidth}{\includegraphics[width=0.125\textwidth]{SVGs/Quercetin_original.png}} &      6.70 &                                 & 0.43 & 2.54 &  1.99 \\
\hline
\multirow{2}{*}[-0.35in]{Remdesivir} & improved &          \parbox[c]{0.125\textwidth}{\includegraphics[width=0.125\textwidth]{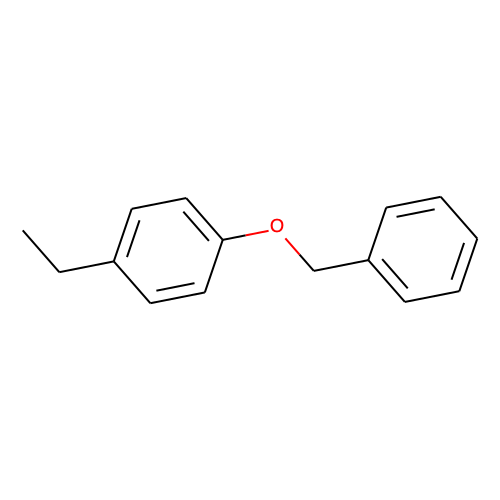}} &      8.21 &  \multirow{2}{*}[-0.35in]{0.13} & 0.75 & 1.25 &  3.83 \\*
           & original &             \parbox[c]{0.125\textwidth}{\includegraphics[width=0.125\textwidth]{SVGs/Remdesivir_original.png}} &      6.32 &                                 & 0.16 & 4.82 &  2.31 \\
\hline
\multirow{2}{*}[-0.35in]{Shikonin} & improved &            \parbox[c]{0.125\textwidth}{\includegraphics[width=0.125\textwidth]{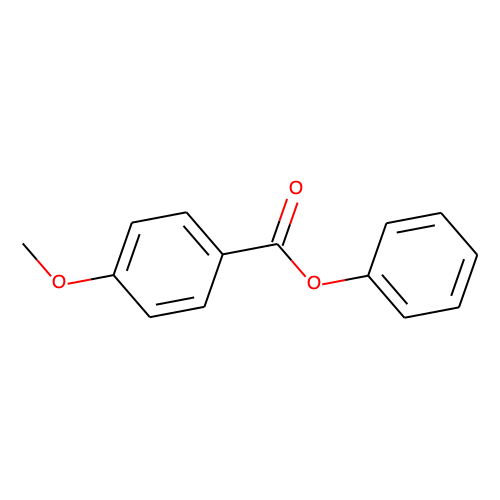}} &      7.92 &  \multirow{2}{*}[-0.35in]{0.11} & 0.60 & 1.31 &  2.91 \\*
           & original &               \parbox[c]{0.125\textwidth}{\includegraphics[width=0.125\textwidth]{SVGs/Shikonin_original.png}} &      5.11 &                                 & 0.58 & 3.41 &  2.12 \\
\hline
\multirow{2}{*}[-0.35in]{Tideglusib} & improved &          \parbox[c]{0.125\textwidth}{\includegraphics[width=0.125\textwidth]{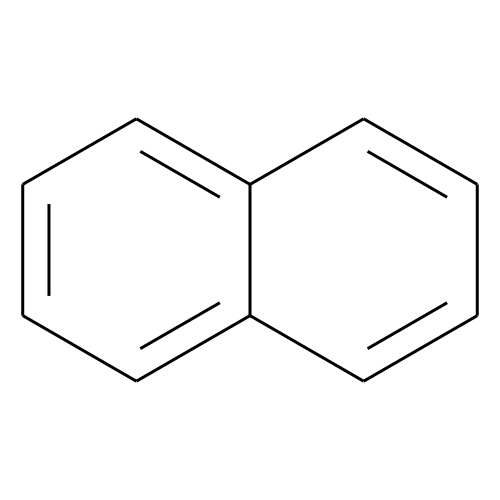}} &      8.16 &  \multirow{2}{*}[-0.35in]{0.19} & 0.51 & 1.00 &  2.84 \\*
           & original &             \parbox[c]{0.125\textwidth}{\includegraphics[width=0.125\textwidth]{SVGs/Tideglusib_original.png}} &      4.94 &                                 & 0.58 & 2.28 &  3.26 \\
\hline
\multirow{2}{*}[-0.35in]{Umifenovir} & improved &          \parbox[c]{0.125\textwidth}{\includegraphics[width=0.125\textwidth]{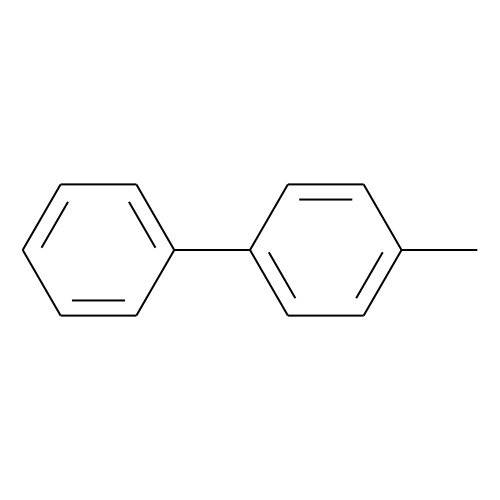}} &      7.51 &  \multirow{2}{*}[-0.35in]{0.13} & 0.61 & 1.07 &  3.66 \\*
           & original &             \parbox[c]{0.125\textwidth}{\includegraphics[width=0.125\textwidth]{SVGs/Umifenovir_original.png}} &      6.36 &                                 & 0.38 & 2.68 &  5.18 \\
\hline
\multirow{2}{*}{Average} & improved & & \textbf{7.56} & \multirow{2}{*}{0.10} & 0.61 & \textbf{1.11} & 3.42 \\
 & original & & 5.79 & & 0.50 & 3.17 & 2.63 \\
\bottomrule
\caption{Results of QMO optimization starting from 23 SARS-CoV-2 inhibitor lead-molecules with M\textsuperscript{pro}-affinity $\ge7.5$, min-SA objective.}
\label{tab:Mpro_SA}
\end{longtable}
\end{center}

\newpage

\begin{center}
\begin{longtable}{ll|p{0.2\textwidth}ccccc}
\toprule
compound & & Molecule & \textcolor{blue}{Affinity} & Similarity & QED & \textcolor{blue}{SA} & logP \\\endfirsthead
\multicolumn{6}{l}{\bfseries \tablename\ \thetable{} continued}\\[10pt]
compound & & Molecule & \textcolor{blue}{Affinity} & Similarity & QED & \textcolor{blue}{SA} & logP \\\endhead
\midrule
\multirow{2}{*}[-0.35in]{0} & improved &        \parbox[c]{0.125\textwidth}{\includegraphics[width=0.125\textwidth]{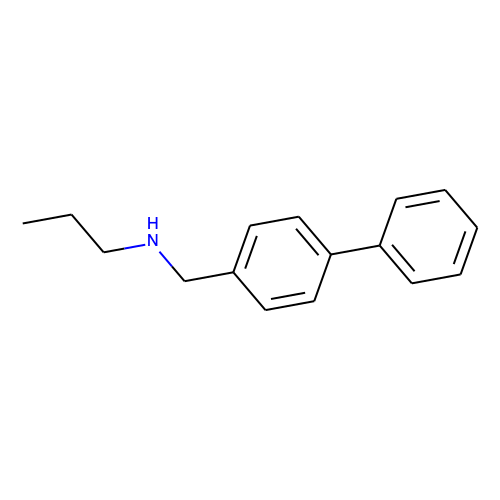}} &      8.08 &  \multirow{2}{*}[-0.35in]{0.06} & 0.76 & 1.40 &  3.85 \\*
           & original &           \parbox[c]{0.125\textwidth}{\includegraphics[width=0.125\textwidth]{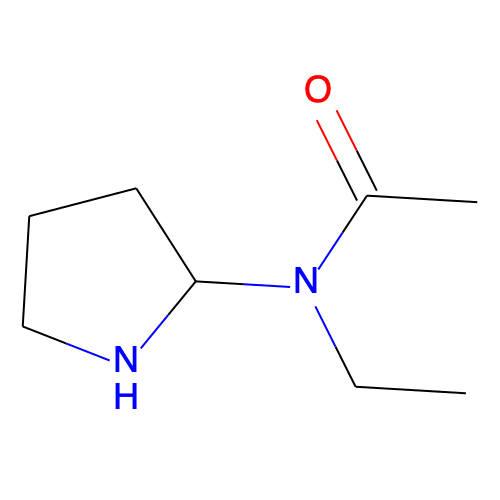}} &      6.46 &                                 & 0.63 & 3.41 &  0.56 \\
\hline
\multirow{2}{*}[-0.35in]{1} & improved &        \parbox[c]{0.125\textwidth}{\includegraphics[width=0.125\textwidth]{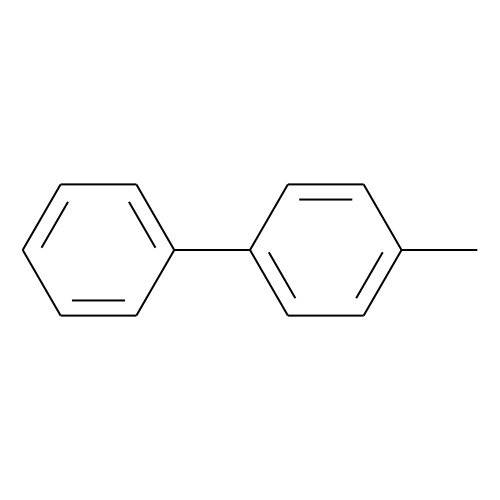}} &      7.51 &  \multirow{2}{*}[-0.35in]{0.11} & 0.61 & 1.07 &  3.66 \\*
           & original &           \parbox[c]{0.125\textwidth}{\includegraphics[width=0.125\textwidth]{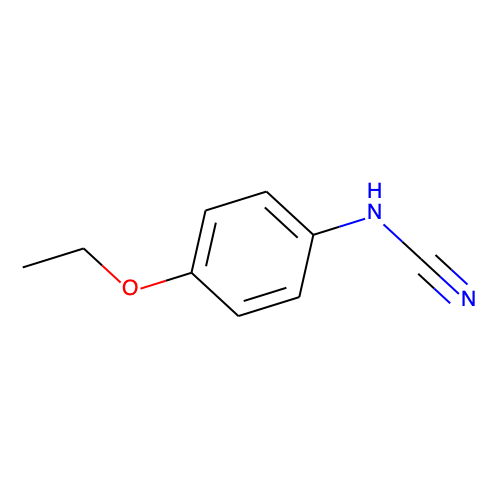}} &      5.87 &                                 & 0.55 & 2.02 &  1.98 \\
\hline
\multirow{2}{*}[-0.35in]{2} & improved &        \parbox[c]{0.125\textwidth}{\includegraphics[width=0.125\textwidth]{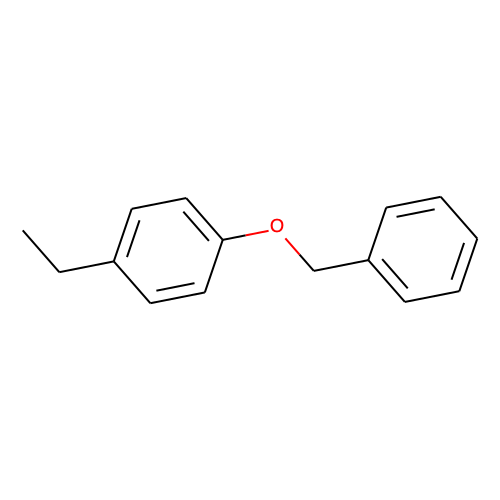}} &      8.21 &  \multirow{2}{*}[-0.35in]{0.11} & 0.75 & 1.25 &  3.83 \\*
           & original &           \parbox[c]{0.125\textwidth}{\includegraphics[width=0.125\textwidth]{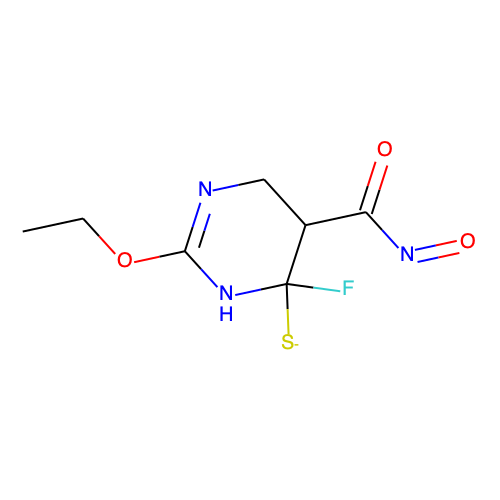}} &      5.96 &                                 & 0.41 & 5.19 &  0.06 \\
\hline
\multirow{2}{*}[-0.35in]{3} & improved &        \parbox[c]{0.125\textwidth}{\includegraphics[width=0.125\textwidth]{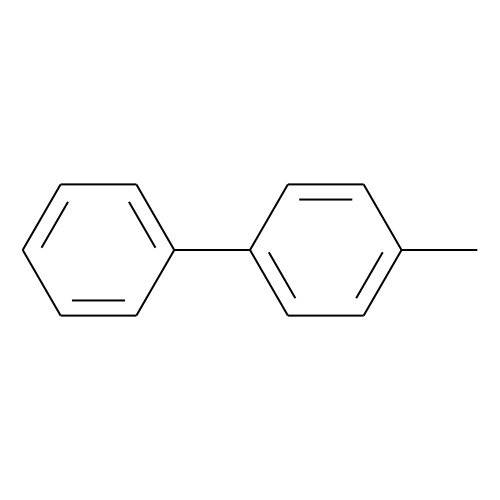}} &      7.51 &  \multirow{2}{*}[-0.35in]{0.05} & 0.61 & 1.07 &  3.66 \\*
           & original &           \parbox[c]{0.125\textwidth}{\includegraphics[width=0.125\textwidth]{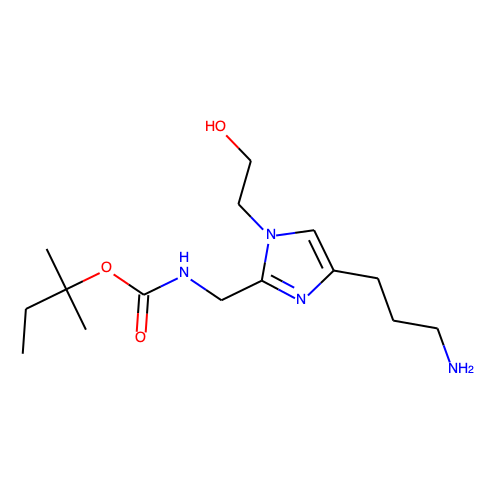}} &      7.68 &                                 & 0.64 & 2.93 &  1.18 \\
\hline
\multirow{2}{*}[-0.35in]{4} & improved &        \parbox[c]{0.125\textwidth}{\includegraphics[width=0.125\textwidth]{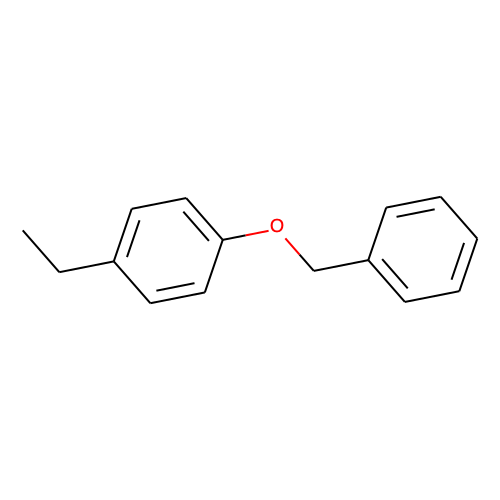}} &      8.21 &  \multirow{2}{*}[-0.35in]{0.09} & 0.75 & 1.25 &  3.83 \\*
           & original &           \parbox[c]{0.125\textwidth}{\includegraphics[width=0.125\textwidth]{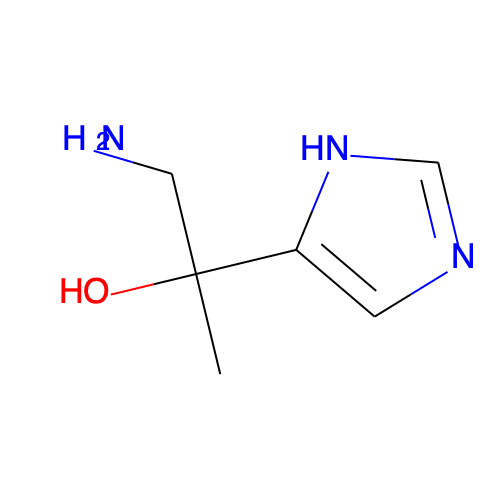}} &      5.58 &                                 & 0.52 & 3.70 & -0.42 \\
\hline
\multirow{2}{*}[-0.35in]{5} & improved &        \parbox[c]{0.125\textwidth}{\includegraphics[width=0.125\textwidth]{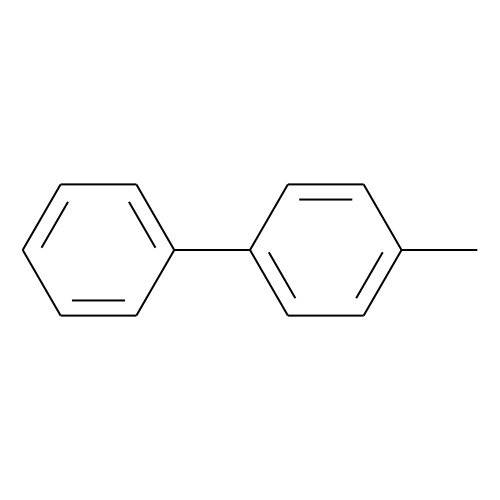}} &      7.51 &  \multirow{2}{*}[-0.35in]{0.08} & 0.61 & 1.07 &  3.66 \\*
           & original &           \parbox[c]{0.125\textwidth}{\includegraphics[width=0.125\textwidth]{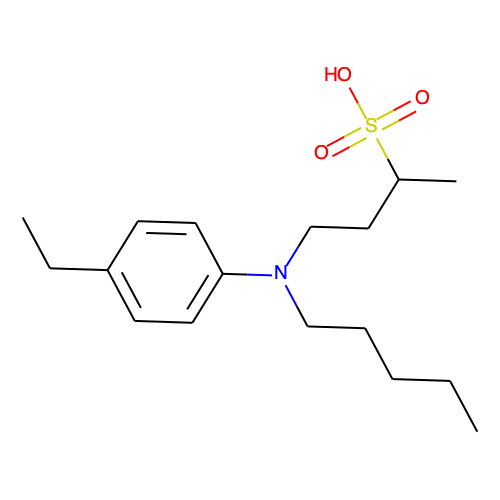}} &      7.34 &                                 & 0.52 & 2.79 &  3.91 \\
\hline
\multirow{2}{*}[-0.35in]{6} & improved &        \parbox[c]{0.125\textwidth}{\includegraphics[width=0.125\textwidth]{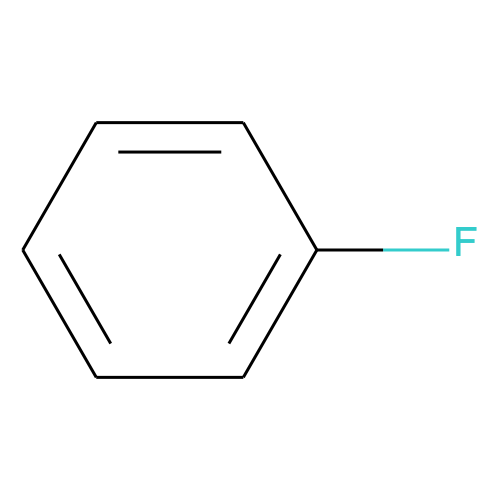}} &      3.90 &  \multirow{2}{*}[-0.35in]{0.13} & 0.46 & 1.00 &  1.83 \\*
           & original &           \parbox[c]{0.125\textwidth}{\includegraphics[width=0.125\textwidth]{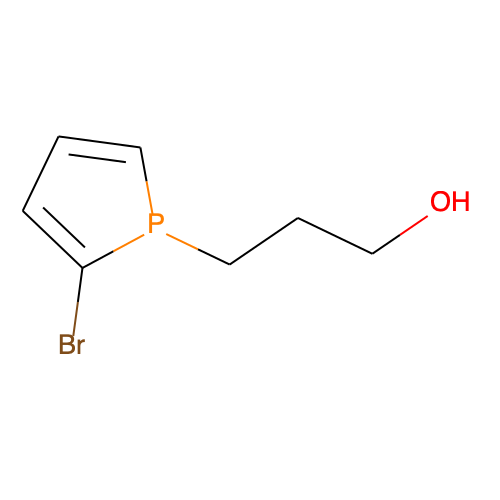}} &      6.39 &                                 & 0.83 & 4.86 &  2.82 \\
\hline
\multirow{2}{*}[-0.35in]{7} & improved &        \parbox[c]{0.125\textwidth}{\includegraphics[width=0.125\textwidth]{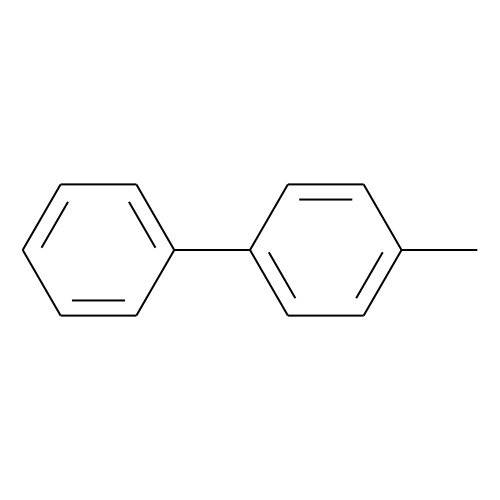}} &      7.51 &  \multirow{2}{*}[-0.35in]{0.04} & 0.61 & 1.07 &  3.66 \\*
           & original &           \parbox[c]{0.125\textwidth}{\includegraphics[width=0.125\textwidth]{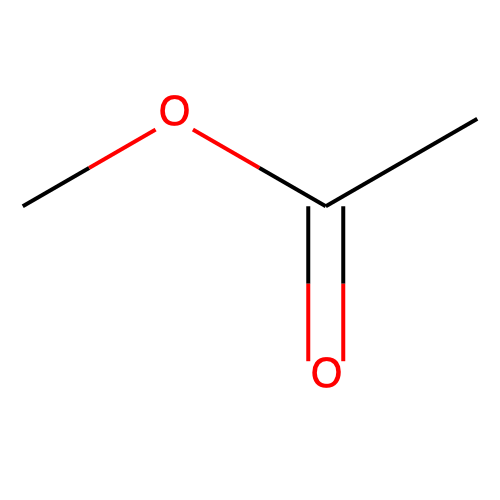}} &      4.91 &                                 & 0.38 & 1.74 &  0.18 \\
\hline
\multirow{2}{*}[-0.35in]{8} & improved &        \parbox[c]{0.125\textwidth}{\includegraphics[width=0.125\textwidth]{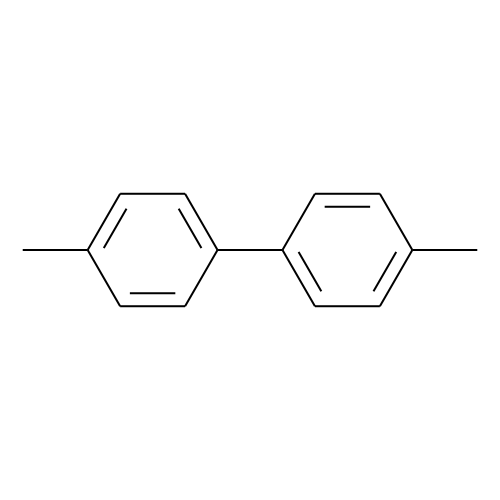}} &      8.12 &  \multirow{2}{*}[-0.35in]{0.09} & 0.63 & 1.00 &  3.97 \\*
           & original &           \parbox[c]{0.125\textwidth}{\includegraphics[width=0.125\textwidth]{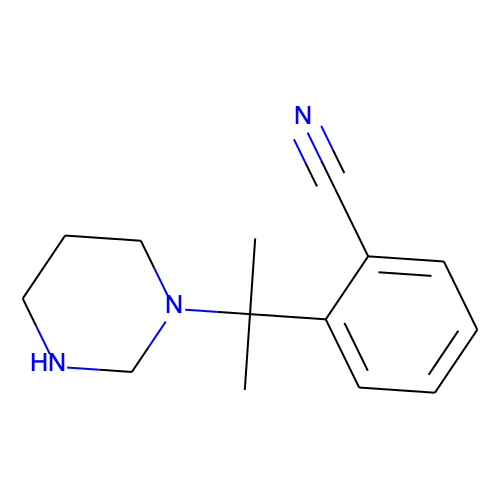}} &      5.00 &                                 & 0.84 & 2.83 &  2.05 \\
\hline
\multirow{2}{*}[-0.35in]{9} & improved &        \parbox[c]{0.125\textwidth}{\includegraphics[width=0.125\textwidth]{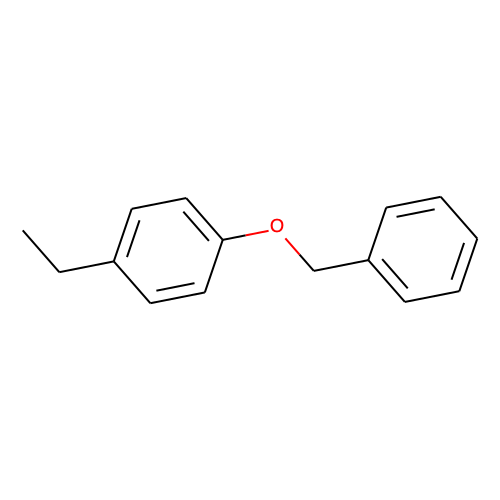}} &      8.21 &  \multirow{2}{*}[-0.35in]{0.04} & 0.75 & 1.25 &  3.83 \\*
           & original &           \parbox[c]{0.125\textwidth}{\includegraphics[width=0.125\textwidth]{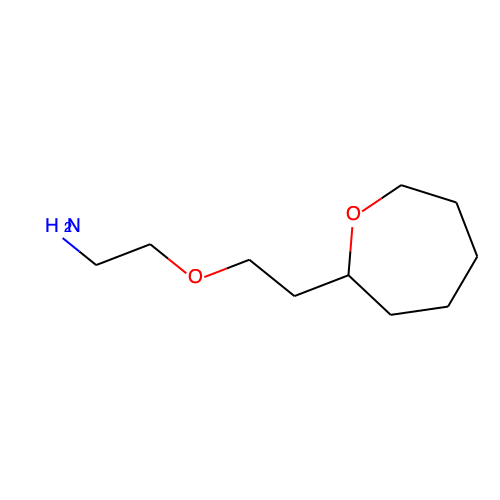}} &      4.66 &                                 & 0.66 & 2.71 &  1.31 \\
\hline
\multirow{2}{*}[-0.35in]{10} & improved &       \parbox[c]{0.125\textwidth}{\includegraphics[width=0.125\textwidth]{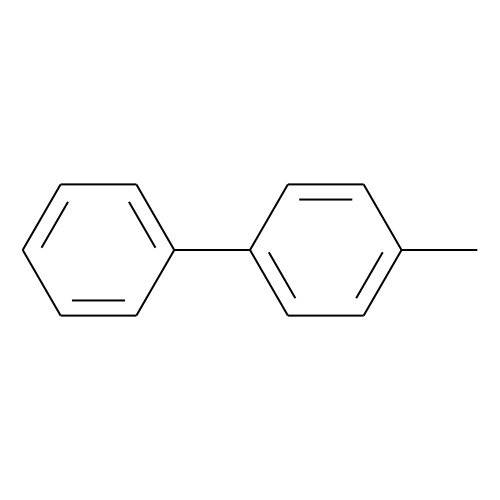}} &      7.51 &  \multirow{2}{*}[-0.35in]{0.11} & 0.61 & 1.07 &  3.66 \\*
           & original &          \parbox[c]{0.125\textwidth}{\includegraphics[width=0.125\textwidth]{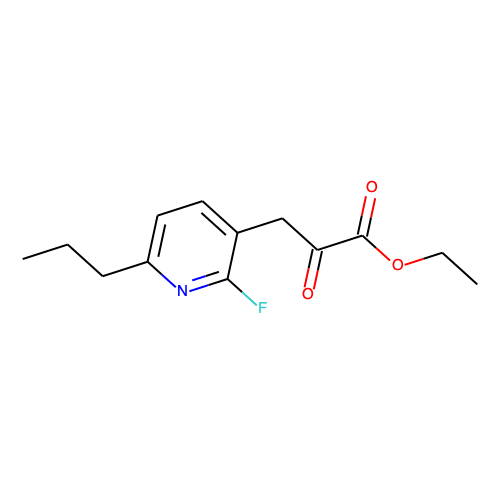}} &      6.11 &                                 & 0.44 & 2.54 &  1.85 \\
\hline
\multirow{2}{*}[-0.35in]{11} & improved &       \parbox[c]{0.125\textwidth}{\includegraphics[width=0.125\textwidth]{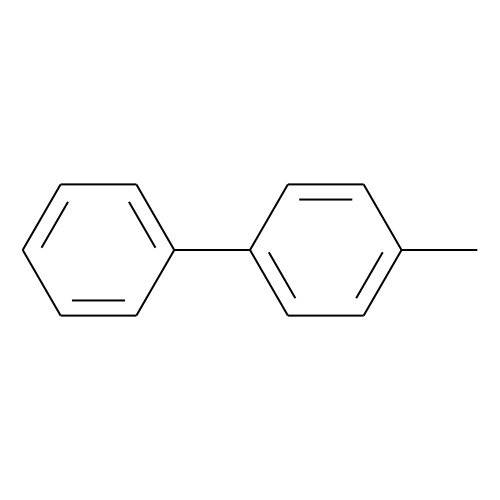}} &      7.51 &  \multirow{2}{*}[-0.35in]{0.10} & 0.61 & 1.07 &  3.66 \\*
           & original &          \parbox[c]{0.125\textwidth}{\includegraphics[width=0.125\textwidth]{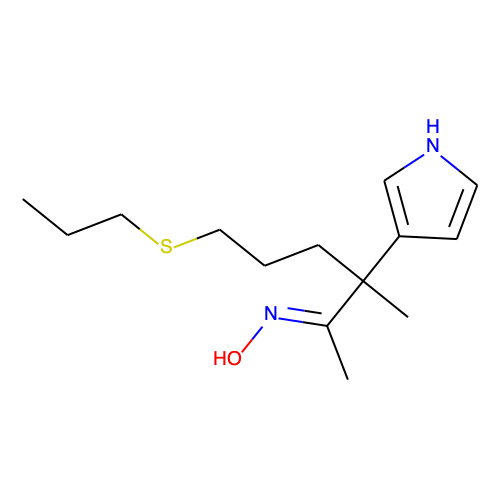}} &      7.33 &                                 & 0.32 & 3.96 &  4.05 \\
\hline
\multirow{2}{*}[-0.35in]{12} & improved &       \parbox[c]{0.125\textwidth}{\includegraphics[width=0.125\textwidth]{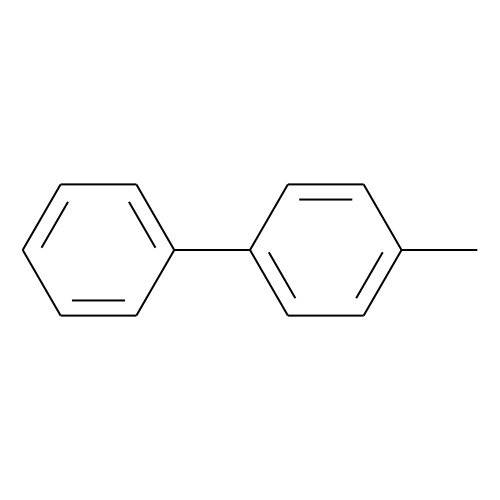}} &      7.51 &  \multirow{2}{*}[-0.35in]{0.07} & 0.61 & 1.07 &  3.66 \\*
           & original &          \parbox[c]{0.125\textwidth}{\includegraphics[width=0.125\textwidth]{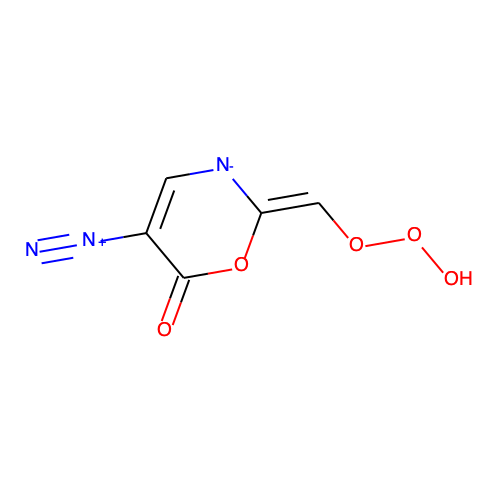}} &      5.82 &                                 & 0.22 & 5.52 &  0.83 \\
\hline
\multirow{2}{*}[-0.35in]{13} & improved &       \parbox[c]{0.125\textwidth}{\includegraphics[width=0.125\textwidth]{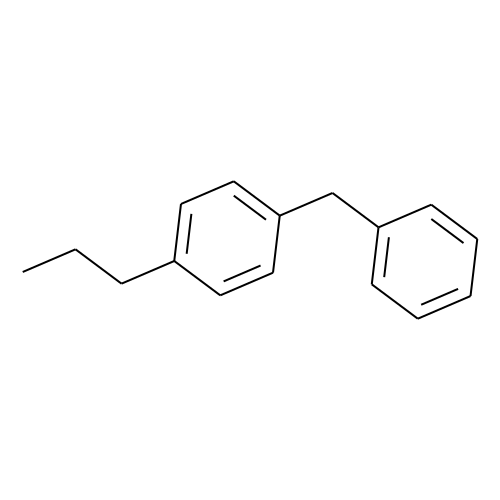}} &      8.29 &  \multirow{2}{*}[-0.35in]{0.15} & 0.71 & 1.34 &  4.23 \\*
           & original &          \parbox[c]{0.125\textwidth}{\includegraphics[width=0.125\textwidth]{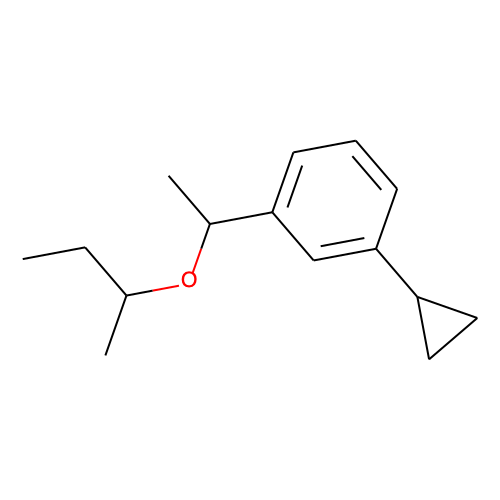}} &      7.26 &                                 & 0.71 & 3.01 &  4.44 \\
\hline
\multirow{2}{*}[-0.35in]{14} & improved &       \parbox[c]{0.125\textwidth}{\includegraphics[width=0.125\textwidth]{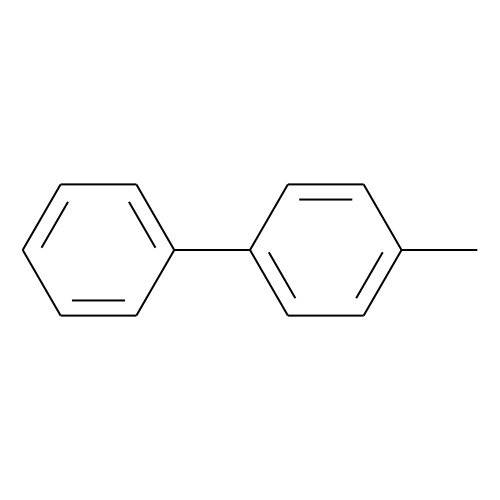}} &      7.51 &  \multirow{2}{*}[-0.35in]{0.00} & 0.61 & 1.07 &  3.66 \\*
           & original &          \parbox[c]{0.125\textwidth}{\includegraphics[width=0.125\textwidth]{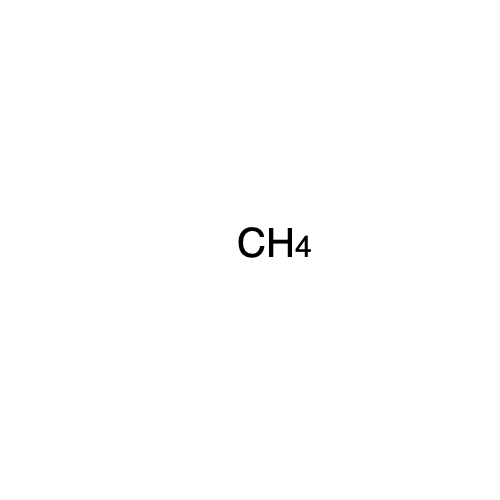}} &      5.58 &                                 & 0.36 & 7.33 &  0.64 \\
\hline
\multirow{2}{*}[-0.35in]{15} & improved &       \parbox[c]{0.125\textwidth}{\includegraphics[width=0.125\textwidth]{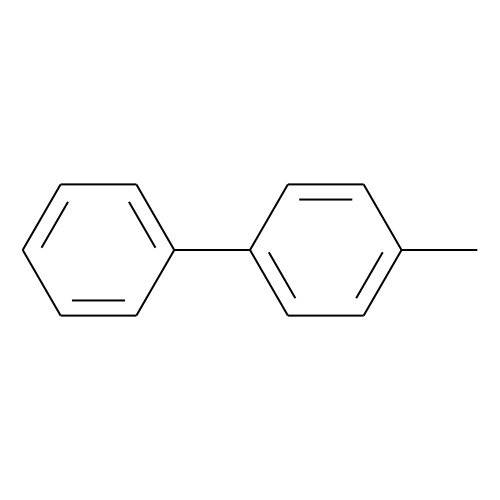}} &      7.51 &  \multirow{2}{*}[-0.35in]{0.07} & 0.61 & 1.07 &  3.66 \\*
           & original &          \parbox[c]{0.125\textwidth}{\includegraphics[width=0.125\textwidth]{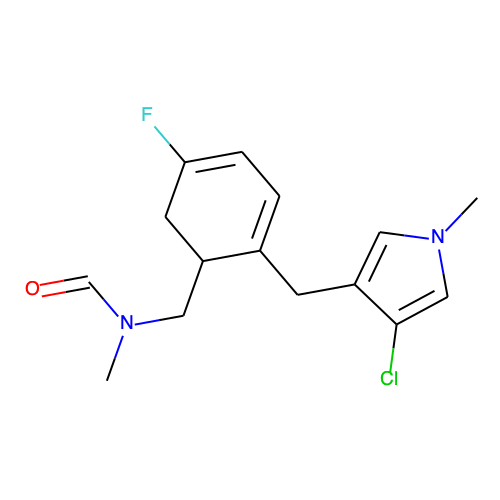}} &      5.35 &                                 & 0.77 & 4.45 &  3.11 \\
\hline
\multirow{2}{*}[-0.35in]{16} & improved &       \parbox[c]{0.125\textwidth}{\includegraphics[width=0.125\textwidth]{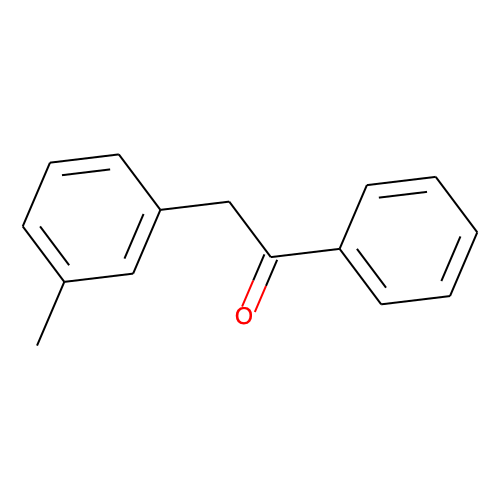}} &      7.75 &  \multirow{2}{*}[-0.35in]{0.04} & 0.71 & 1.42 &  3.42 \\*
           & original &          \parbox[c]{0.125\textwidth}{\includegraphics[width=0.125\textwidth]{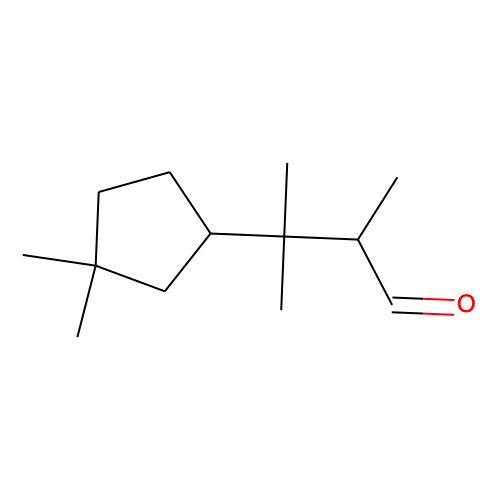}} &      5.00 &                                 & 0.63 & 3.98 &  3.67 \\
\hline
\multirow{2}{*}[-0.35in]{17} & improved &       \parbox[c]{0.125\textwidth}{\includegraphics[width=0.125\textwidth]{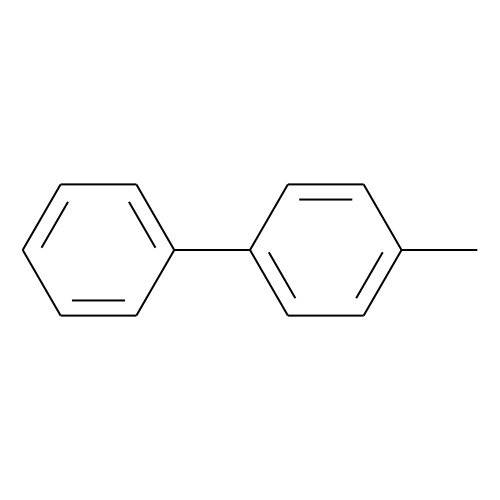}} &      7.51 &  \multirow{2}{*}[-0.35in]{0.11} & 0.61 & 1.07 &  3.66 \\*
           & original &          \parbox[c]{0.125\textwidth}{\includegraphics[width=0.125\textwidth]{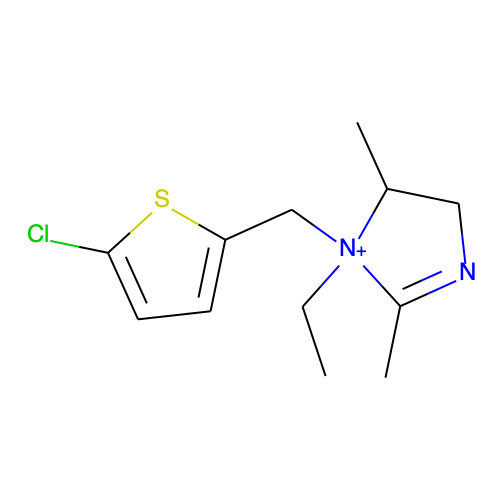}} &      6.39 &                                 & 0.73 & 4.50 &  3.56 \\
\hline
\multirow{2}{*}[-0.35in]{18} & improved &       \parbox[c]{0.125\textwidth}{\includegraphics[width=0.125\textwidth]{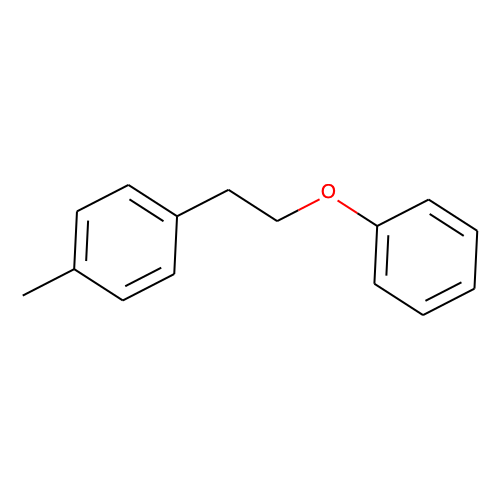}} &      8.55 &  \multirow{2}{*}[-0.35in]{0.16} & 0.75 & 1.31 &  3.62 \\*
           & original &          \parbox[c]{0.125\textwidth}{\includegraphics[width=0.125\textwidth]{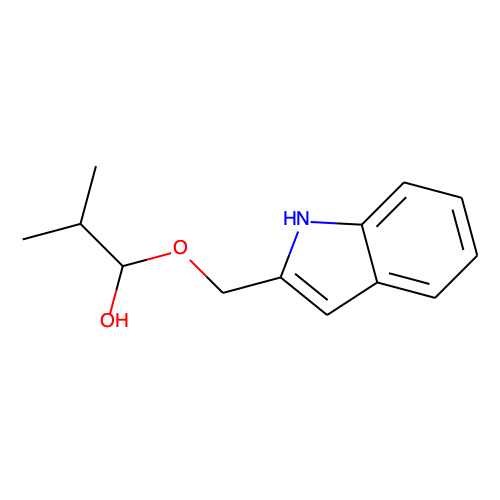}} &      7.46 &                                 & 0.78 & 2.99 &  2.66 \\
\hline
\multirow{2}{*}[-0.35in]{19} & improved &       \parbox[c]{0.125\textwidth}{\includegraphics[width=0.125\textwidth]{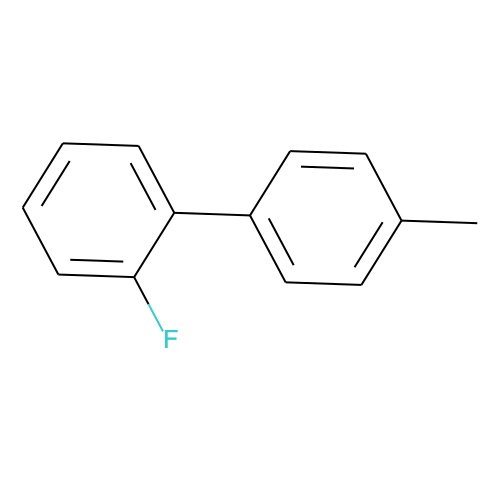}} &      8.40 &  \multirow{2}{*}[-0.35in]{0.04} & 0.64 & 1.28 &  3.80 \\*
           & original &          \parbox[c]{0.125\textwidth}{\includegraphics[width=0.125\textwidth]{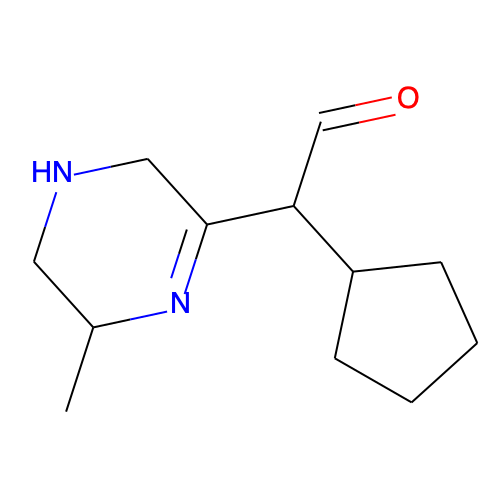}} &      5.62 &                                 & 0.71 & 4.36 &  1.42 \\
\hline
\multirow{2}{*}[-0.35in]{20} & improved &       \parbox[c]{0.125\textwidth}{\includegraphics[width=0.125\textwidth]{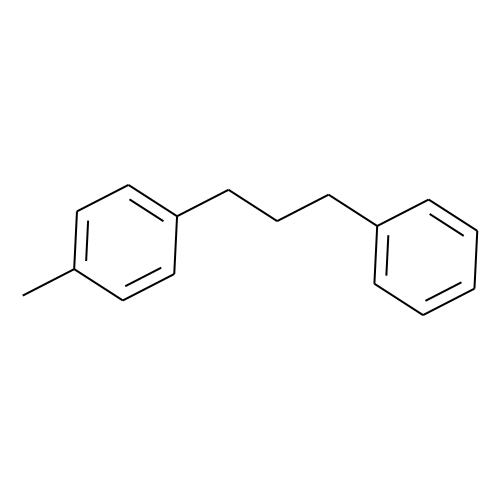}} &      7.91 &  \multirow{2}{*}[-0.35in]{0.05} & 0.71 & 1.29 &  4.17 \\*
           & original &          \parbox[c]{0.125\textwidth}{\includegraphics[width=0.125\textwidth]{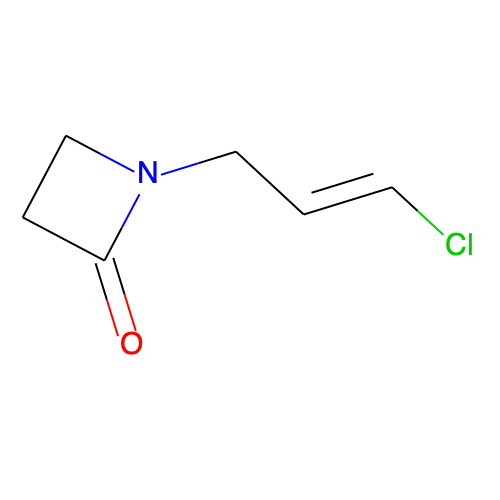}} &      5.42 &                                 & 0.53 & 3.04 &  0.97 \\
\hline
\multirow{2}{*}[-0.35in]{21} & improved &       \parbox[c]{0.125\textwidth}{\includegraphics[width=0.125\textwidth]{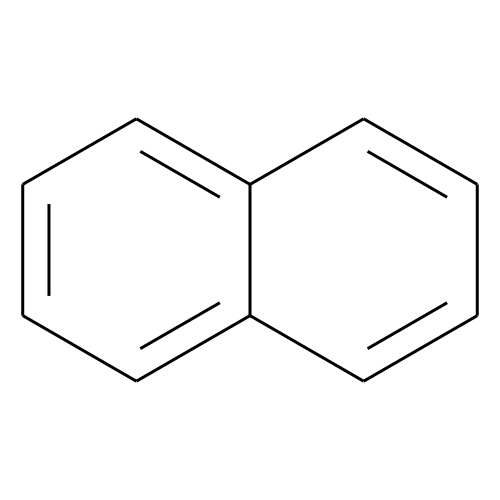}} &      8.16 &  \multirow{2}{*}[-0.35in]{0.06} & 0.51 & 1.00 &  2.84 \\*
           & original &          \parbox[c]{0.125\textwidth}{\includegraphics[width=0.125\textwidth]{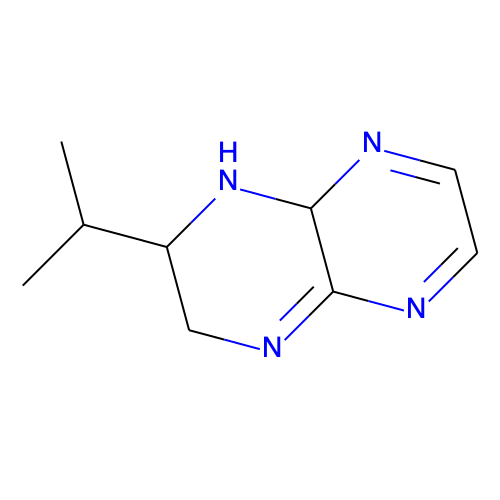}} &      4.75 &                                 & 0.62 & 4.99 &  0.49 \\
\hline
\multirow{2}{*}[-0.35in]{22} & improved &       \parbox[c]{0.125\textwidth}{\includegraphics[width=0.125\textwidth]{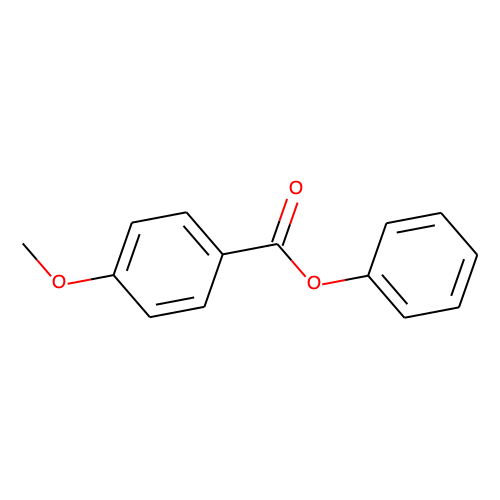}} &      7.92 &  \multirow{2}{*}[-0.35in]{0.11} & 0.60 & 1.31 &  2.91 \\*
           & original &          \parbox[c]{0.125\textwidth}{\includegraphics[width=0.125\textwidth]{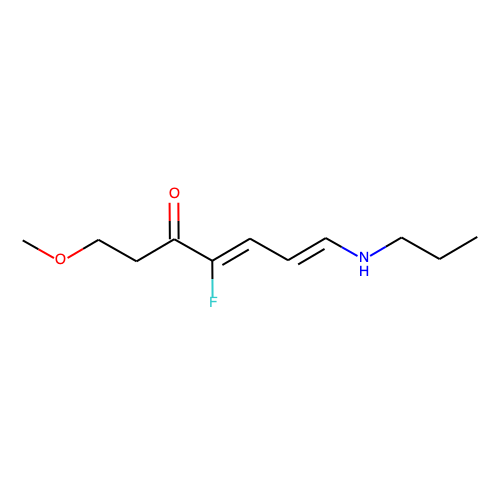}} &      6.86 &                                 & 0.38 & 3.45 &  1.96 \\
\hline
\multirow{2}{*}{Average} & improved &  &  \textbf{7.69} &  \multirow{2}{*}{0.08} & 0.64 & \textbf{1.17} &  3.60 \\*
           & original &  &      6.03 &                                 & 0.57 & 3.75 &  1.88 \\
\bottomrule
\caption{Results of QMO optimization starting from 23 randomly generated molecules with M\textsuperscript{pro}-affinity $\ge7.5$, min-SA objective.}
\label{tab:Mpro_rand_SA}
\end{longtable}
\end{center}

\newpage

\begin{center}
\begin{longtable}{ll|p{0.2\textwidth}ccccc}
\toprule
compound & & Molecule & \textcolor{blue}{Affinity} & Similarity & \textcolor{blue}{QED} & SA & logP \\\endfirsthead
\multicolumn{6}{l}{\bfseries \tablename\ \thetable{} continued}\\[10pt]
compound & & Molecule & \textcolor{blue}{Affinity} & Similarity & \textcolor{blue}{QED} & SA & logP \\\endhead
\midrule
\multirow{2}{*}[-0.35in]{0} & improved &        \parbox[c]{0.125\textwidth}{\includegraphics[width=0.125\textwidth]{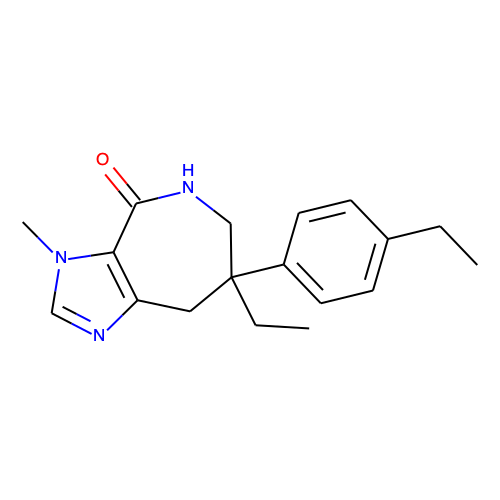}} &      7.64 &  \multirow{2}{*}[-0.35in]{0.11} & 0.95 & 3.39 &  2.62 \\*
           & original &            \parbox[c]{0.125\textwidth}{\includegraphics[width=0.125\textwidth]{SVGs/0_rand_original.png}} &      6.46 &                                 & 0.63 & 3.41 &  0.56 \\
\hline
\multirow{2}{*}[-0.35in]{1} & improved &        \parbox[c]{0.125\textwidth}{\includegraphics[width=0.125\textwidth]{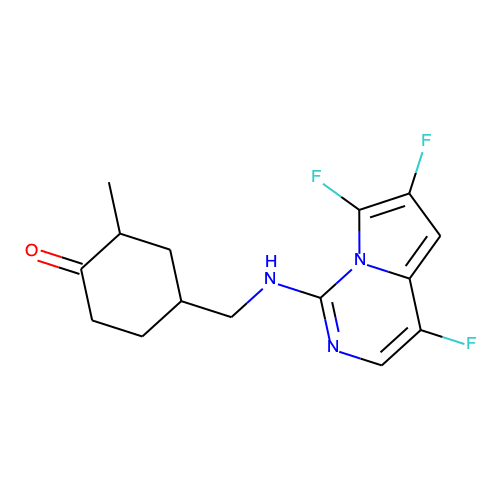}} &      8.29 &  \multirow{2}{*}[-0.35in]{0.09} & 0.95 & 4.08 &  3.17 \\*
           & original &            \parbox[c]{0.125\textwidth}{\includegraphics[width=0.125\textwidth]{SVGs/1_rand_original.png}} &      5.87 &                                 & 0.55 & 2.02 &  1.98 \\
\hline
\multirow{2}{*}[-0.35in]{2} & improved &        \parbox[c]{0.125\textwidth}{\includegraphics[width=0.125\textwidth]{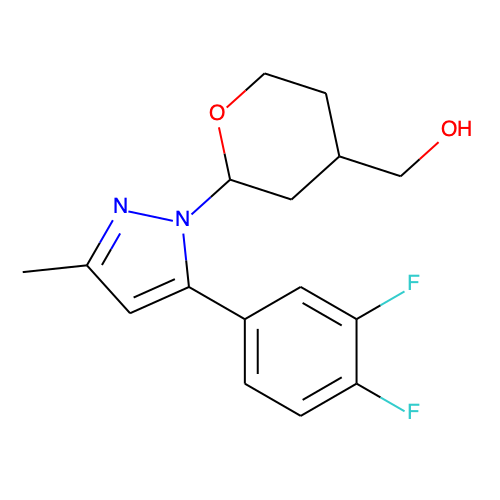}} &      7.58 &  \multirow{2}{*}[-0.35in]{0.10} & 0.95 & 3.46 &  3.05 \\*
           & original &            \parbox[c]{0.125\textwidth}{\includegraphics[width=0.125\textwidth]{SVGs/2_rand_original.png}} &      5.96 &                                 & 0.41 & 5.19 &  0.06 \\
\hline
\multirow{2}{*}[-0.35in]{3} & improved &        \parbox[c]{0.125\textwidth}{\includegraphics[width=0.125\textwidth]{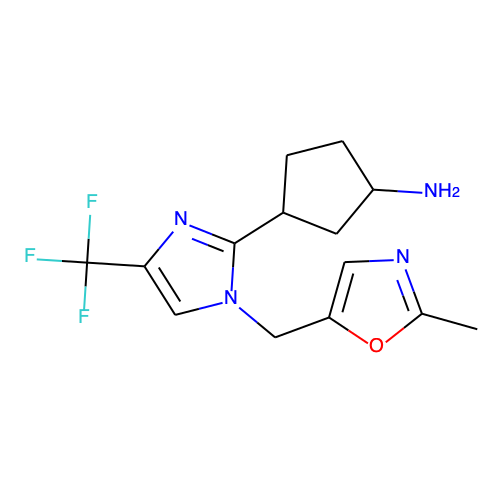}} &      7.51 &  \multirow{2}{*}[-0.35in]{0.15} & 0.95 & 3.93 &  2.84 \\*
           & original &            \parbox[c]{0.125\textwidth}{\includegraphics[width=0.125\textwidth]{SVGs/3_rand_original.png}} &      7.68 &                                 & 0.64 & 2.93 &  1.18 \\
\hline
\multirow{2}{*}[-0.35in]{4} & improved &        \parbox[c]{0.125\textwidth}{\includegraphics[width=0.125\textwidth]{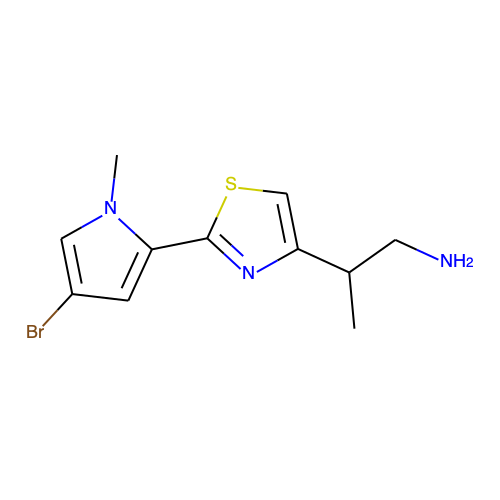}} &      7.57 &  \multirow{2}{*}[-0.35in]{0.12} & 0.95 & 3.49 &  2.97 \\*
           & original &            \parbox[c]{0.125\textwidth}{\includegraphics[width=0.125\textwidth]{SVGs/4_rand_original.png}} &      5.58 &                                 & 0.52 & 3.70 & -0.42 \\
\hline
\multirow{2}{*}[-0.35in]{5} & improved &        \parbox[c]{0.125\textwidth}{\includegraphics[width=0.125\textwidth]{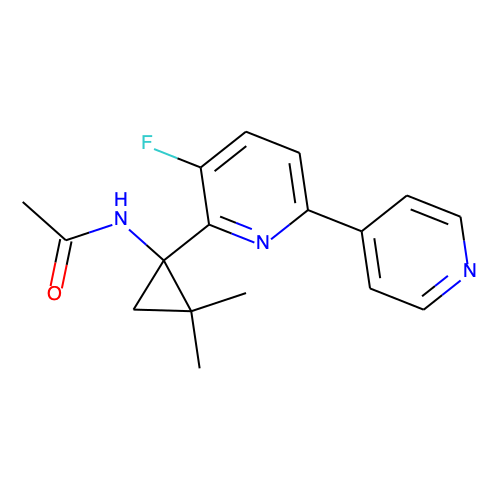}} &      7.77 &  \multirow{2}{*}[-0.35in]{0.08} & 0.95 & 3.30 &  3.04 \\*
           & original &            \parbox[c]{0.125\textwidth}{\includegraphics[width=0.125\textwidth]{SVGs/5_rand_original.png}} &      7.34 &                                 & 0.52 & 2.79 &  3.91 \\
\hline
\multirow{2}{*}[-0.35in]{6} & improved &        \parbox[c]{0.125\textwidth}{\includegraphics[width=0.125\textwidth]{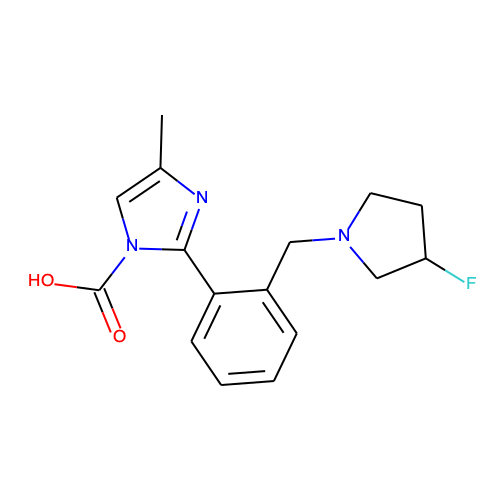}} &      7.78 &  \multirow{2}{*}[-0.35in]{0.11} & 0.95 & 3.15 &  2.93 \\*
           & original &            \parbox[c]{0.125\textwidth}{\includegraphics[width=0.125\textwidth]{SVGs/6_rand_original.png}} &      6.39 &                                 & 0.83 & 4.86 &  2.82 \\
\hline
\multirow{2}{*}[-0.35in]{7} & improved &        \parbox[c]{0.125\textwidth}{\includegraphics[width=0.125\textwidth]{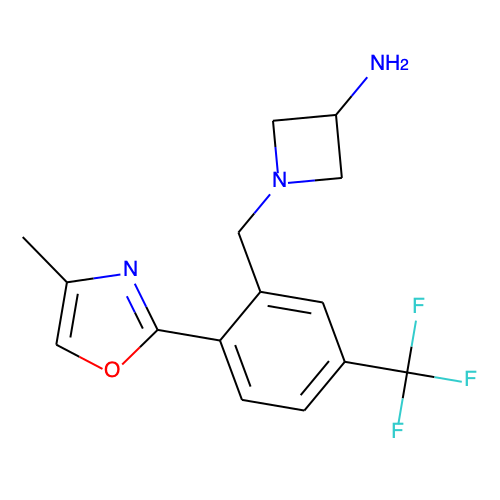}} &      7.76 &  \multirow{2}{*}[-0.35in]{0.02} & 0.95 & 2.66 &  2.81 \\*
           & original &            \parbox[c]{0.125\textwidth}{\includegraphics[width=0.125\textwidth]{SVGs/7_rand_original.png}} &      4.91 &                                 & 0.38 & 1.74 &  0.18 \\
\hline
\multirow{2}{*}[-0.35in]{8} & improved &        \parbox[c]{0.125\textwidth}{\includegraphics[width=0.125\textwidth]{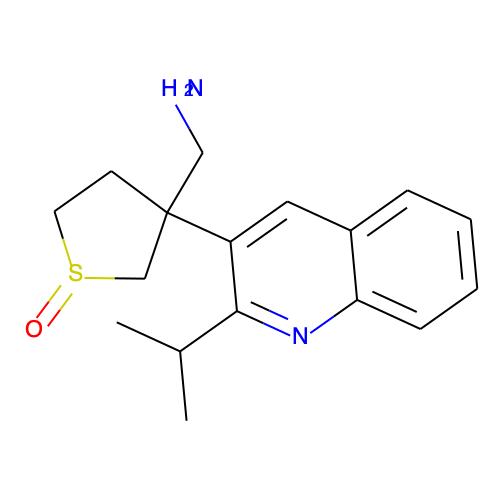}} &      7.67 &  \multirow{2}{*}[-0.35in]{0.13} & 0.95 & 3.89 &  2.71 \\*
           & original &            \parbox[c]{0.125\textwidth}{\includegraphics[width=0.125\textwidth]{SVGs/8_rand_original.png}} &      5.00 &                                 & 0.84 & 2.83 &  2.05 \\
\hline
\multirow{2}{*}[-0.35in]{9} & improved &        \parbox[c]{0.125\textwidth}{\includegraphics[width=0.125\textwidth]{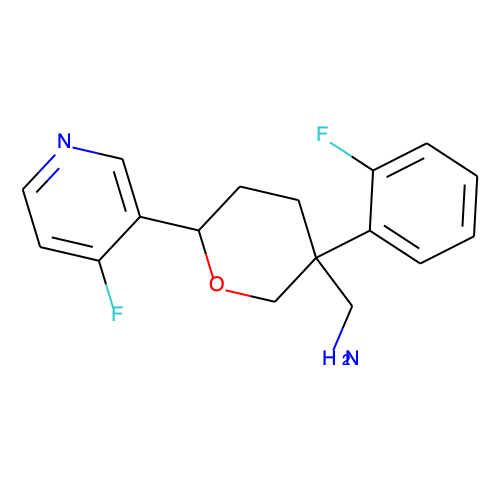}} &      8.21 &  \multirow{2}{*}[-0.35in]{0.14} & 0.95 & 3.54 &  3.11 \\*
           & original &            \parbox[c]{0.125\textwidth}{\includegraphics[width=0.125\textwidth]{SVGs/9_rand_original.png}} &      4.66 &                                 & 0.66 & 2.71 &  1.31 \\
\hline
\multirow{2}{*}[-0.35in]{10} & improved &       \parbox[c]{0.125\textwidth}{\includegraphics[width=0.125\textwidth]{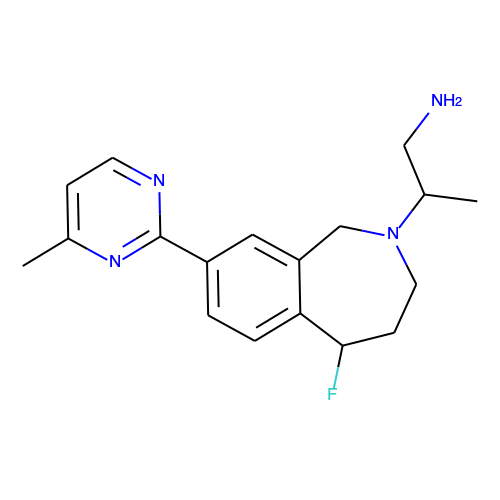}} &      7.56 &  \multirow{2}{*}[-0.35in]{0.13} & 0.95 & 3.45 &  3.02 \\*
           & original &           \parbox[c]{0.125\textwidth}{\includegraphics[width=0.125\textwidth]{SVGs/10_rand_original.png}} &      6.11 &                                 & 0.44 & 2.54 &  1.85 \\
\hline
\multirow{2}{*}[-0.35in]{11} & improved &       \parbox[c]{0.125\textwidth}{\includegraphics[width=0.125\textwidth]{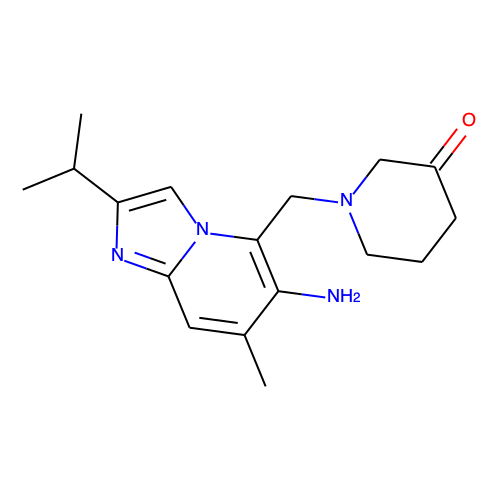}} &      7.51 &  \multirow{2}{*}[-0.35in]{0.06} & 0.95 & 3.00 &  2.51 \\*
           & original &           \parbox[c]{0.125\textwidth}{\includegraphics[width=0.125\textwidth]{SVGs/11_rand_original.png}} &      7.33 &                                 & 0.32 & 3.96 &  4.05 \\
\hline
\multirow{2}{*}[-0.35in]{12} & improved &       \parbox[c]{0.125\textwidth}{\includegraphics[width=0.125\textwidth]{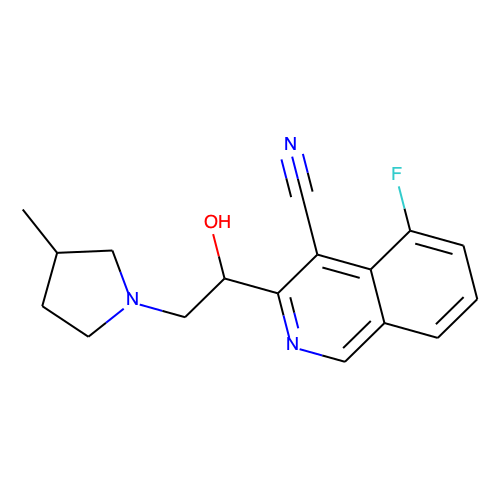}} &      8.19 &  \multirow{2}{*}[-0.35in]{0.06} & 0.95 & 3.45 &  2.62 \\*
           & original &           \parbox[c]{0.125\textwidth}{\includegraphics[width=0.125\textwidth]{SVGs/12_rand_original.png}} &      5.82 &                                 & 0.22 & 5.52 &  0.83 \\
\hline
\multirow{2}{*}[-0.35in]{13} & improved &       \parbox[c]{0.125\textwidth}{\includegraphics[width=0.125\textwidth]{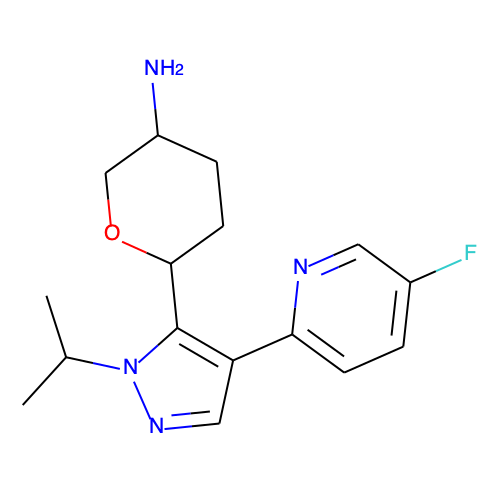}} &      8.07 &  \multirow{2}{*}[-0.35in]{0.13} & 0.95 & 3.75 &  2.84 \\*
           & original &           \parbox[c]{0.125\textwidth}{\includegraphics[width=0.125\textwidth]{SVGs/13_rand_original.png}} &      7.26 &                                 & 0.71 & 3.01 &  4.44 \\
\hline
\multirow{2}{*}[-0.35in]{14} & improved &       \parbox[c]{0.125\textwidth}{\includegraphics[width=0.125\textwidth]{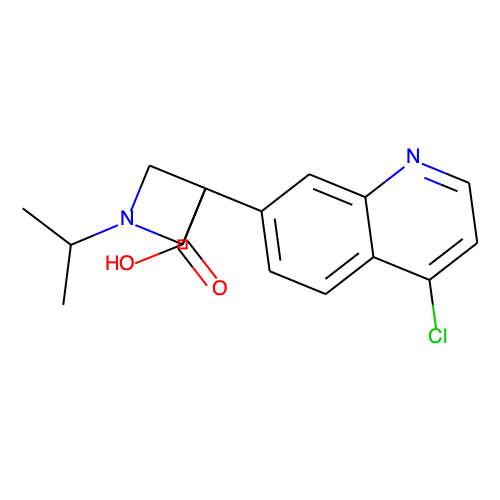}} &      8.32 &  \multirow{2}{*}[-0.35in]{0.00} & 0.95 & 2.69 &  2.93 \\*
           & original &           \parbox[c]{0.125\textwidth}{\includegraphics[width=0.125\textwidth]{SVGs/14_rand_original.png}} &      5.58 &                                 & 0.36 & 7.33 &  0.64 \\
\hline
\multirow{2}{*}[-0.35in]{15} & improved &       \parbox[c]{0.125\textwidth}{\includegraphics[width=0.125\textwidth]{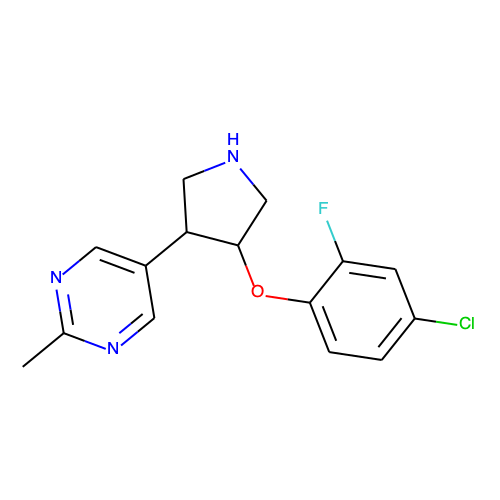}} &      7.76 &  \multirow{2}{*}[-0.35in]{0.11} & 0.95 & 3.43 &  2.71 \\*
           & original &           \parbox[c]{0.125\textwidth}{\includegraphics[width=0.125\textwidth]{SVGs/15_rand_original.png}} &      5.35 &                                 & 0.77 & 4.45 &  3.11 \\
\hline
\multirow{2}{*}[-0.35in]{16} & improved &       \parbox[c]{0.125\textwidth}{\includegraphics[width=0.125\textwidth]{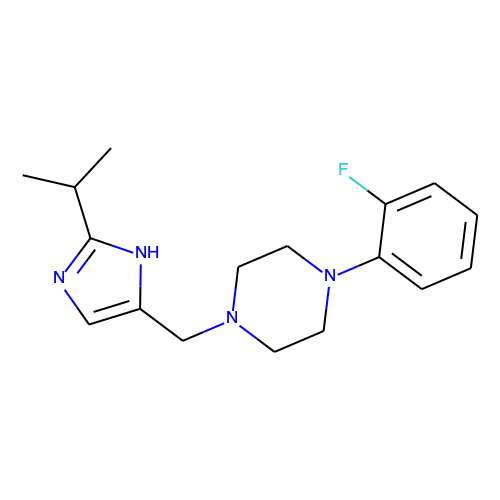}} &      7.66 &  \multirow{2}{*}[-0.35in]{0.08} & 0.94 & 2.27 &  2.99 \\*
           & original &           \parbox[c]{0.125\textwidth}{\includegraphics[width=0.125\textwidth]{SVGs/16_rand_original.png}} &      5.00 &                                 & 0.63 & 3.98 &  3.67 \\
\hline
\multirow{2}{*}[-0.35in]{17} & improved &       \parbox[c]{0.125\textwidth}{\includegraphics[width=0.125\textwidth]{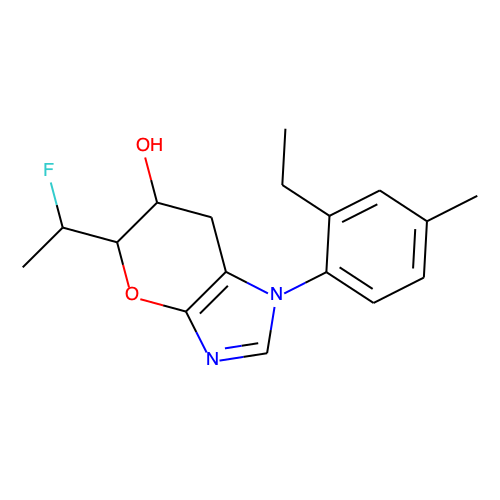}} &      7.65 &  \multirow{2}{*}[-0.35in]{0.15} & 0.95 & 4.07 &  2.77 \\*
           & original &           \parbox[c]{0.125\textwidth}{\includegraphics[width=0.125\textwidth]{SVGs/17_rand_original.png}} &      6.39 &                                 & 0.73 & 4.50 &  3.56 \\
\hline
\multirow{2}{*}[-0.35in]{18} & improved &       \parbox[c]{0.125\textwidth}{\includegraphics[width=0.125\textwidth]{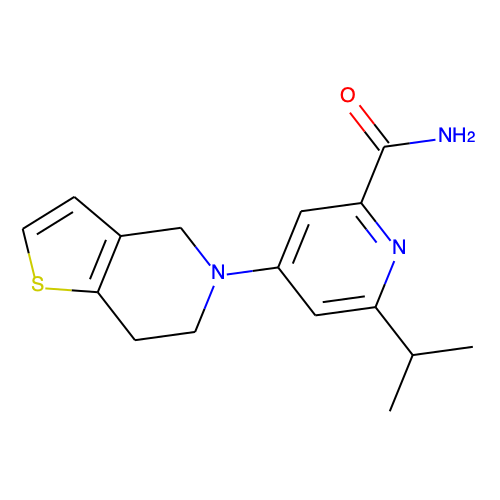}} &      7.71 &  \multirow{2}{*}[-0.35in]{0.13} & 0.95 & 2.71 &  2.93 \\*
           & original &           \parbox[c]{0.125\textwidth}{\includegraphics[width=0.125\textwidth]{SVGs/18_rand_original.png}} &      7.46 &                                 & 0.78 & 2.99 &  2.66 \\
\hline
\multirow{2}{*}[-0.35in]{19} & improved &       \parbox[c]{0.125\textwidth}{\includegraphics[width=0.125\textwidth]{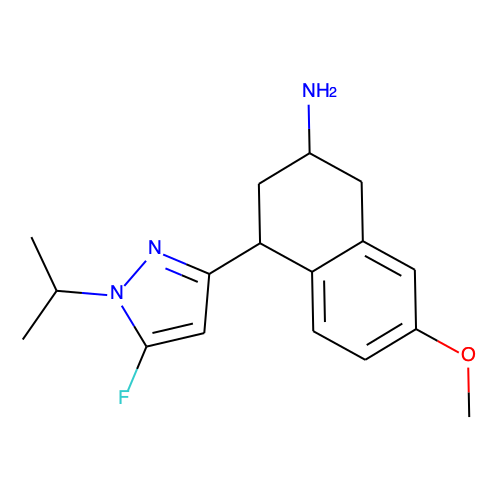}} &      5.86 &  \multirow{2}{*}[-0.35in]{0.08} & 0.95 & 3.81 &  3.02 \\*
           & original &           \parbox[c]{0.125\textwidth}{\includegraphics[width=0.125\textwidth]{SVGs/19_rand_original.png}} &      5.62 &                                 & 0.71 & 4.36 &  1.42 \\
\hline
\multirow{2}{*}[-0.35in]{20} & improved &       \parbox[c]{0.125\textwidth}{\includegraphics[width=0.125\textwidth]{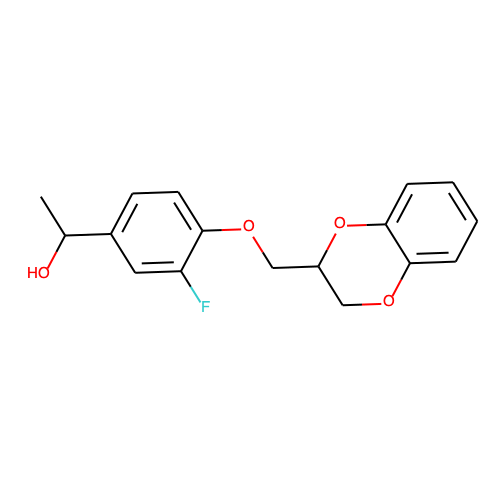}} &      8.11 &  \multirow{2}{*}[-0.35in]{0.05} & 0.94 & 2.89 &  3.10 \\*
           & original &           \parbox[c]{0.125\textwidth}{\includegraphics[width=0.125\textwidth]{SVGs/20_rand_original.png}} &      5.42 &                                 & 0.53 & 3.04 &  0.97 \\
\hline
\multirow{2}{*}[-0.35in]{21} & improved &       \parbox[c]{0.125\textwidth}{\includegraphics[width=0.125\textwidth]{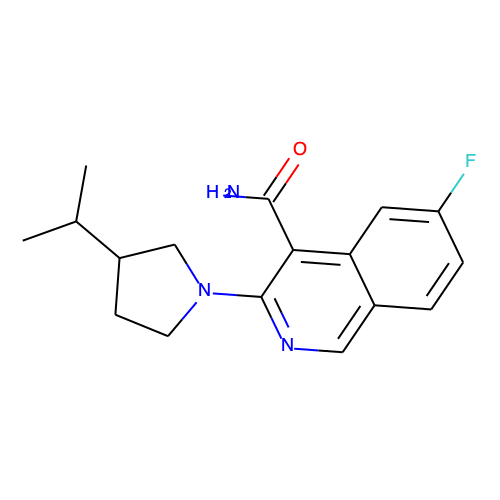}} &      7.75 &  \multirow{2}{*}[-0.35in]{0.14} & 0.95 & 2.97 &  2.96 \\*
           & original &           \parbox[c]{0.125\textwidth}{\includegraphics[width=0.125\textwidth]{SVGs/21_rand_original.png}} &      4.75 &                                 & 0.62 & 4.99 &  0.49 \\
\hline
\multirow{2}{*}[-0.35in]{22} & improved &       \parbox[c]{0.125\textwidth}{\includegraphics[width=0.125\textwidth]{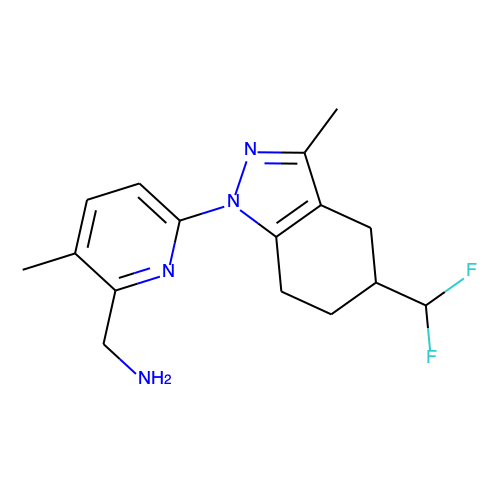}} &      7.58 &  \multirow{2}{*}[-0.35in]{0.04} & 0.95 & 3.55 &  2.71 \\*
           & original &           \parbox[c]{0.125\textwidth}{\includegraphics[width=0.125\textwidth]{SVGs/22_rand_original.png}} &      6.86 &                                 & 0.38 & 3.45 &  1.96 \\
\hline
\multirow{2}{*}{Average} & improved &  &  \textbf{7.72} &  \multirow{2}{*}{0.10} & \textbf{0.95} & 3.34 &  2.89 \\
        & original &  &      6.03 &                                 & 0.57 & 3.75 &  1.88 \\
\bottomrule
\caption{Results of QMO optimization on 23 randomly generated molecules with M\textsuperscript{pro}-affinity $\ge7.5$, max-QED objective.}
\label{tab:Mpro_rand_QED}
\end{longtable}
\end{center}

\end{document}